\documentclass[journal]{IEEEtran}
\usepackage{cite}
\usepackage{color}
\usepackage{amsmath,amssymb,amsfonts}
\usepackage{algorithmic}
\usepackage{subfigure}
\usepackage{graphicx}
\usepackage{textcomp}
\usepackage{underscore}
\usepackage{leftidx}
\usepackage{algorithm}
\usepackage{algorithmic}
\usepackage{graphicx}
\usepackage{multirow}
\usepackage{multicol}
\usepackage{booktabs}
\usepackage{float}

\begin{document}
\title{Super-Resolution Domain Adaptation Networks for Semantic Segmentation via Pixel and Output Level Aligning}

\author{Junfeng Wu,
        Zhenjie Tang,
        Congan Xu,
        Enhai Liu,
        Long Gao,
        Wenjun Yan
\thanks{This research was funded by the National Natural Science Foundation of China under the grant number 62001251, 62001252, 61790550, 61790554 and 61971432, Young Elite Scientists Sponsorship Program by CAST under the grant number 2020-JCJQ-QT-XXX, the Beijing-Tianjin-Hebei Basic Research Cooperation Project under the grant number F2021203109, .}% <-this % stops a space
\thanks{Junfeng Wu, Congan Xu, Long Gao and Wenjun Yan are with the Naval Aeronautical University, Shandong 264000, China (e-mail: patrickwu0609@163.com; xcatougao@163.com; gaolong14@nudt.edu.cn; 434942047@qq.com).}% <-this % stops a space
\thanks{Zhenjie Tang and Enhai Liu are with the School of Artificial Intelligence, Hebei University of Technology, Tianjin 300401, China, and also with the Hebei Province Key Laboratory of Big Data Calculation, Tianjin 300401, China (e-mail:liuenhai@scse.hebut.edu.cn; tangzhenjie.hebut@gmail.com).}
}


%\markboth{IEEE TRANSACTIONS ON IMAGE PROCESSING}%

\markboth{ }%
{Shell \MakeLowercase{\textit{et al.}}: Bare Demo of IEEEtran.cls for IEEE Journals}



\maketitle

\begin{abstract}
Recently, Unsupervised Domain Adaptation (UDA) has attracted increasing attention to address the domain shift problem in the semantic segmentation task. Although previous UDA methods have achieved promising performance, they still suffer from the distribution gaps between source and target domains, especially the resolution discrepany in the remote sensing images. To address this problem, this paper designs a novel end-to-end semantic segmentation network, namely Super-Resolution Domain Adaptation Network (SRDA-Net).  SRDA-Net can simultaneously achieve the super-resolution task and the domain adaptation task, thus satisfying the requirement of semantic segmentation for remote sensing images which usually involve various resolution images. The proposed SRDA-Net includes three parts: a Super-Resolution and Segmentation (SRS) model which focuses on recovering high-resolution image and predicting segmentation map, a Pixel-level Domain Classifier (PDC) for determining which domain the pixel belongs to, and an Output-space Domain Classifier (ODC) for distinguishing which domain the pixel contribution is from. By jointly optimizing SRS with two classifiers, the proposed method can not only eliminates the resolution difference between source and target domains, but also improve the performance of the semantic segmentation task. Experimental results on two remote sensing datasets with different resolutions demonstrate that SRDA-Net performs favorably against some state-of-the-art methods in terms of accuracy and visual quality. Code and models are available at https://github.com/tangzhenjie/SRDA-Net.
\end{abstract}


% \fi
\IEEEpeerreviewmaketitle

\section{Introduction}
\IEEEPARstart{R}{emote} sensing imagery semantic segmentation, aiming at assigning a semantic label for each pixel, has enabled various high-level applications, such as land-use survey, urban planning and environmental protection \cite{Zheng2017, Pan2019, Mou2020}. Deep convolutional neural networks (CNNs) have already shown amazing performance in the semantic segmentation task \cite{Long2015Fully, Chen2018Deeplab, Wang2019, Pan2019DSSNet}. To guarantee the superior representation ability, CNNs usually require a huge number of manually labeled training data. However, the manually annotating process for each pixel is time-consuming and labor-intensive.

UDA tries to learn a well-performed model for the target domain only under the supervision of the source data, and has become to be a powerful technology to handle the problem of insufficient labeling. Most UDA-related works focus on aligning features of source and target domains in a deep network by extracting domain-invariant features \cite{Zhang2018Fully, Wu2019Ace}. In recent years, some works begin seeking to minimize domain shift at the pixel-level, by means of turning source domain images into target-like images through adversarial training \cite{Zhang2018Fully, Li2019Bidirectional}. In addition, some studies are proposed to addresss this problem by reducing the spatial structure domain discrepancies in the output space \cite{Tsai2018Learning, Vu2019}.

However, these typical algorithms mainly address the semantic segmentation problem on natural scene image, and the performance would be influenced when applied on remote sensing images because of the spatial resolution difference. Spatial resolution \cite{Pan2019Super,Liu2019} is one of the important characteristics of remote sensing images. Unlike natural scene images, the sensors used to acquire remote sensing images usually have significant differences, which results in different spatial resolutions. Moreover, the definition of resolution in remote sensing image is not the same as that in natural scene. For example, there may be both large and small cars in a natural scene image, however, a car in a 4m-resolution remote sensing image can never be the same size as a car in a 1m-resolution image. On the other hand, if we only considered UDA for remote sensing images with the same resolution \cite{Liu2020, Tasar2020}, the available data should be severely compressed. Therefore, we may conclude that UDA for remote sensing images should not only narrow the gaps between source and target domains, but also address the issue of different resolutions.

To the best of our knowledge, there are few UDA algorithms for remote sensing images that explicitly consider the resolution problem. The existing algorithms usually neglecte the resolution problem when the resolution differences between source and target domains are not obvious \cite{Yan2020Triplet, Zhang2020Domain}, or deal with the problem by simply interpolation \cite{Zhang2020Unsupervised} or adjusting the parameters of kernel function \cite{Liu2020MultiKernel}. For instance, Yan \emph{et al}. \cite{Yan2020Triplet} proposed a triplet adversarial domain adaptation method to learn a domain-invariant classifier in output-space by a novel domain discriminator, without considering resolution problem between the source and target domains. Instead of matching the distributions in output-space, Zhang \emph{et al}. \cite{Zhang2020Unsupervised} proposed to eliminate the domain shift by aligning the distributions of the source and target data in the feature space, where the resolution problem was dealt with interpolation. Liu \emph{et al}. \cite{Liu2020MultiKernel} minimized the feature distributions distance between source and target domains through metric under different kernel functions, which reduced the effect of resolution problem by adjusting the parameters of kernel function. However, the existing UDA methods for remote sensing images have not explicitly studied the resolution problem.

 \begin{figure}[!t]
 \centering
 \includegraphics[width=8.5cm, height=9cm]{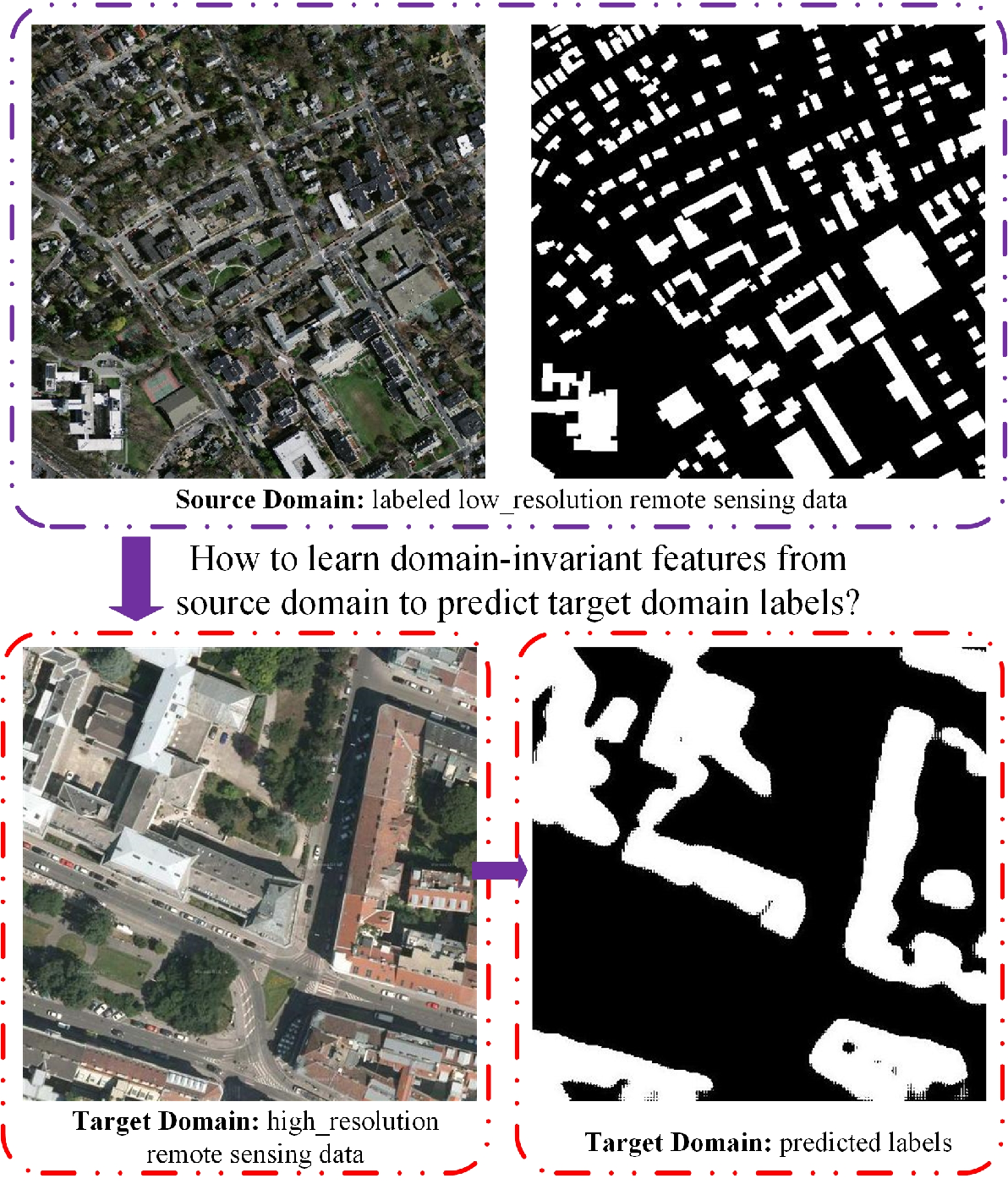}
 \caption{The description of the problem setting: given a source domain composed of labeled low-resolution remote sensing data, and a target domain made up of unlabeled high-resolution remote sensing data, this task intends on predicting the label map for image from the target domain using a semantic segmentation model trained by source domain images.}
 \label{fig:example}
 \end{figure}

In this paper, explicitly considering the resolution problem, a novel end-to-end network is designed, which can simultaneously conduct Super-Resolution and Domain Adaptation, to improve the segmentation performance from low-resolution remote sensing data to high-resolution remote sensing data. Figure~\ref{fig:example} briefly depicts the problem setting: source domain (low-resolution remote sensing images) with labels and target domain (high-resolution remote sensing images) without labels. SRDA-Net is motivated by two recent research: 1) super-resolution and semantic segmentation can promote each other, 2) adversarial training based UDA methods for semantic segmentation.  Recently, some studies have shown that super-resolution and semantic segmentation can boost each other. For instance, researchers indicate that super-resolution results can be improved by semantic priors, such as semantic segmentation probability maps \cite{Wang2018Recovering} or segmentation labels \cite{Rad2019Srobb}. In the field of remote sensing, high-resolution images contain more detailed informaition, and this is very important for image segmentation \cite{Lei2019Simultaneous}. Lei \emph{et al}. \cite{Lei2019Simultaneous} proposed to embed image super-resolution into segmentation network to improve the performance on both super-resolution and segmentation tasks. Furthermore, most of the UDA methods successfully reduce the domain discrepancies drawing support from the adversarial training. For instance, Zhang \emph{et al}. \cite{Zhang2018Fully} applied the adversarial loss to the lower layers of the segmentation network because the lower layers mainly capture the appearance information of the images. Tsai \emph{et al}. \cite{Tsai2018Learning} employed the adversarial feature learning in the output space over the base segmentation model. Vu \emph{et al}. \cite{Vu2019} also reduced the discrepancies of feature distributions in output space through Adversarial Entropy Minimization.

To be specific, the SRDA-Net consists of three networks: a multi-task model for Super-Resolution and semantic Segmentation (SRS), a Pixel-level Domain Classifier (PDC) and an Output-space Domain Classifier (ODC). By integrating a super-resolution network and a segmentation network into one architecture, SRS can eliminate the resolution gap between the source and target domains, and further enhances the semantic segmentation capability. PDC is fed with high-resolution images generated by super-resolution network, and outputs their domain (source or target domain) for each pixel. ODC is fed with the predicted label distributions from segmentation network, then outputs the domain class for each pixel label distribution. Similar to generative adversarial networks (GANs) \cite{Goodfellow2014Generative}, SRS model can be regarded as a generator, while PDC/ODC models can be treated as two discriminators. Through the adversarial training, SRS model can learn domain-invariant features at both the pixel and output-space levels.

To summarise, the major contributions of SRDA-Net can be stated as follows:
\begin{itemize}
\item A new UDA method named SRDA-Net is proposed for semantic segmentation, to adapt the changes from low-resolution remote sensing images to high-resolution remote sensing images.
\item A multi-task model composed of super-resolution and segmentation is built, which not only eliminates the resolution difference between source and target domains, but also obtains improvements on the semantic segmentation task.
\item Two domain classifiers are designed at the pixel-level and output space, to pursue domain alignment. With the help of adversarial training, the domain gap can be effectively reduced.
\end{itemize}

%\subsection{Subsection Heading Here}
%Subsection text here.

%\IEEEpubidadjcol

%\subsubsection{Subsubsection Heading Here}
%Subsubsection text here.
\begin{figure*}[!ht]
 \centering
 \includegraphics[width=18cm]{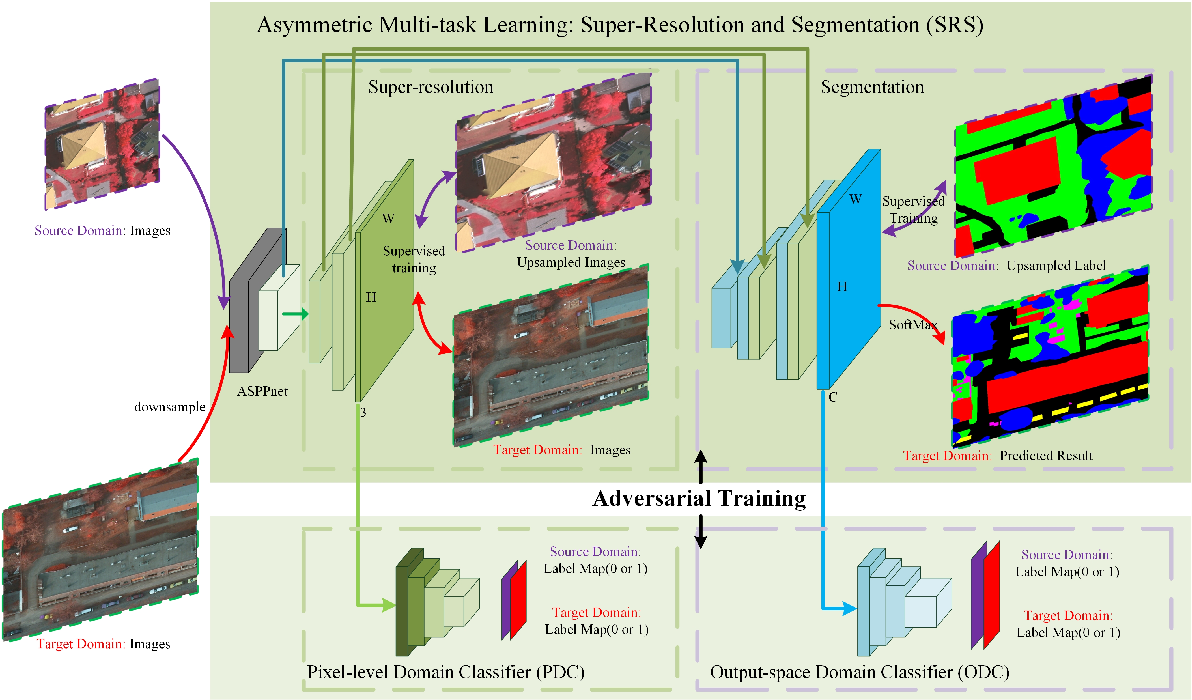}
 \caption{The overview of the proposed Super-Resolution Domain Adaptation Network (SRDA-Net). The two model, Super-Resolution model and the Segmentation model, are integrated together based on asymmetric multi-task learning, as presented in the upper part. At the stage of training, two images randomly selected from source domain and target domain respectively are input into the SRS model. The \textcolor[RGB]{128,0,128}{purple} and \textcolor{red}{red} curve respectively indicate the input/output of source and target domains. And, the two-way arrow indicates the data flow involved in the training process. Note that source images are used for training the Super-Resolution model and Segmentation model in the supervised mode, while the target images only participate the super-resolution training. The two domain classifiers, PDC and ODC, are displayed in left and right bottom parts respectively. The Super-Resolution images from SRS are fed into PDC, and the predicted label distributions from SRS are fed into ODC. The SRS can be optimized by adversarially training SRS and the two classifiers. In the testing stage, the downsampled test images are input into the SRS for predicting the segmentation maps.}
 \label{fig:SRDA-net}
 \end{figure*}

\section{RELATED WORKS}
This section briefly reviews some important works about semantic segmentation, single image super resolution, unsupervised domain adaptation.

\subsection{Semantic Segmentation}
Semantic segmentation aims to assign a semantic label to each pixel in an image. It plays an important role in many fields, such as autonomous driving, urban planning, etc. In 2014, Fully Convolutional Network (FCN) \cite{Long2015Fully} presents amazing performance in some pixel-wise tasks (such as semantic segmentation). After that, the models based on FCN have made significant improvements on several segmentation benchmarks \cite{Maggiori2017Can}. Some model variants are then proposed to exploit the contextual information by adopting multi-scale inputs \cite{Chen2018Deeplab, Ding2020Semantic} or employing probabilistic graphical models \cite{Zheng2017}. For instance, Chen \emph{et al}. \cite{Chen2018Deeplab} proposed a dilated convolution operation to aggregate multi-scale contextual information. Ding \emph{et al}. \cite{Ding2020Semantic} introduced a two-stage multiscale training strategy to incorporate enough context information.

\subsection{Single Image Super Resolution}
Single Image Super Resolution \cite{Freeman2000Learning} attempts to recover high-resolution images from the corresponding low-resolution ones, which has been applied broadly in many occasions, such as product quality inspection, medical diagnosis and remote sensing image reconstruction. The conventional non-CNNs methods mainly focus on the domain and feature priors. For example, interpolation methods such as bicubic and Lanczos generate the high-resolution pixels by the weighted average of neighboring low-resolution pixels. However, CNNs-based methods \cite{Haut2018A,Arun2020CNN} consider the super resolution as a mapping from the low-resolution space to high-resolution space in an end-to-end manner, showing great breakthrough. For example, Arun \emph{et al}. designed a 3-D super resolution neural network for hyperspectral images. Besides, some researchers proposed the perceptual loss \cite{Johnson2016Perceptual} and adversarial training \cite{Lei2020Coupled} to improve perceptual quality of super resolution result.

\subsection{Unsupervised Domain Adaptation}
Considering the work mainly focuses on visual semantic segmentation, the review of UDA is limited in this task as well. Many UDA-based segmentation approaches \cite{Zhang2018Fully, Wu2019Ace} employ adversarial training to minimize cross-domain discrepancy in the feature space. Some works \cite{Tsai2018Learning,Vu2019} propose to align the predicted label distributions in the output space. Tsai \emph{et al}. \cite{Tsai2018Learning} carried out the alignment on the prediction of the segmentation network and Vu \emph{et al}. \cite{Vu2019} proposed to do it on entropy minimization of the prediction probability. In contrast, pixel-level domain adaptation \cite{Zheng2020Do, Tasar2020, Li2019Bidirectional} makes use of generative networks to turn source domain images into target-like images. Li \emph{et al}. \cite{Li2019Bidirectional} presented a bidirectional learning system for semantic segmentation, which is a closed loop to learn the segmentation adaptation model and the image translation model alternatively, causing the domain gap to be reduced gradually at the pixel-level. Besides, a curriculum learning strategy is proposed in \cite{Zhang2019A} by leveraging information from global label distributions and local super-pixel distributions of target domain.

\section{The Proposed Approach}
In this section, we discuss the methodology of the proposed SRDA-Net and the overall framework is shown in Figure~\ref{fig:SRDA-net}. In order to reduce the resolution domain gap, we integrate the super-resolution into the segmentation model to eliminate the impact of different resolutions. By optimizing the SRS as well as two domain classifiers (PDC and ODC) with adversarial optimizing, the domain gap in pixel-level and output-space can be gradually reduced, thus improving the performance.

\subsection{Problem Description}
It is worthy of defining cross-domain semantic segmentation with mathematical notations, before illustrating the method in detail. Formally, Let's suppose $\mathcal{S}$ as a source domain from a low-resolution remote sensing dataset, where low-resolution images ${I}_\mathcal{S}$ and pixel-level annotations ${A}_\mathcal{S}$ are provided; $\mathcal{T}$ as a target domain from high-resolution remote sensing dataset, which only provides high-resolution images ${I}_\mathcal{T}$. Note that the label space of $\mathcal{S}$ and $\mathcal{T}$ is the same, denoted as $\mathbb{R}^{C}$, where the $C$ denotes the number of categories. In a word, given ${I}_\mathcal{S}$, ${A}_\mathcal{S}$, ${I}_\mathcal{T}$, our goal is to reduce the domain gap (including resolution difference) between $\mathcal{S}$ and $\mathcal{T}$ and learn a segmentation model to predict pixel-wise category of $\mathcal{T}$. In the following, we first describe the asymmetric multi-task (super-resolution and segmentation) model. Then, the adversarial domain adaptation (pixel-level and output-space) is presented in details.

\subsection{Multi-task Model: Super-Resolution and Segmentation} %Super Resolution for Segmentation
% reason
Over the past few years, CNNs-based methods have been widely applied to solve the semantic segmentation problem. However, those methods may perform worse when generalizing to the unseen images, especially the domain gap between the training (source domain) and test (target domain) images are obvious. This problem is critical for remote sensing images, because the resolution of them usually changes dramatically, which seriously affects the generalization ability of the segmentation models. Therefore, it is important to eliminate the resolution difference for cross-domain semantic segmentation in remote sensing.

Recently, some studies \cite{Wang2018Recovering,Lei2019Simultaneous} show that super-resolution and semantic segmentation can boost each other. Although super-resolution and segmentation are two different and challenging tasks, they may have certain relationship. Super-resolution can provide images with more details, which is helpfull for improving the segmentation accuracy. Label maps from segmentation dataset or semantic segmentation probability maps may contribute to recover textures faithful to semantic classes during super-resolution process.

Based on the above discussions, we propose a novel model based on the aysmmetric multi-task learning, which consists of Super-Resolution and Segmentation models. In order to make super-resolution and segmentation boost each other, two strategies are adopted: (1) introducing a pyramid feature fusion structure between the two tasks; (2) imposing the the cross-entropy segmentation loss to train the segmentation network, for the generated high-resolution images of source domain. During traing, a source domain image, pairing with a downsampled target domain image, are taken as the inputs to the SRS network. The source doamin images are used to train both the super-resolution network and the segmentation network, while the target images only participate in the super-resolution training process, shown as Figure~\ref{fig:SRDA-net}. At the testing stage, the downsampled test images are input into the SRS network to obtain the pixel-wise scope maps.

To be specific, due to GPU memory limitation, we employ residual Atrous Spatrial Pyramid Pooling (ASPP) Module \cite{Wang2019Learning} as the shared feature extractor. For the super-resolution model, we only use a few deconvolutions to recover the high-resolution images, without using pixelshuffle \cite{ShiReal}. To transfer the low-level features effectively from super-resolution stream to segmentation stream, we introduce the pyramid feature fusion structure \cite{Lin2017Feature} between the two streams. Moreover, the super-resolution results of source domain are also fed to the segmentation stream. Meanwhile, the segmentation stream also ensures to recover textures faithful to semantic classes during super-resolution stream.

%0.5 \times
The proposed SRS model is optimized through the following loss:
\begin{equation}
\begin{split}
\mathcal{L}_{S R S}= \alpha \mathcal{L}_{s e g} + \beta \left( \mathcal{L}_{i d T} + \mathcal{L}_{i d S}  \right)
\end{split}
\end{equation}
\begin{equation}
\mathcal{L}_{s e g} = \mathcal{L}_{c e l}\left(\mathbf{S}\left(I_{\mathcal{S}}\right), \uparrow A_{\mathcal{S}}\right) + \mathcal{L}_{c e l}\left(\mathbf{S} \left(\downarrow \mathbf{R}\left(I_{\mathcal{S}}\right)\right), \uparrow A_{\mathcal{S}}\right)
\end{equation}
\begin{equation}
\mathcal{L}_{i d T} =  \mathcal{L}_{m s e}\left(\mathbf{R}\left(\downarrow I_{\mathcal{T}} \right), I_{\mathcal{T}}\right)
\end{equation}
\begin{equation}
\mathcal{L}_{i d S} = \mathcal{L}_{p e r}\left(\mathbf{R}\left(I_{\mathcal{S}}\right), \uparrow I_{\mathcal{S}}\right) + 0.5 \times \mathcal{L}_{f p}
\end{equation}
\begin{equation}
\mathcal{L}_{f p}=\mathcal{L}_{L 1}\left( \mathbf{E} \left( \downarrow \mathbf{R}\left(I_{\mathcal{S}}\right)\right), \mathbf{E} \left( I_{\mathcal{S}}\right)\right)
\end{equation}
where $\mathcal{L}_{c e l}$ represents the 2D Cross Entropy Loss, the standard supervised pixel-wise classification objective function \cite{Wang2019}; $\mathcal{L}_{m s e}$ is the pixel-wise Mean Squared Error (MSE) loss,  which is widely applied to optimize the objective function for image super resolution; $\mathcal{L}_{p e r}$ is the perceptual loss \cite{Johnson2016Perceptual}; $\mathcal{L}_{L 1}$ represents the L1 norm loss; and $\mathcal{L}_{f p}$ is the fixpoint loss \cite{Kotovenko2019A}. The $\uparrow$ and $\downarrow$ denote upsampling and downsampling operations, respectively. $\mathbf{S}$, $\mathbf{R}$, and $\mathbf{E}$ denote Segmentation model, super-Resolution model and the shared feature Extractor, respectively. Note that, in order to easily superimpose the style of target domain and stabilize the adversarial training process,  we use the $\mathcal{L}_{p e r}$ and $\mathcal{L}_{f p}$ to train the super-resolution model of source domain images. $\alpha$, $\beta$ denote the weighting factors for semantic segmentation and super-resolution.

\subsection{Adversarial Domain Adaptation}
Although the proposed asymmetric multi-task model can eliminate the resolution difference between source and target domains, some other domain gaps (e.g. color, texture etc) are still exist. Affected by various human and natural factors, such as sensors, weather conditions, imaging locations, these differences are inherent in remote sensing imagery. Therefore, how to learn the domain-invariant features for remote sensing imagery is a critical problem.

An effective framework to deal with the above-mentioned problem is adversarial learning. It consists of two main parts: a generator network and a discriminator network. Its main idea is to train the discriminator to predict the domain label of the data, while the generator network attemps to fool it, as well as implements the segmentation task on source domain data. Through training the two networks alternately, the feature domain gap can be gradually reduced, thus obtaining the domain-invariant representations.

The proposed method also takes the adersarial learning to allevate the domain gap. Specially, the PDC and ODC are disigned as discriminators, and the SRS is adopted as the generator. By the adversarial training, SRS will learn the domain-invariant features that fool the PDC and ODC.

\subsection{Pixel-Level Adaptation}
The proposed SRS mode eliminates the resolution difference between source and target domains, while does not reduce the gap in other aspects. To address this problem ulteriorly, PDC is designed to receive the high-resolution images from source or target domain, and classify the domain for each pixel. Concretely, the PatchGAN \cite{Li2016Precomputed} is applied as PDC and the network architecture is shown at the bottom left of Figure~\ref{fig:SRDA-net}. 

The loss objective of PDC can be formulated as following:
\begin{equation}
\begin{split}
I_{\mathcal{S}}^{R} = \mathbf{R} \left(I_{\mathcal{S}}\right) \in \mathbb{R}^{H \times W \times 3}
\end{split}
\end{equation}
\begin{equation}
\begin{split}
I_{fake} = \mathbf{D}_{pdc} \left( I_{\mathcal{S}}^{R} \right) \in \mathbb{R}^{H \times W \times 1}
\end{split}
\end{equation}
\begin{equation}
\begin{split}
I_{true} = \mathbf{D}_{pdc} \left( I_{\mathcal{T}} \right) \in \mathbb{R}^{H \times W \times 1}
\end{split}
\end{equation}
\begin{equation}
\begin{split}
\mathcal{L}_{P D C} &= \mathbb{E}_{I_{fake} \sim p_{\text {data }}(I_{fake})} \left[ (I_{fake}-1)^{2} \right] \\
& + \mathbb{E}_{I_{true} \sim p_{\text {data }}(I_{true})}\left[ (I_{true})^{2} \right]
\end{split}
\end{equation}
where $\mathbf{D}_{pdc}$ is the PDC model,  $H$ and $W$ denote the height and width of the high-resolution target domain image.

Accordingly, the inverse of PDC loss is calculated by:
\begin{equation}
\begin{split}
\mathcal{L}_{{P D C}_{inv}} &= \mathbb{E}_{I_{true} \sim p_{\text {data }}(I_{true})} \left[ (I_{true}-1)^{2} \right] \\
& + \mathbb{E}_{I_{fake} \sim p_{\text {data }}(I_{fake})}\left[ (I_{fake})^{2} \right]
\end{split}
\end{equation}

Finally, the adversarial objective functions are given as:
\begin{equation}
\label{objectiveFunctions1ofPixelLevelAdaptation}
\begin{split}
\min _{\theta_{SRS}} \mathcal{L}_{SRS} + \mathcal{L}_{{P D C}}
\end{split}
\end{equation}
\begin{equation}
\label{objectiveFunctions2ofPixelLevelAdaptation}
\begin{split}
\min _{\theta_{P D C}} \mathcal{L}_{{P D C}_{inv}}
\end{split}
\end{equation}
where $\theta_{SRS}$ and $\theta_{P D C}$ denote the network parameters of SRS and PDC, respectively. During the training phase, the parameters of the two models are updated in turns using Eq. (\ref{objectiveFunctions1ofPixelLevelAdaptation}) and Eq.(\ref{objectiveFunctions2ofPixelLevelAdaptation}).

%$\mathbf{R}$ tries to generate images $\mathbf{R}\left(I_{\mathcal{S}}\right)$ that look similar to images from domain $\mathcal{T}$, while $\mathbf{D}_{pdc}$ aims to distinguish between the translated samples $\mathbf{R}\left(I_{\mathcal{S}}\right)$ and real samples $I_{\mathcal{T}}$. $\mathbf{R}$ aims to minimize this objective function, while $\mathbf{D}_{pdc}$ tries to maximize it.

\subsection{Output-Space Adaptation}
Different from image classification task that based on global features, the generated high-dimensional features for semantic segmentation encode complex detailed representations, which will result in comtextual relationships among neighboring pixels. Therefore, adaptation only in the pixel space may not enough for semantic segmentation. On the other hand, although segmentation outputs are in the low-dimensional space, they contain rich information, e.g., scene layout and context. Moreover, in the segmentation task of remote sensing, images from the source or target domain should share strong similarities both in spatial and local representations. For example, the rectangular road region may cover the part of cars, pedestrians, and green plants often grow around the buildings. Thus, we adapt the low-dimensional softmax outputs of segmentation predictions via an adversarial learning scheme.

To be specific, we design ODC to distinguish domain source for the distribution of pixels, which receives the segmentation softmax output: $P=\mathbf{S}(I) \in \mathbb{R}^{H \times W \times C}$, where $C$ is the number of categories. We forward $P$ to ODC using a cross-entropy loss $\mathcal{L}_{O D C}$ for the two classes (i.e., source and target). The ODC loss can be written as:
\begin{equation}
\begin{split}
P_{\mathcal{S}}=\mathbf{S}(I_{\mathcal{S}}) \in \mathbb{R}^{H \times W \times C}
\end{split}
\end{equation}
\begin{equation}
\begin{split}
P_{\mathcal{T}}=\mathbf{S}(\downarrow I_{\mathcal{T}}) \in \mathbb{R}^{H \times W \times C}
\end{split}
\end{equation}
\begin{equation}
\begin{split}
P_{true} = \mathbf{D}_{odc} \left( P_{\mathcal{S}} \right) \in \mathbb{R}^{H \times W \times 1}
\end{split}
\end{equation}
\begin{equation}
\begin{split}
P_{fake} = \mathbf{D}_{odc} \left( P_{\mathcal{T}} \right) \in \mathbb{R}^{H \times W \times 1}
\end{split}
\end{equation}
\begin{equation}
\begin{split}
\mathcal{L}_{O D C} &= -\sum_{h, w}(1 - z) \log \left(P_{fake}\right) + z \log \left(P_{true}\right)
\end{split}
\end{equation}
where $\mathbf{D}_{odc}$ denotes the ODC model.

Accordingly, the inverse of ODC loss is defined as:
\begin{equation}
\begin{split}
\mathcal{L}_{{O D C}_{inv}} &= -\sum_{h, w}(1 - z) \log \left(P_{true}\right) + z \log \left(P_{fake}\right)
\end{split}
\end{equation}

In the end, the adversarial objective functions are expressed as follows:
\begin{equation}
\label{objectiveFunction1ofOutputSpaceAdaptation}
\begin{split}
\min _{\theta_{SRS}} \mathcal{L}_{SRS} + \mathcal{L}_{{O D C}}
\end{split}
\end{equation}
\begin{equation}
\label{objectiveFunction2ofOutputSpaceAdaptation}
\begin{split}
\min _{\theta_{O D C}} \mathcal{L}_{{O D C}_{inv}}
\end{split}
\end{equation}
where $\theta_{SRS}$ and $\theta_{O D C}$ represent the parameters of SRS and ODC networks, respectively. They can be optimized in turns by minimizing Eq. (\ref{objectiveFunction1ofOutputSpaceAdaptation}) and Eq. (\ref{objectiveFunction2ofOutputSpaceAdaptation}) during the training stage.

\subsection{Final objective function}
To initialize parameters of the network better, we first use the following loss function to pre-train the model:
\begin{equation}
\label{R}
 \min _{\theta_{R}} \beta \left( \mathcal{L}_{i d T} + \mathcal{L}_{i d S}  \right) + \mathcal{L}_{P D C}
\end{equation}
\begin{equation}
\label{PDC}
\begin{split}
\min _{\theta_{P D C}} \mathcal{L}_{{P D C}_{inv}}
\end{split}
\end{equation}
where $\beta$ denotes a weighting factor for super-resolution, $\theta_{R}$ is the parameters of $\mathbf{R}$ network. During training stage, the $\mathbf{R}$ and PDC networks are optimized in turns using Eq.(\ref{R}) and Eq.(\ref{PDC}).

For the whole models training, including SRS, PDC and ODC, the objective functions can be formulated as:
\begin{equation}
\label{fullObjectiveFunction1}
\min _{\theta_{SRS}} \mathcal{L}_{SRS} + \mathcal{L}_{P D C} + \mathcal{L}_{O D C}
\end{equation}
\begin{equation}
\label{fullObjectiveFunction2}
\begin{split}
\min _{\theta_{D}} \mathcal{L}_{{P D C}_{inv}} + \mathcal{L}_{{O D C}_{inv}}
\end{split}
\end{equation}
where $\theta_{D}$ denotes the network parameters of PDC and ODC. During training phase, the parameters of SRS, PDC and ODC are optimized in turns by minimizing Eq.(\ref{fullObjectiveFunction1}) and Eq.(\ref{fullObjectiveFunction2}). The training procedure of our proposed SRDA-Net is illustrated in Algorithm \ref{alg:Framwork}.
\begin{algorithm}
\caption{the proposed SRDA-Net.} \label{alg:Framwork}
\begin{algorithmic}[1]
\REQUIRE ~~\\ Source Domain low-resolution image $I_{\mathcal{S}}$, Target Domain
high-resolution image $I_{\mathcal{T}}$,
Source Domain low-resolution label $A_{\mathcal{S}}$, The weighting factors for semantic segmentation and super-resolution: $\alpha$, $\beta=10$. \\
\ENSURE ~~\\
High-resolution source domain image with style of target domain: $I_{\mathcal{S}}^{R}$\\
Predict label of target domain: $A_{\mathcal{T}}$ \\

 \REPEAT
 {
 \STATE
 \% Super-Resolution images by the $\mathbf{R}$ model\\
$I_{\mathcal{S}}^{R}, I_{\mathcal{T}}^{R} = \mathbf{R} \left(I_{\mathcal{S}}, \downarrow I_{\mathcal{T}} \right) \in \mathbb{R}^{H \times W \times 3}$

 \STATE
 \% Segmentation softmax outputs from the $\mathbf{S}$ model \\
$P_{\mathcal{S}}, P_{\mathcal{T}} = \mathbf{S} \left(I_{\mathcal{S}}, \downarrow I_{\mathcal{T}}\right) \in \mathbb{R}^{H \times W \times C}$

\STATE
\% Predict label of target domain image \\
$A_{\mathcal{T}} = max \left( P_{\mathcal{T}} \right) \in \mathbb{R}^{H \times W \times 1}$
\STATE
\% Distinguish the pixels of super-resolution images\\
$I_{fake}, I_{true} = \mathbf{D}_{pdc} \left(I_{\mathcal{S}}^{R}, I_{\mathcal{T}}\right) \in \mathbb{R}^{H \times W \times 1}$
\STATE
\% Distinguish the pixel distributions of softmax outputs\\
$P_{true}, P_{fake} = \mathbf{D}_{odc} \left(P_{\mathcal{S}}, P_{\mathcal{T}}\right)  \in \mathbb{R}^{H \times W \times 1}$

 \STATE
 \% Adversarial training\\
 $\mathbf{R}$ and $\mathbf{S}$ can be optimized according to equation (23).\\
 $\mathbf{D}_{pdc}$ and $\mathbf{D}_{odc}$ are updated by minimizing the inverse loss (24).
 }
 \UNTIL convergence
\end{algorithmic}
\end{algorithm}

\section{Experimental Results}
%In this section, we first introduce the two datasets we constructed: Mass-Inria Dataset and Vaih-Pots Dataset. Then, the experimental setup and implementation details are presented. Next, extensive experimental results are shown to demonstrate the effectiveness of our method on the two datasets. Finally, the super-resolution results with style transfer of source domain images from CycleGAN and SRDA-Net are presented, which confirm that the semantic prior of segmentation improves the perform of super-resolution.
In this secation, we validate the performance of the proposed SRDA-Net. Firstly, the experimental dataset and implementation details are described. And then, the exerimental results are reported and analyzed to demonstrate the effectiveness of SRDA-Net. Finally, the two strategies to achieve mutual promotion of super-resolution and segmentation in SRS are discussed.
\subsection{Datasets Description}
%With the development of deep learning in remote sensing, there have been many large remote sensing datasets with different resolutions. Based on this, we constructed two datasets: a single-category Mass-Inria Dataset, and a multi-category Vaih-Pots Dataset.

\subsubsection{Mass-Inria}  the following two UDA datasets are used for single-category semantic segmentation.
\begin{itemize}
\item {\bfseries Massachusetts Buildings Dataset} \cite{Volodymyr2013Machine} contains 151 aerial images of the Boston area at 1m spatial resolution . The ground truth provides two semantic classes: building and nonbuilding. The whole dataset is divided into three parts: a training set with 137 images, a testing set with 10 images and a validation set with 4 images. Among these sets, the training set is considered as the source domain.

\item {\bfseries Inria Aerial Image Labeling Dataset} \cite{Maggiori2017Can} is comprised of 360 tiles with a resolution of 0.3m on ten cities across the globe. Ground-truth provides two semantic classes, \textit{building} and \textit{not building} classes. We split the training set (image 1 to 5 of each location for validation, 6 to 36 of each location for training). We consider the training set of this dataset as target domain. We finally validate the results of the algorithm on the validation set of this dataset.
\end{itemize}
% Due to the differences in image shooting time, shooting location, and shooting sensor, these two datasets have strong domain discrepancies in addition to different resolutions, so they meet the experimental requirements.

\begin{table*}[!ht]
\small    % 控制字体
\renewcommand{\arraystretch}{1.3} %控制行距
\begin{center}
\caption{The comparison results of domain adaptation from Mass to Inria val datasets}
\label{tab:MassToInria}
\setlength{\tabcolsep}{7.4mm}{%调整宽度
\begin{tabular}{lllll}
\toprule[1.2pt] %加粗的横线
\noalign{\smallskip} %
Methods \%    & BaseNet  & Source domain   & Target domain  & {\bfseries IoU}    \\
\noalign{\smallskip}
\hline
\noalign{\smallskip}
NoAdapt \cite{Tsai2018Learning}& Resnet-101 \cite{He2016Deep}& Mass &  Inria  &32.9   \\
AdaptSegNet \cite{Tsai2018Learning}&Resnet-101 \cite{He2016Deep}&Mass&$\downarrow$ Inria&35.0 \\
AdaptSegNet \cite{Tsai2018Learning}&Resnet-101 \cite{He2016Deep}&$\uparrow$ Mass&Inria&48.5 \\
\noalign{\smallskip}
\hline
\noalign{\smallskip}
NoAdapt \cite{Zhang2018Fully}&Resnet-101 \cite{He2016Deep}&Mass&     Inria   &32.9   \\
CycleGan-FCAN \cite{Zhang2018Fully}&Resnet-101 \cite{He2016Deep}&Mass&$\downarrow$ Inria&41.8  \\
CycleGan-FCAN \cite{Zhang2018Fully}&Resnet-101 \cite{He2016Deep}&$\uparrow$ Mass&Inria&49.7   \\
\noalign{\smallskip}
\hline
\noalign{\smallskip}
NoAdapt & ResidualASPP \cite{Wang2019Learning}  & Mass & Inria & 31.9
\\
SRS & ResidualASPP \cite{Wang2019Learning}  & Mass & Inria & 36.7           \\
SRS + PDC    & ResidualASPP \cite{Wang2019Learning}   & Mass & Inria & 46.0 	         \\
SRS + ODC  & ResidualASPP \cite{Wang2019Learning}     & Mass & Inria & 39.4           \\
Full (SRDA-Net)& ResidualASPP \cite{Wang2019Learning}& Mass & Inria &{\bfseries 52.8}\\
\noalign{\smallskip}
\bottomrule[1.2pt] %加粗的横线
\end{tabular}}
\end{center}
\end{table*}

\begin{figure*}
\begin{center}
%\caption{Qualitative results of the Inria val dataset (Source domain: Massachusetts Buildings}
\subfigure[Input image]{
\begin{minipage}[b]{0.11\linewidth}
\includegraphics[width=1\linewidth]{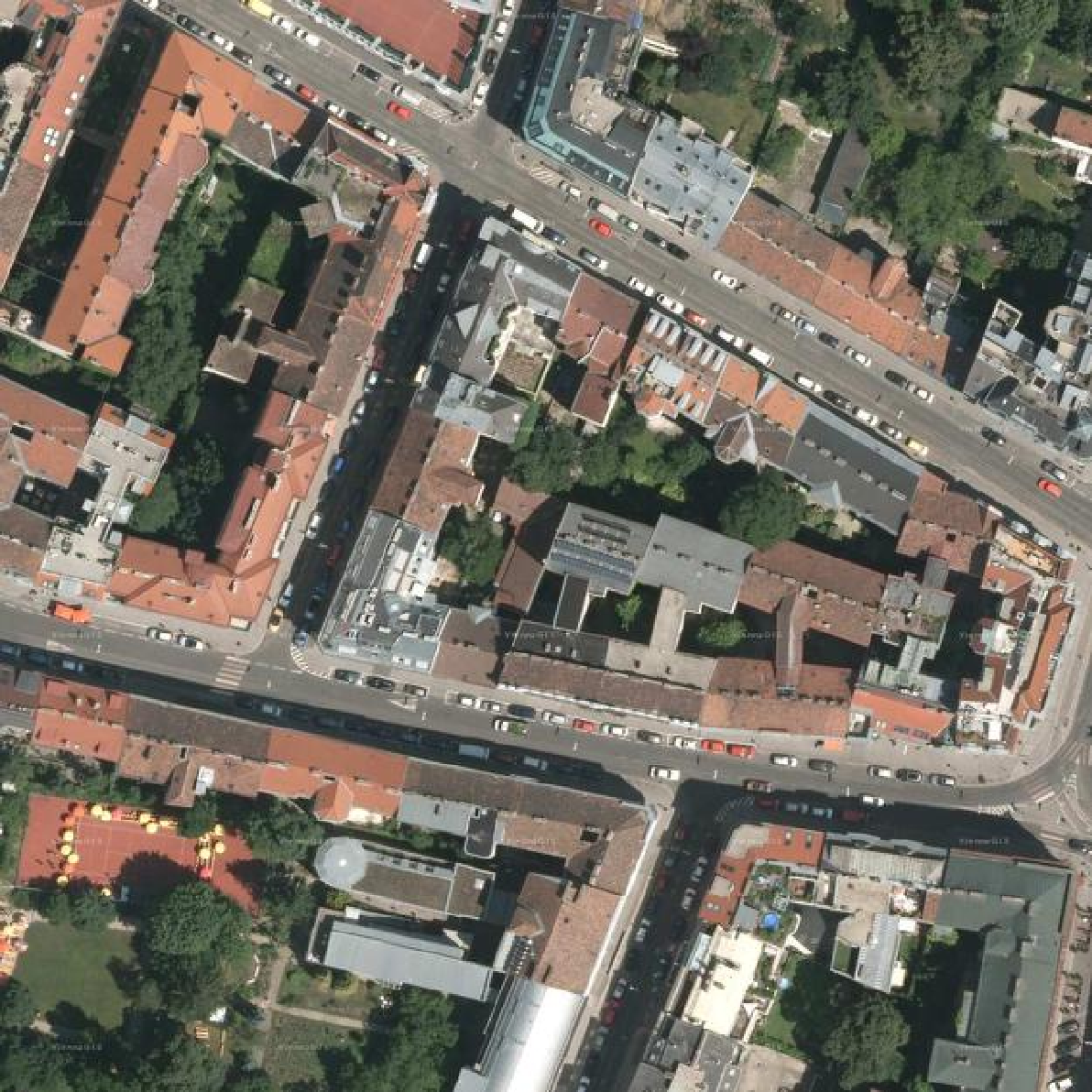}\vspace{4pt}
\includegraphics[width=1\linewidth]{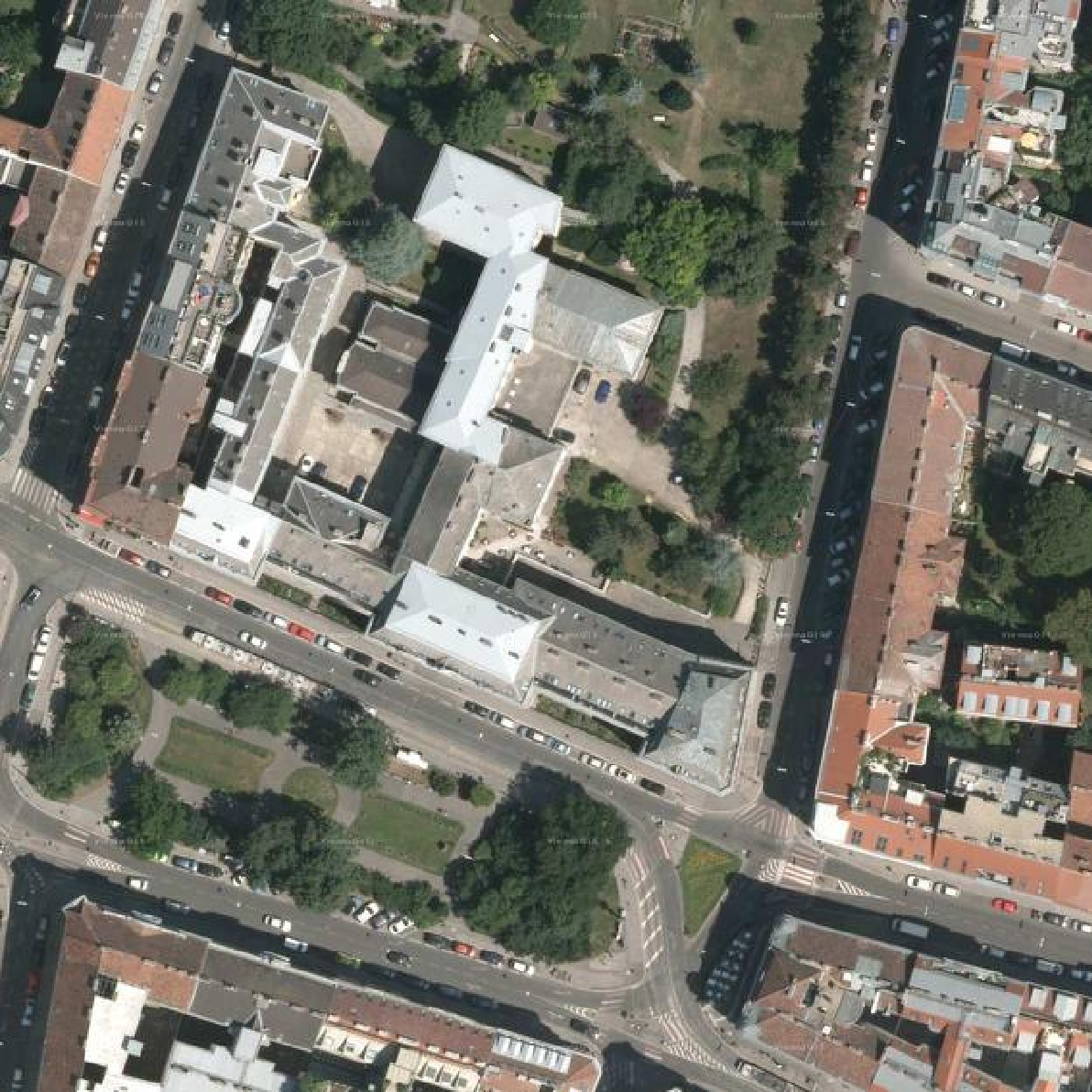}\vspace{4pt}
\includegraphics[width=1\linewidth]{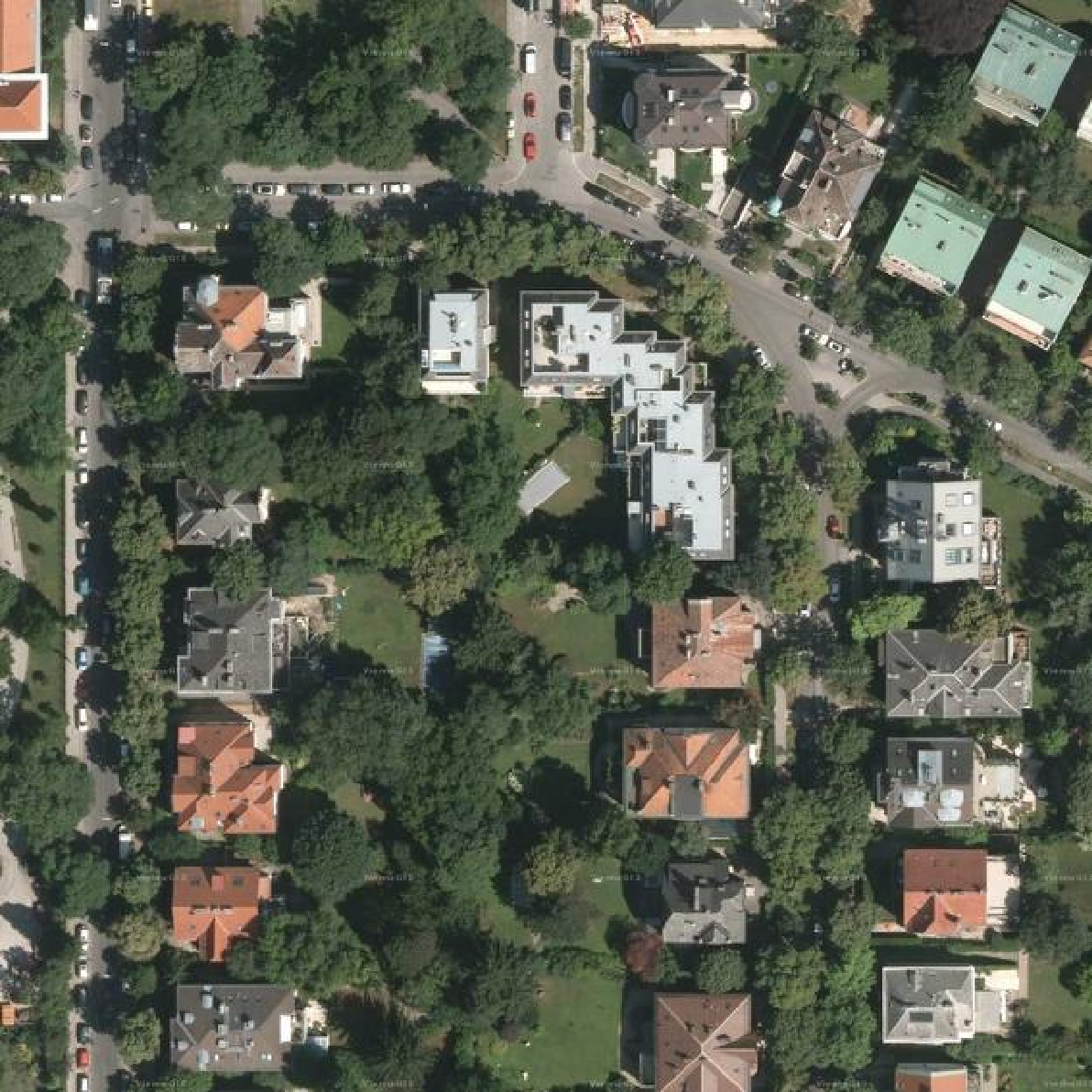}
\end{minipage}}
\subfigure[Ground Truth]{
\begin{minipage}[b]{0.11\linewidth}
\includegraphics[width=1\linewidth]{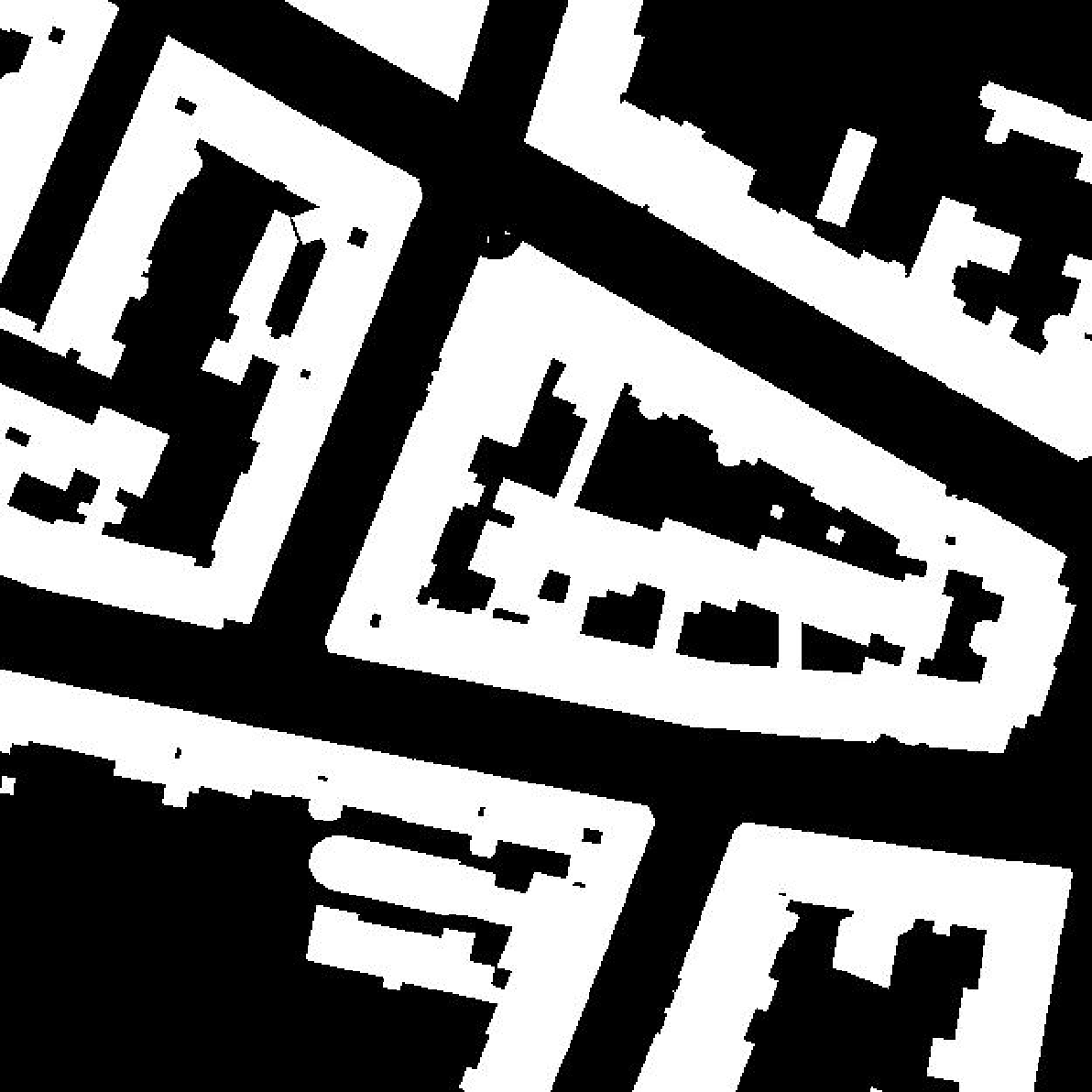}\vspace{4pt}
\includegraphics[width=1\linewidth]{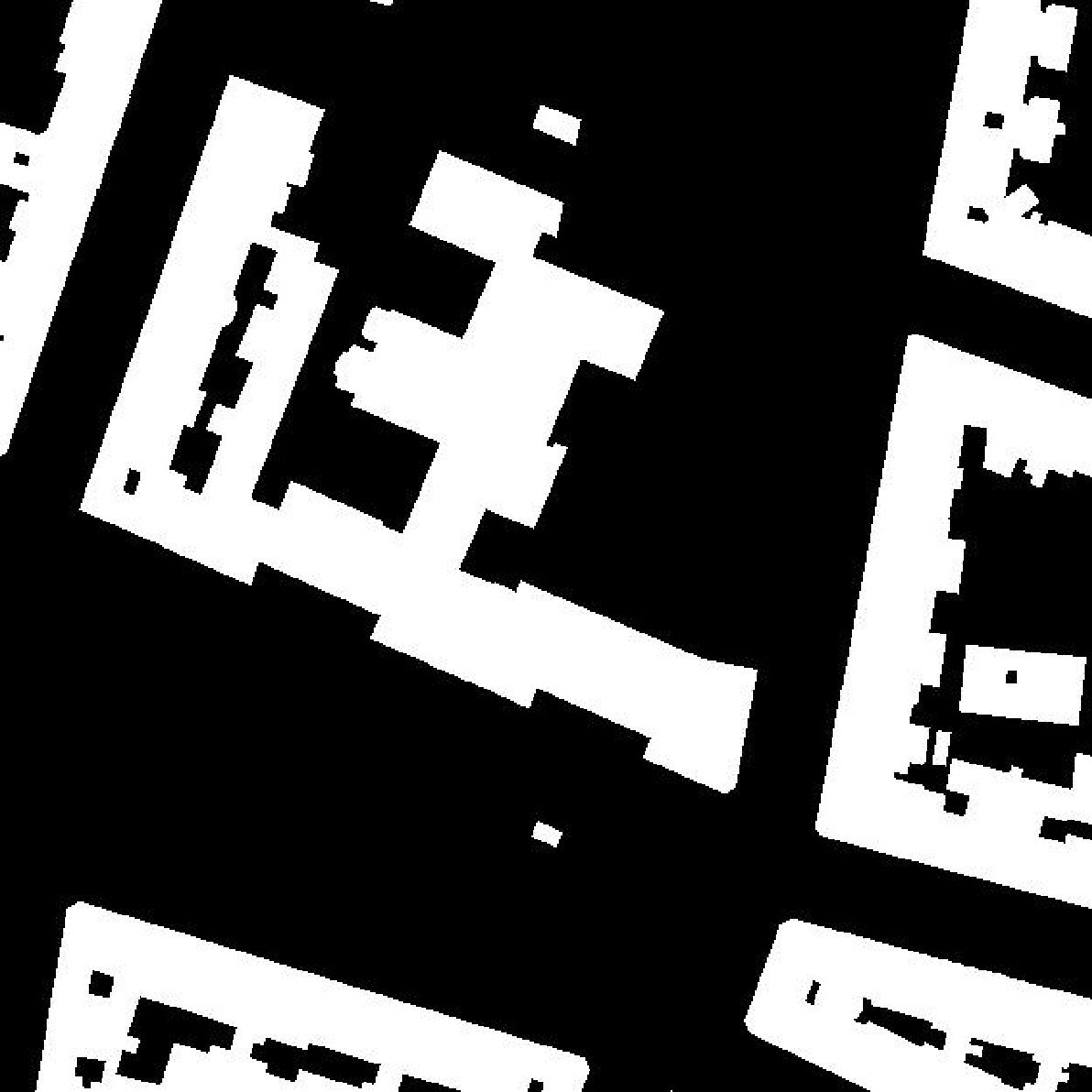}\vspace{4pt}
\includegraphics[width=1\linewidth]{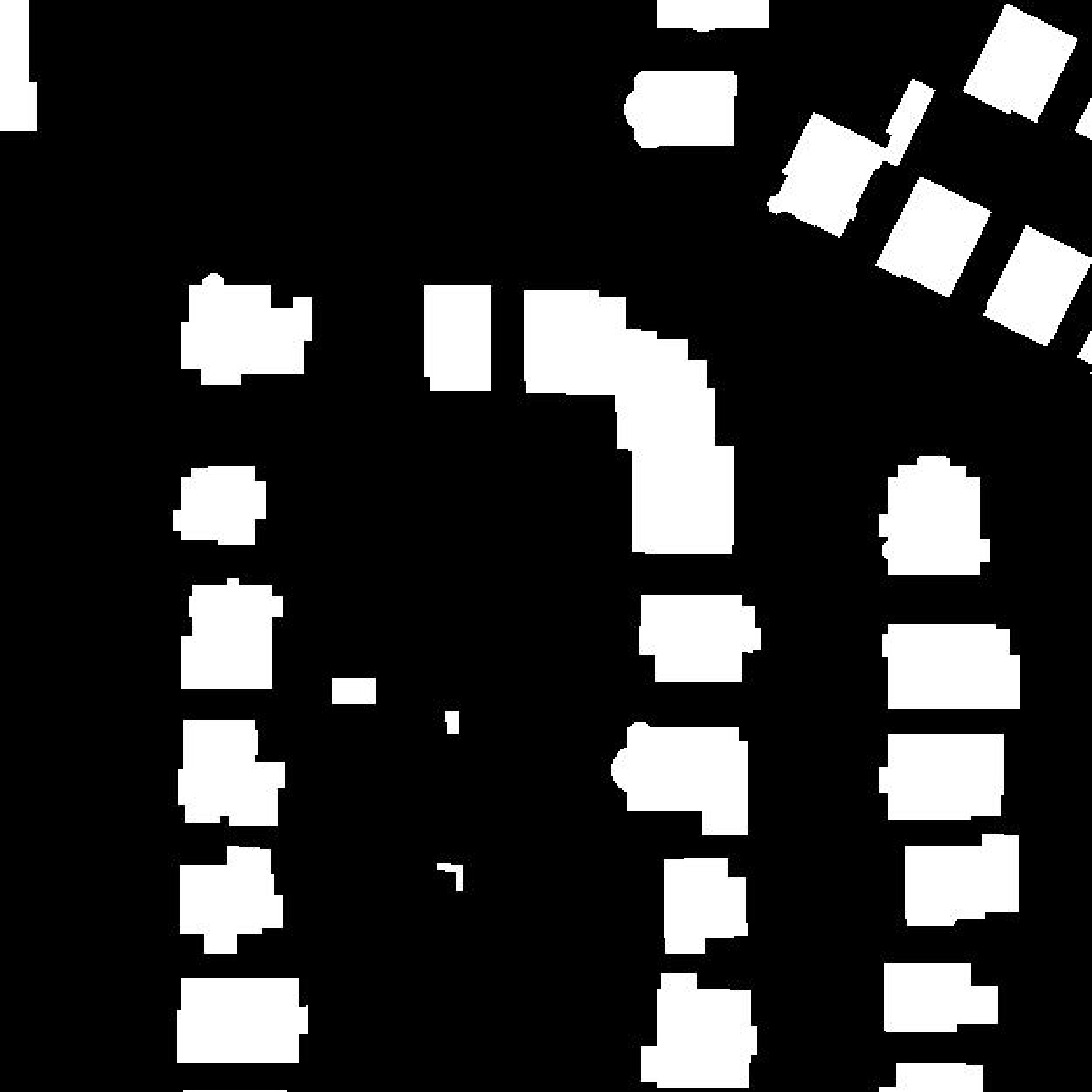}
\end{minipage}}
\subfigure[SRS]{
\begin{minipage}[b]{0.11\linewidth}
\includegraphics[width=1\linewidth]{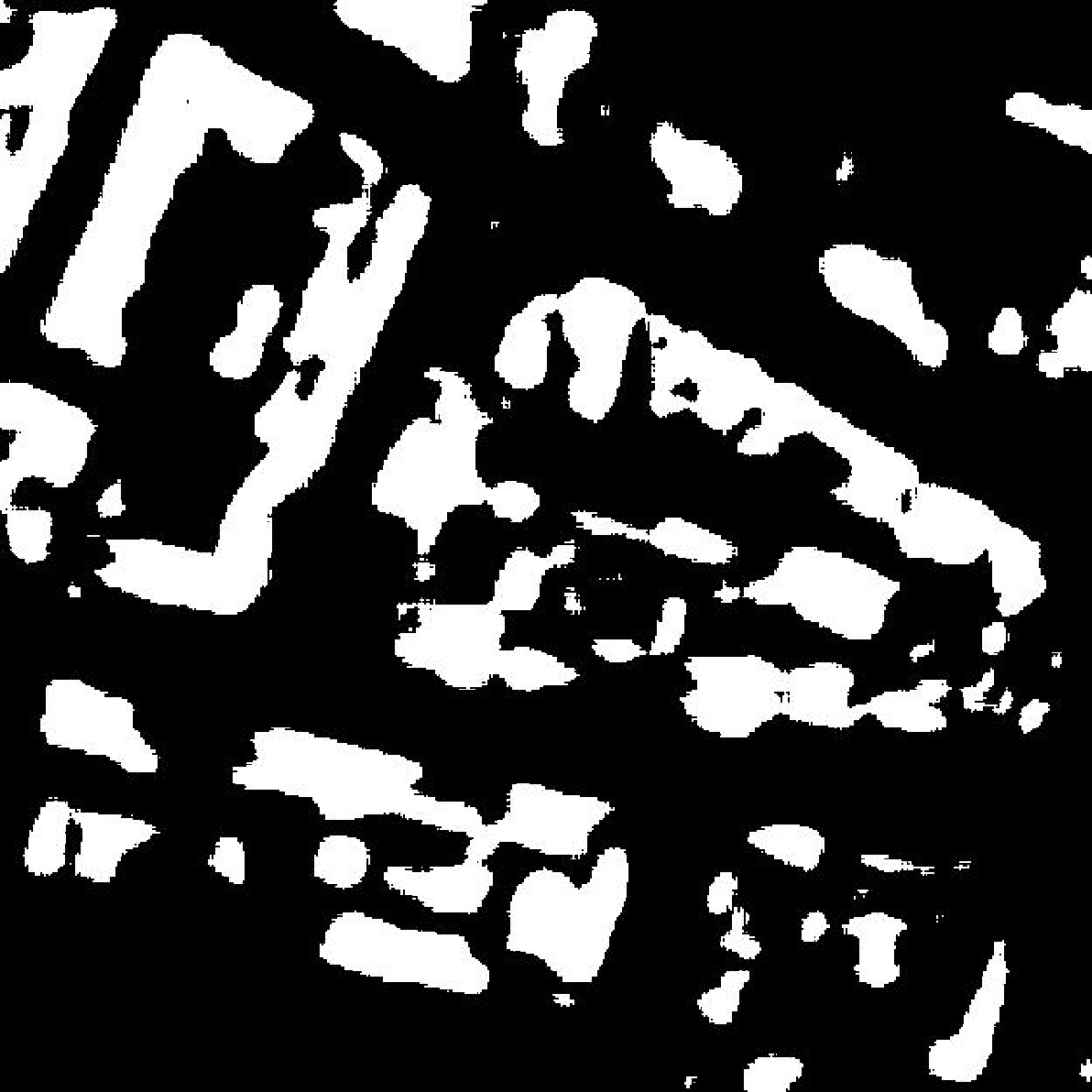}\vspace{4pt}
\includegraphics[width=1\linewidth]{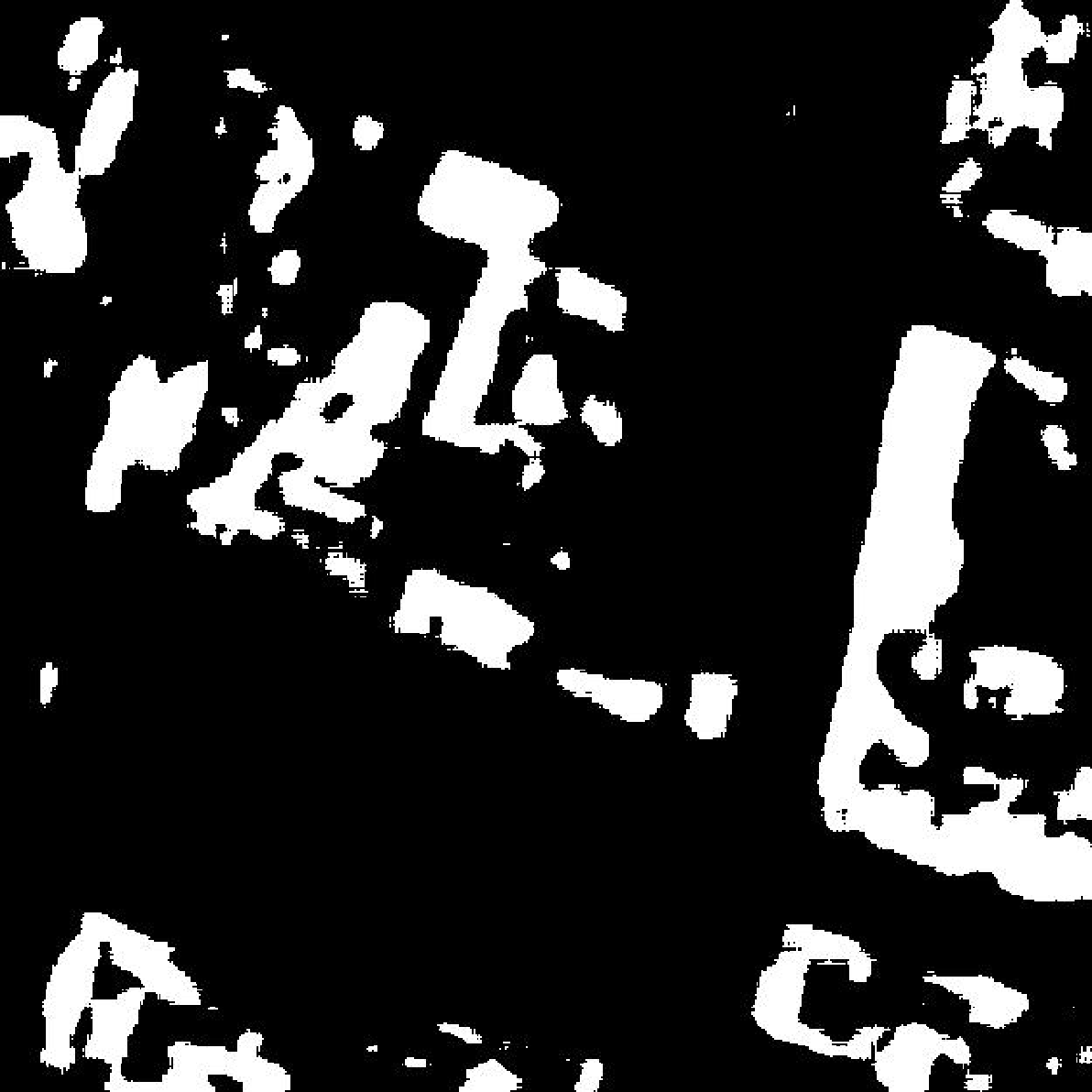}\vspace{4pt}
\includegraphics[width=1\linewidth]{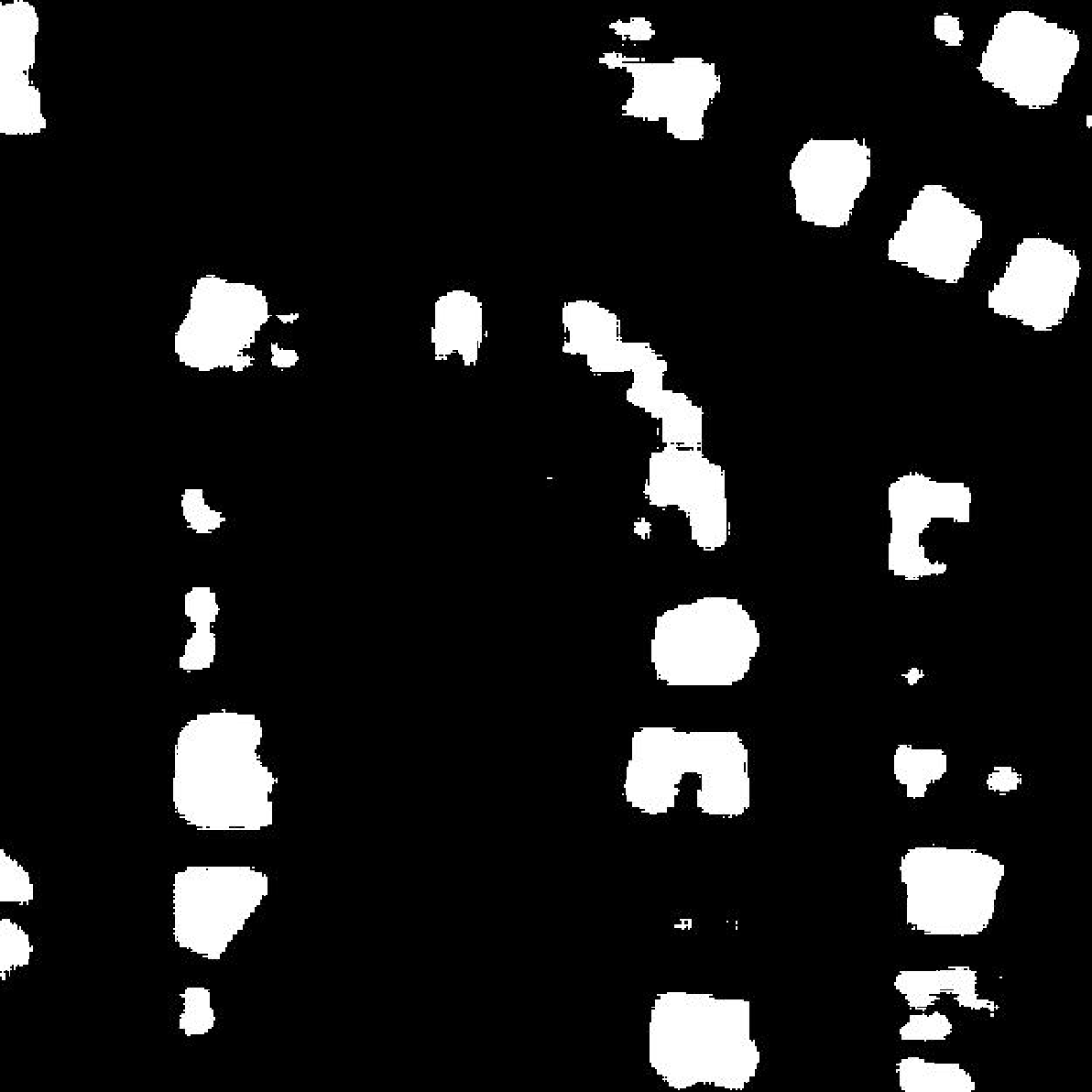}
\end{minipage}}
\subfigure[SRS + PDC]{
\begin{minipage}[b]{0.11\linewidth}
\includegraphics[width=1\linewidth]{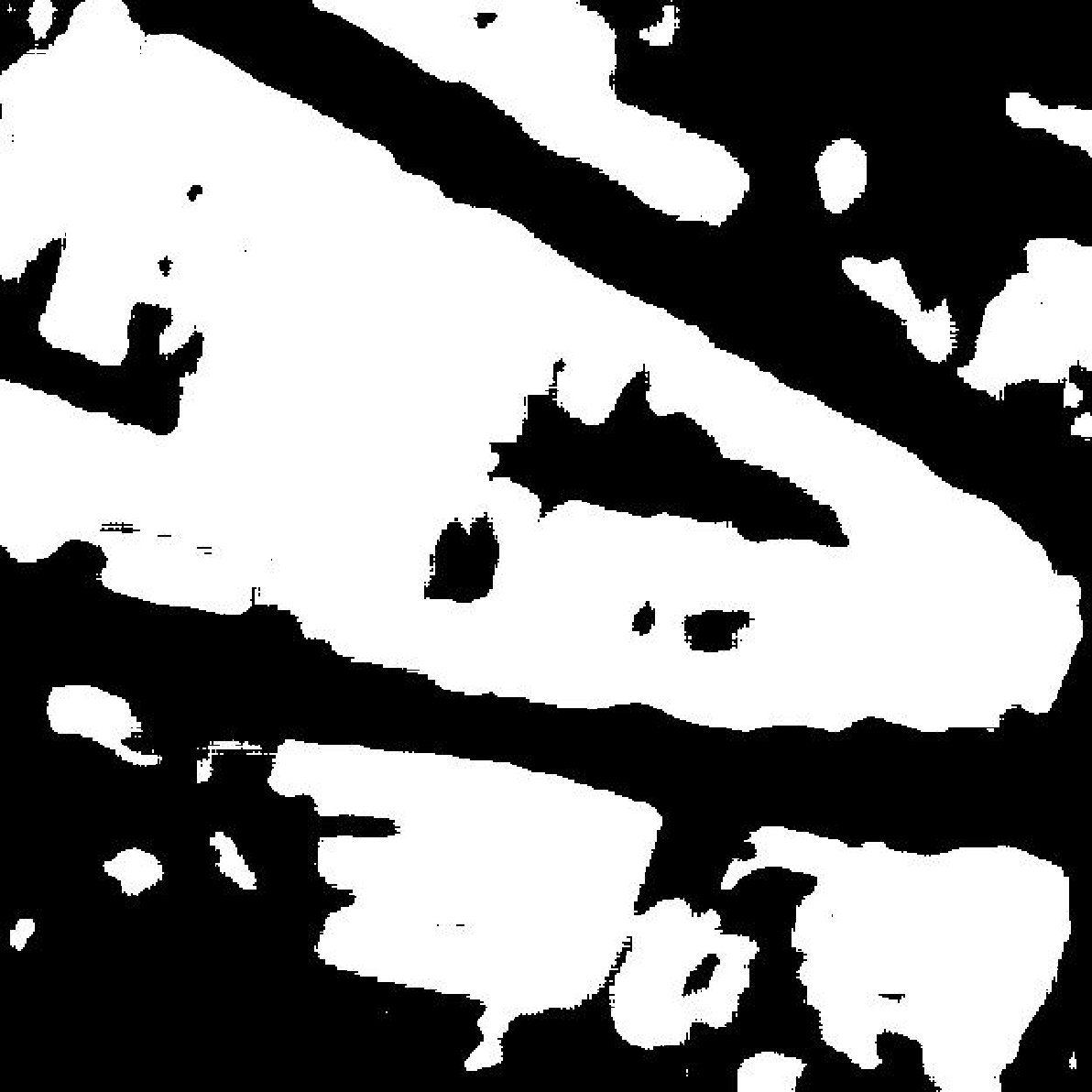}\vspace{4pt}
\includegraphics[width=1\linewidth]{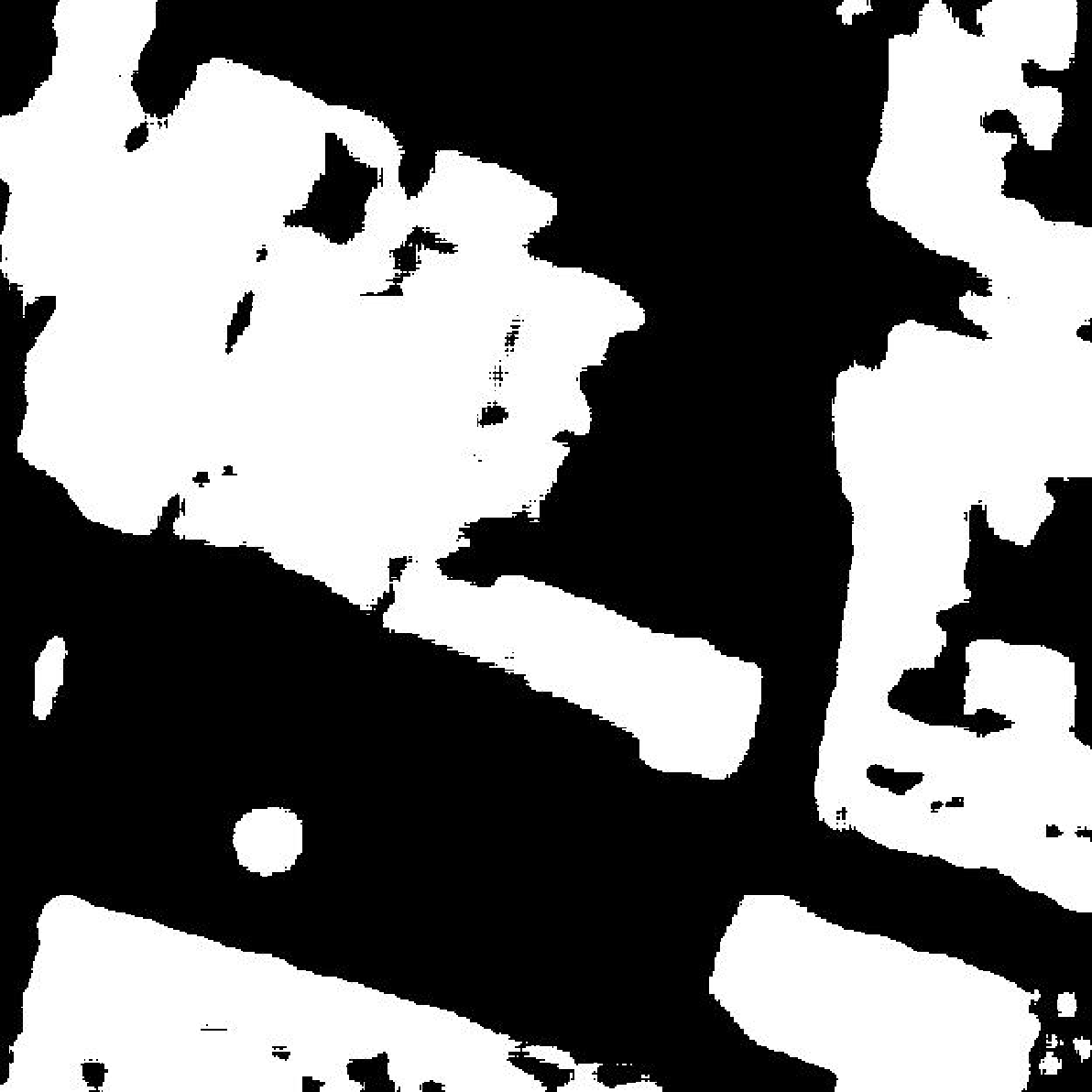}\vspace{4pt}
\includegraphics[width=1\linewidth]{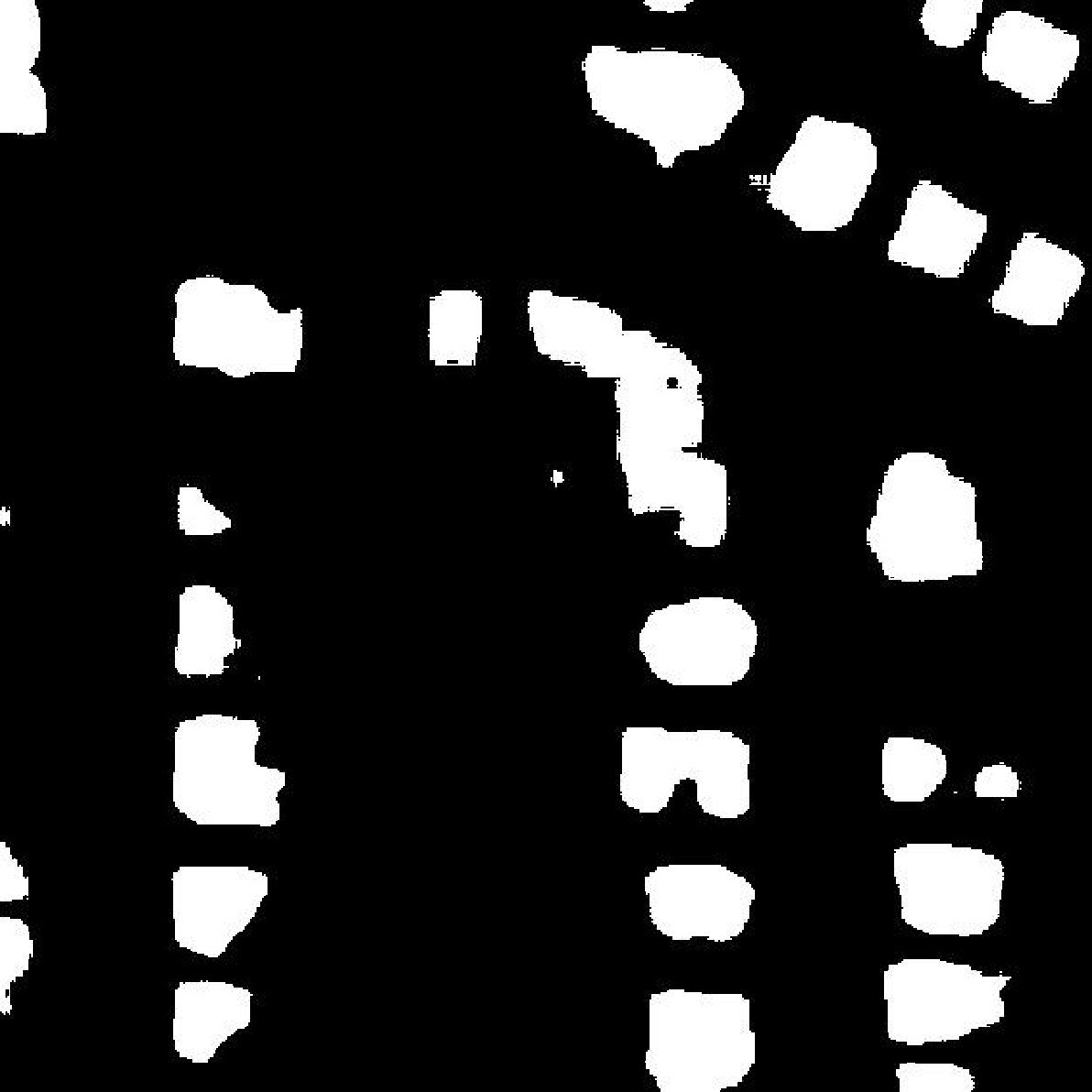}
\end{minipage}}
\subfigure[SRS + ODC]{
\begin{minipage}[b]{0.11\linewidth}
\includegraphics[width=1\linewidth]{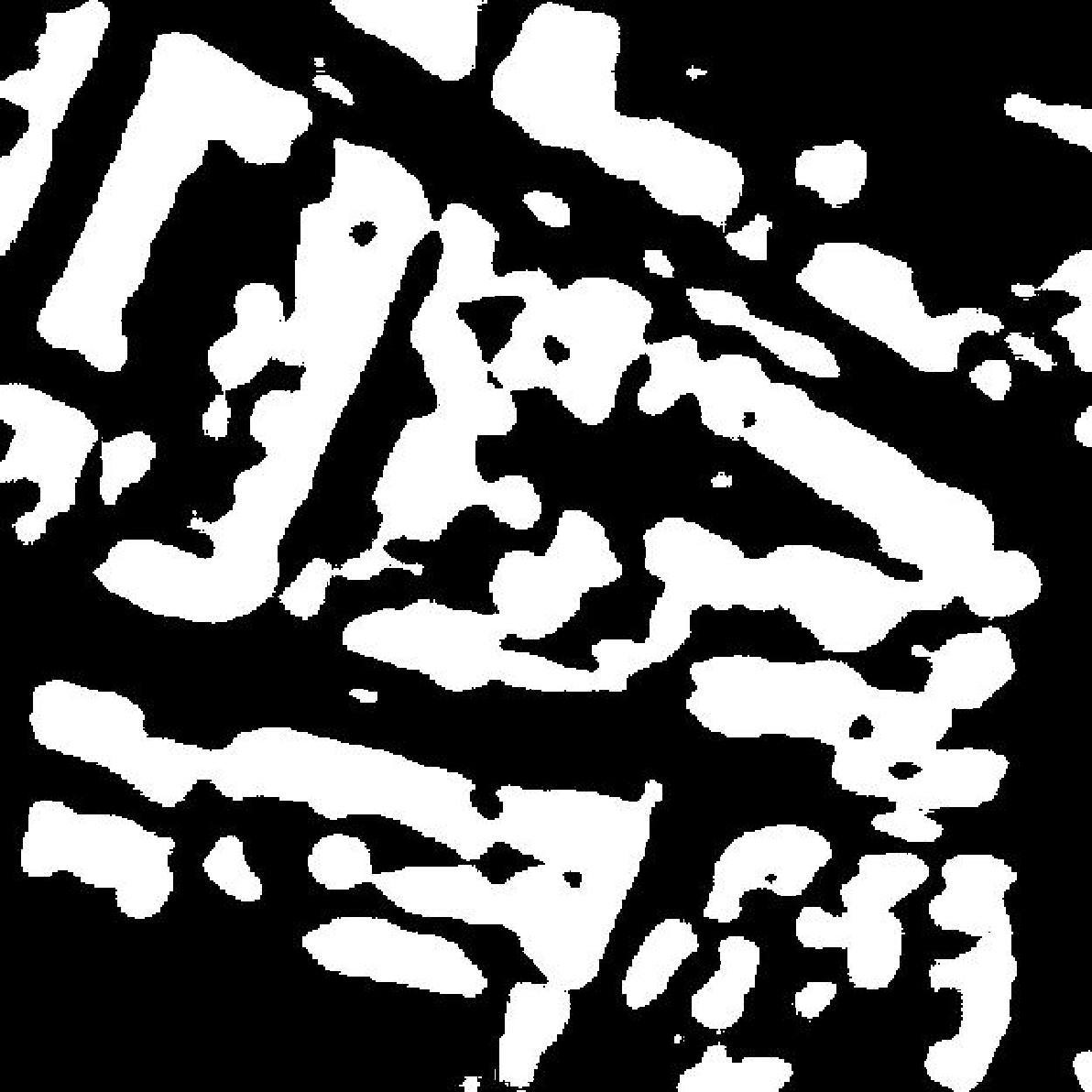}\vspace{4pt}
\includegraphics[width=1\linewidth]{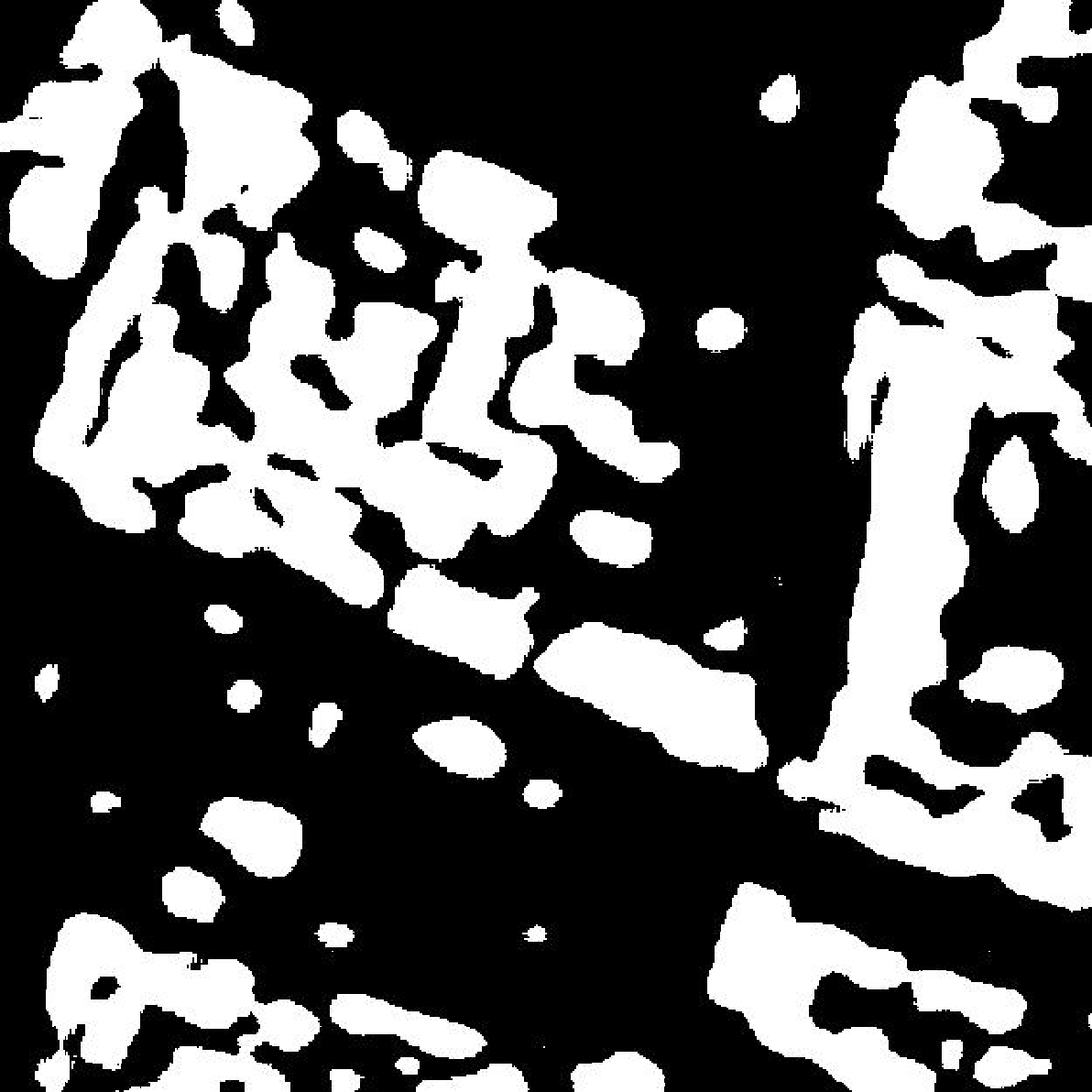}\vspace{4pt}
\includegraphics[width=1\linewidth]{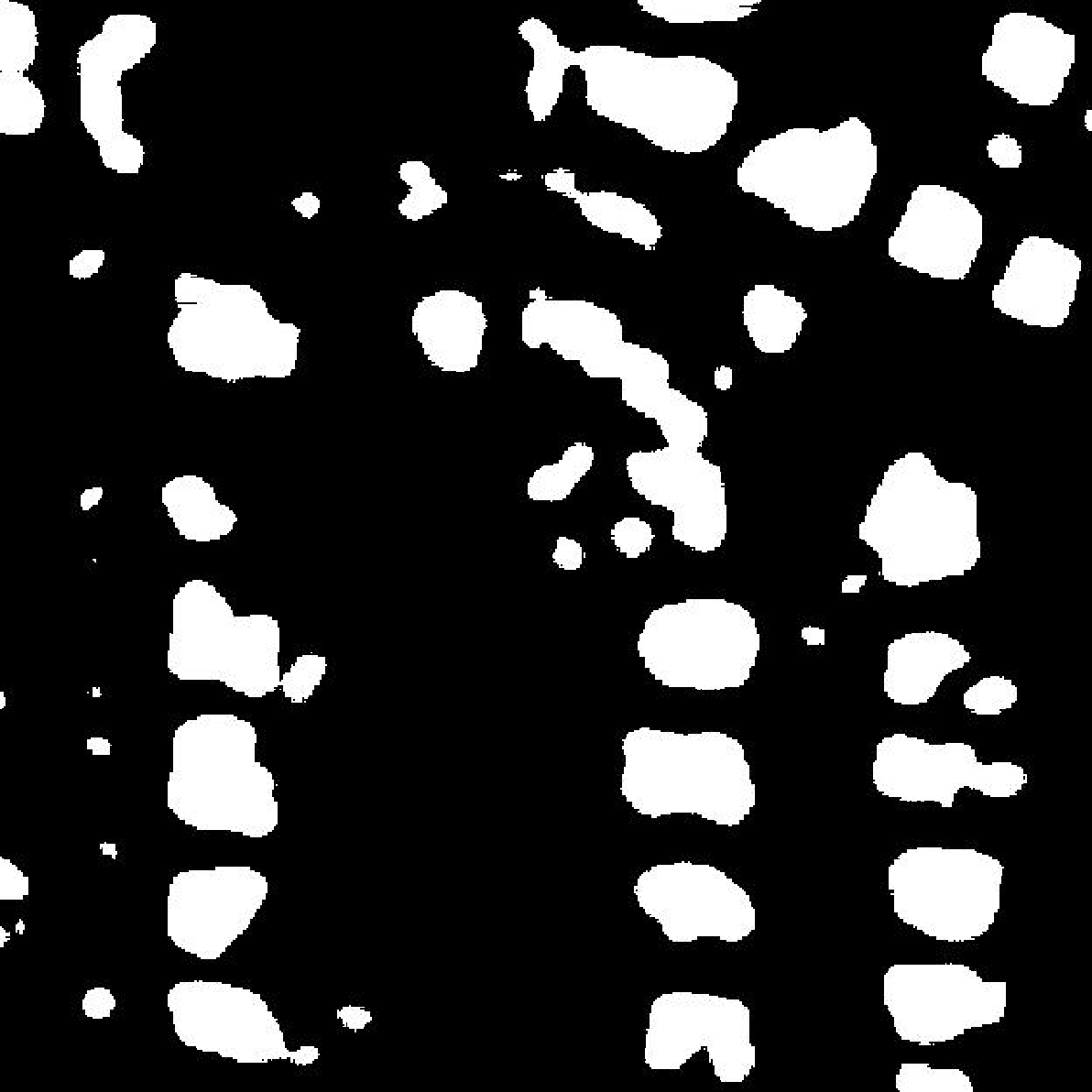}
\end{minipage}}
\subfigure[{\bfseries SRDA-Net}]{
\begin{minipage}[b]{0.11\linewidth}
\includegraphics[width=1\linewidth]{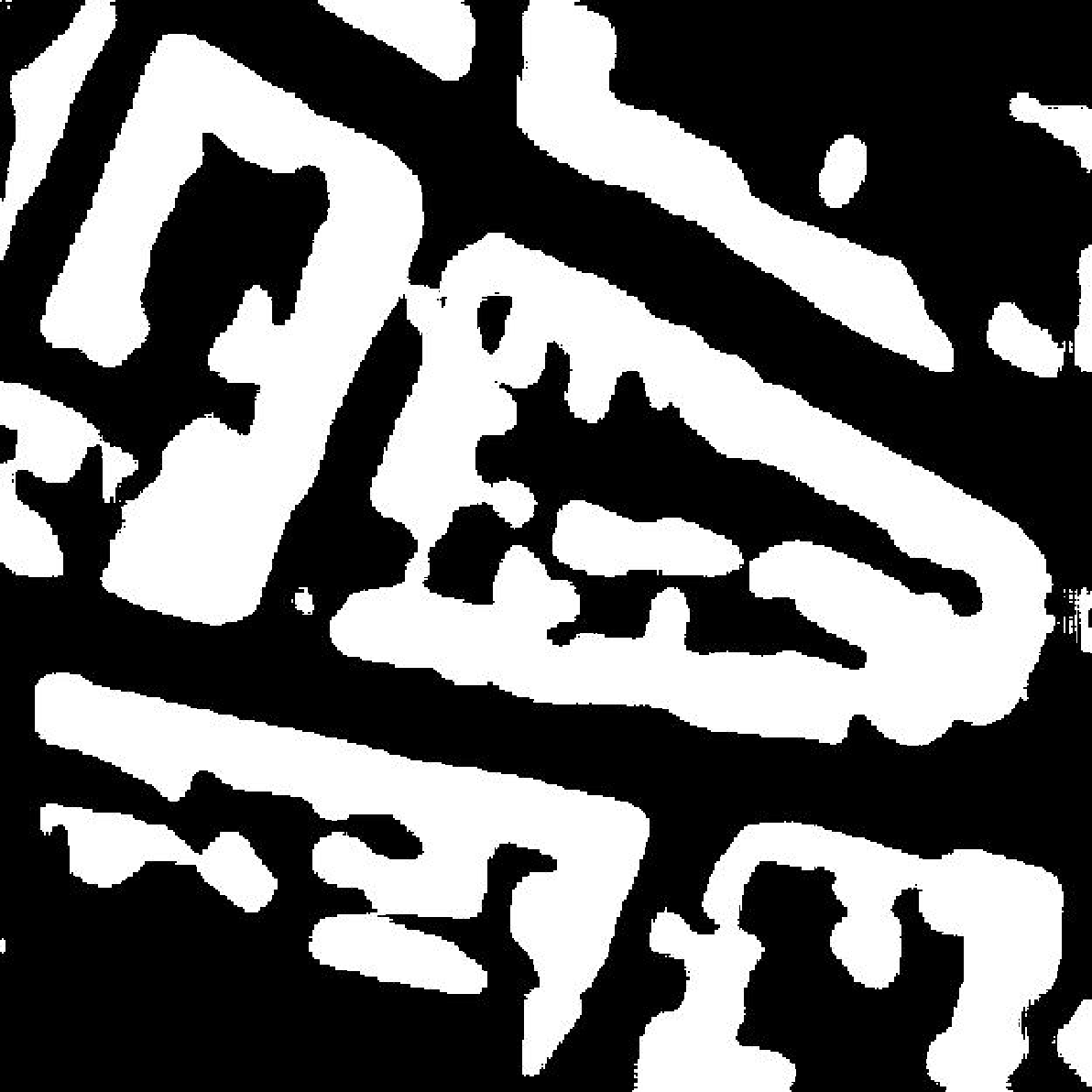}\vspace{4pt}
\includegraphics[width=1\linewidth]{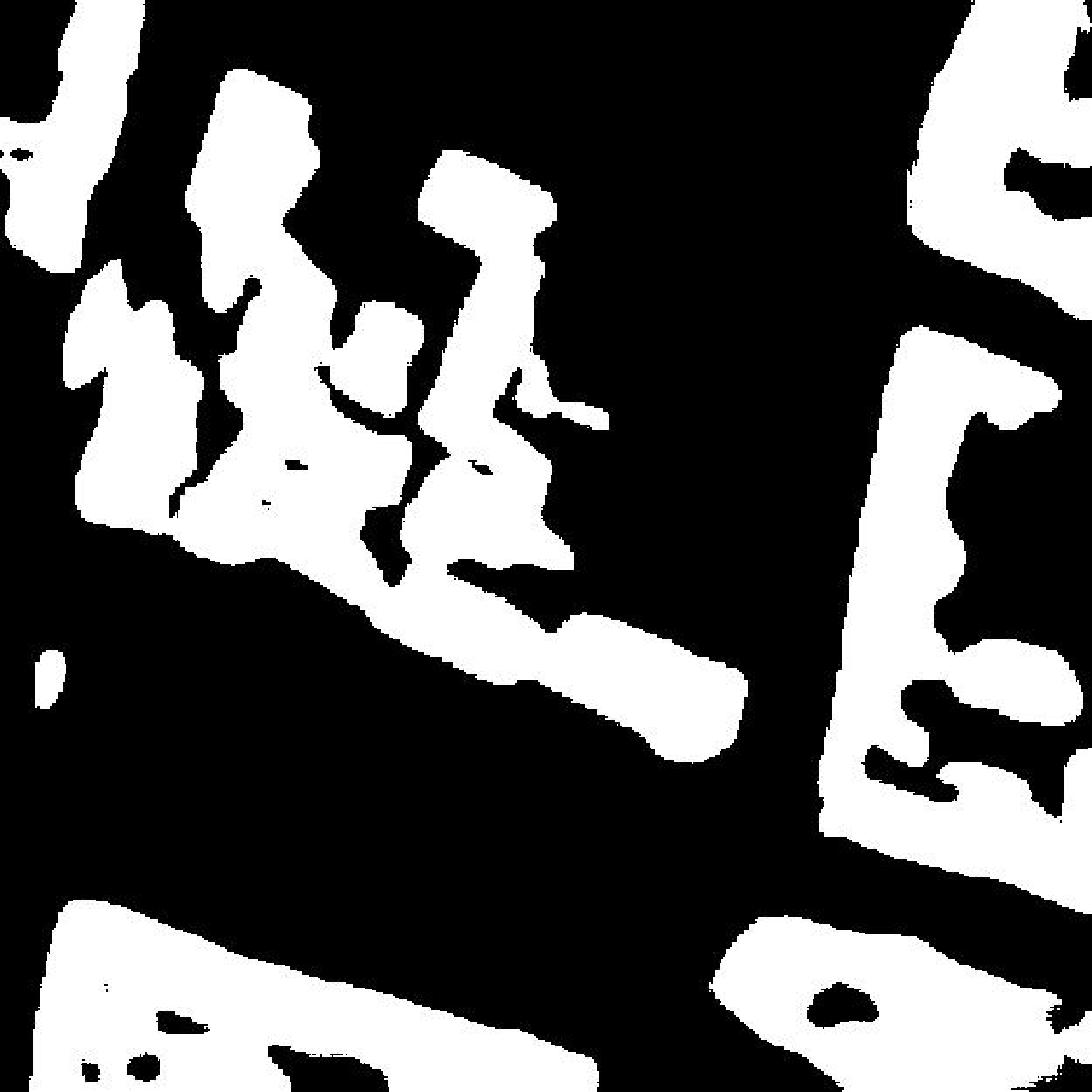}\vspace{4pt}
\includegraphics[width=1\linewidth]{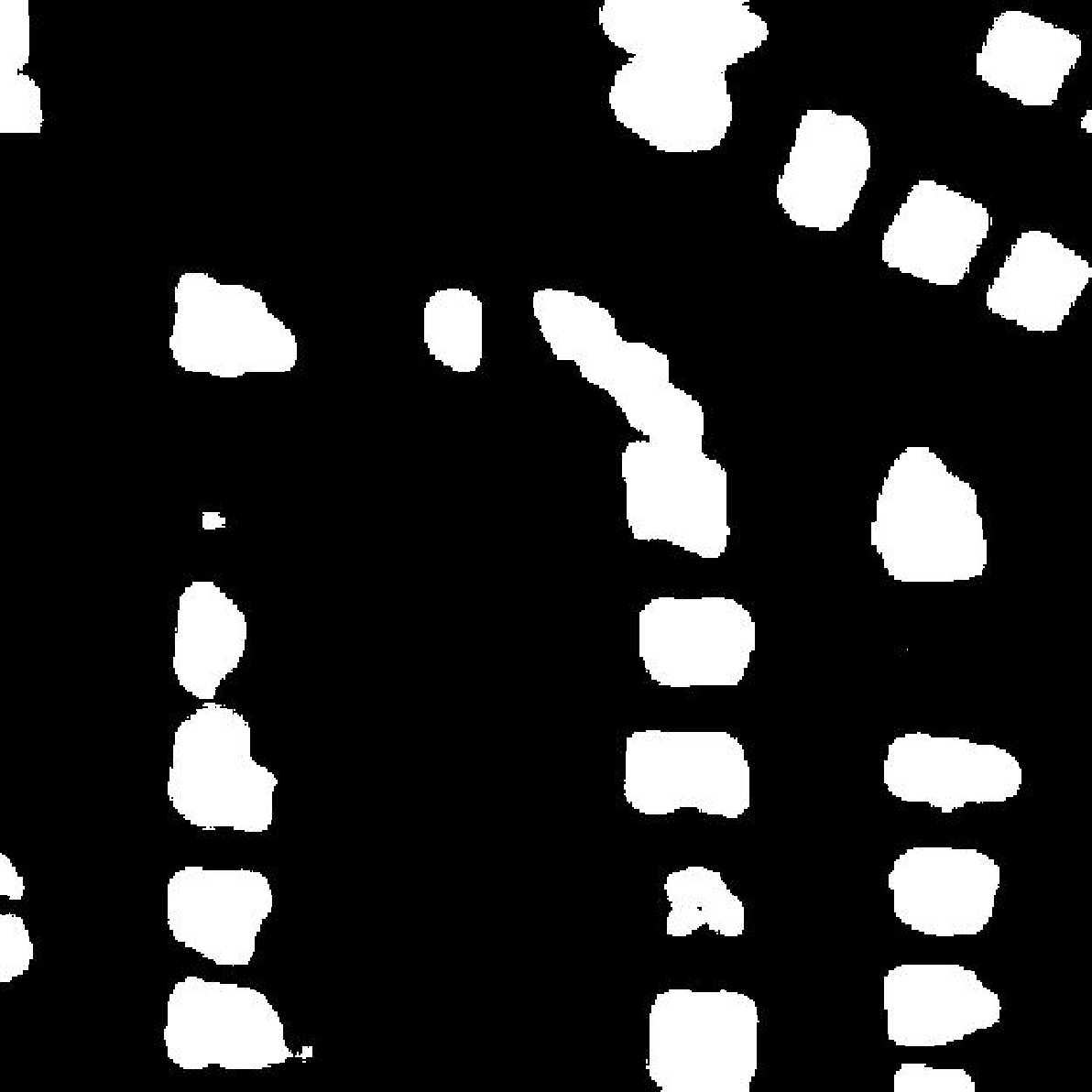}
\end{minipage}}
\subfigure[AdSgNt(48.5)]{
\begin{minipage}[b]{0.11\linewidth}
\includegraphics[width=1\linewidth]{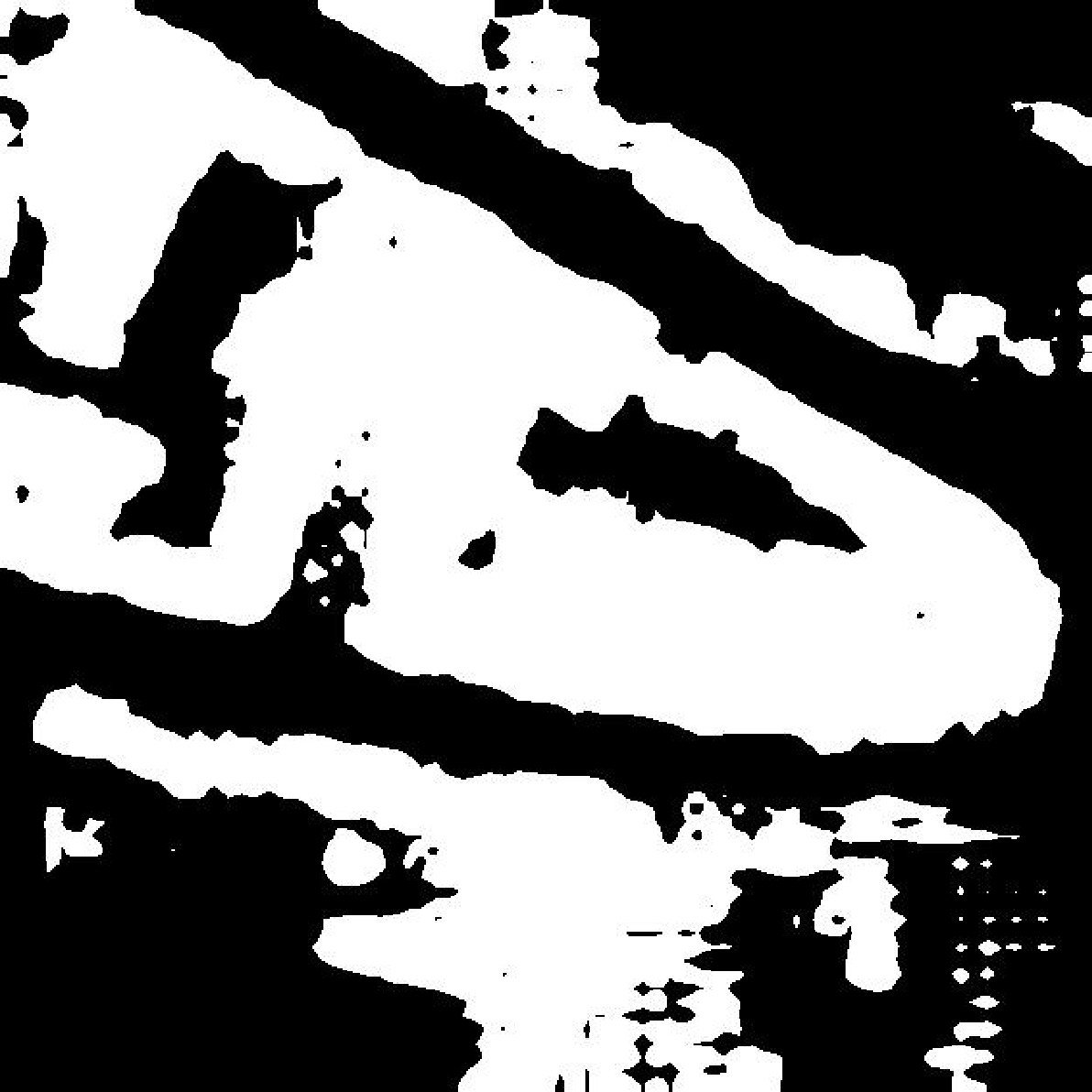}\vspace{4pt}
\includegraphics[width=1\linewidth]{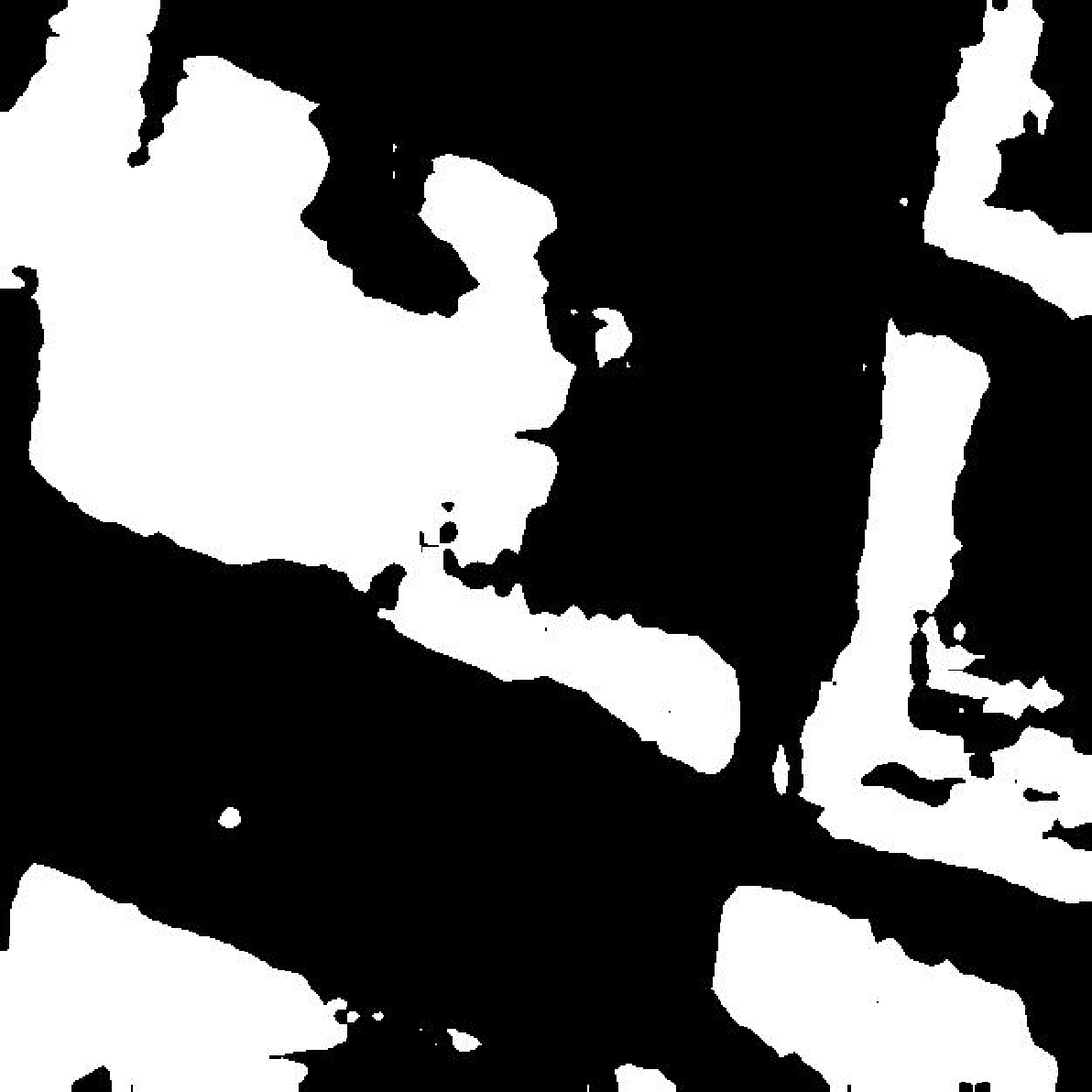}\vspace{4pt}
\includegraphics[width=1\linewidth]{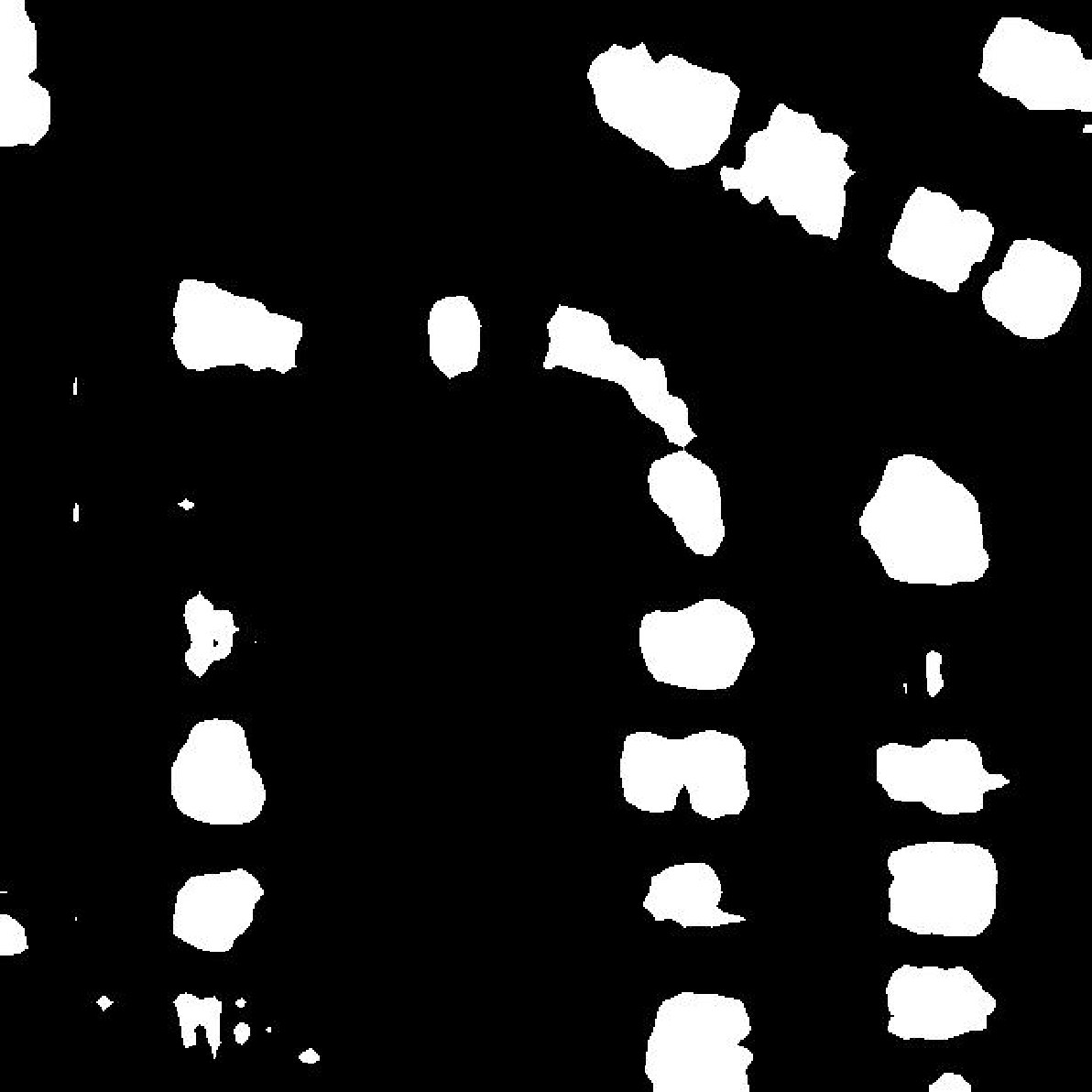}
\end{minipage}}
\subfigure[CFCAN(49.7)]{
\begin{minipage}[b]{0.11\linewidth}
\includegraphics[width=1\linewidth]{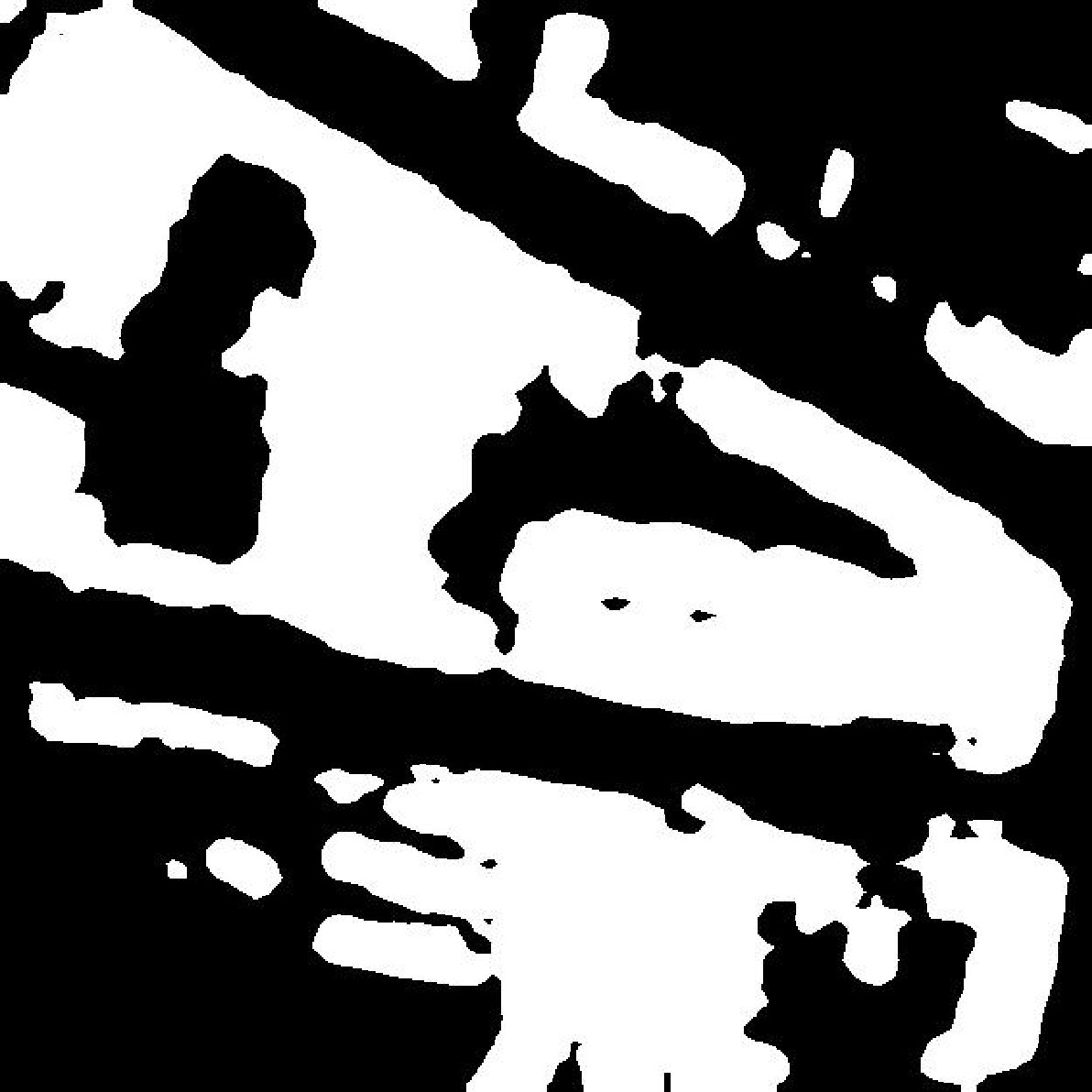}\vspace{4pt}
\includegraphics[width=1\linewidth]{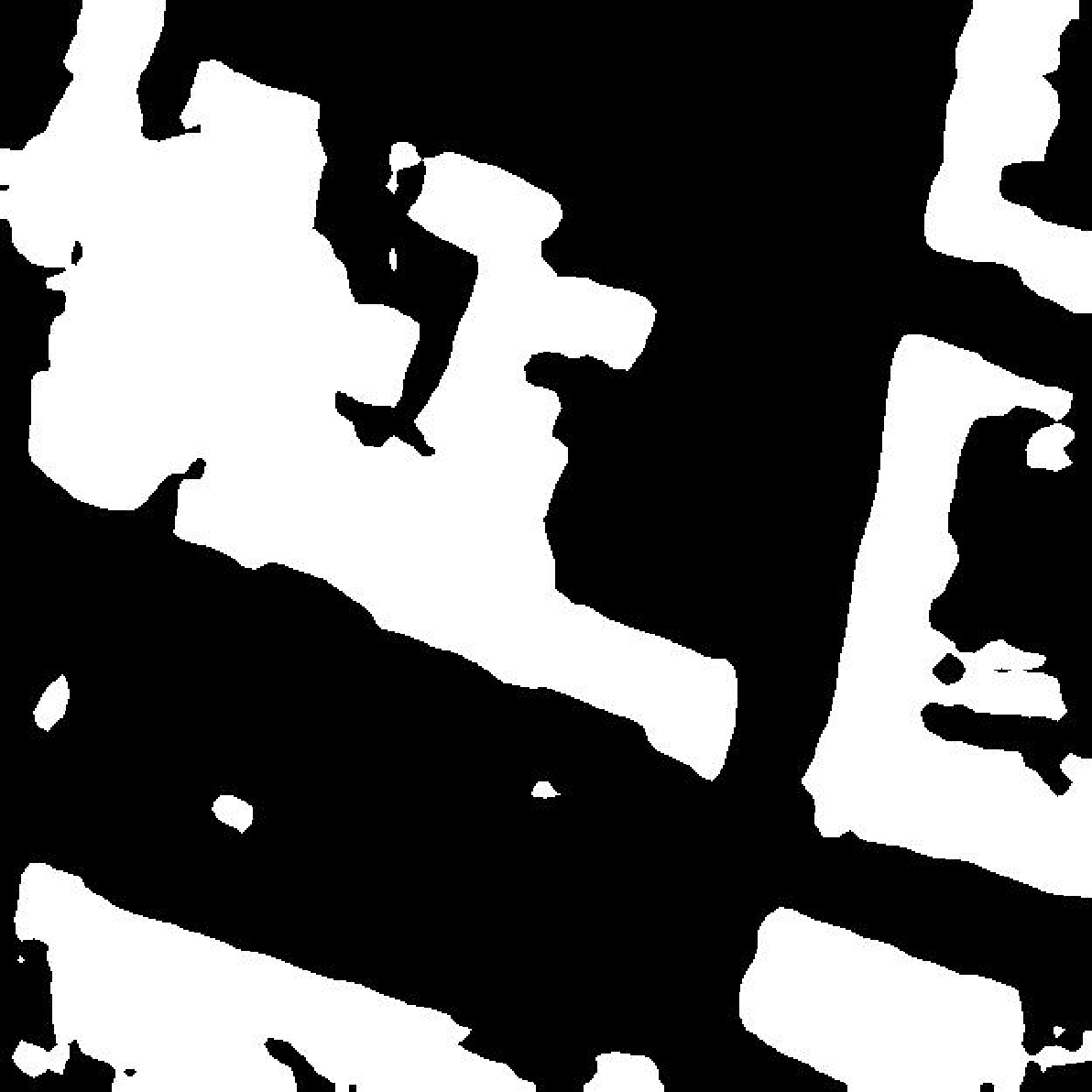}\vspace{4pt}
\includegraphics[width=1\linewidth]{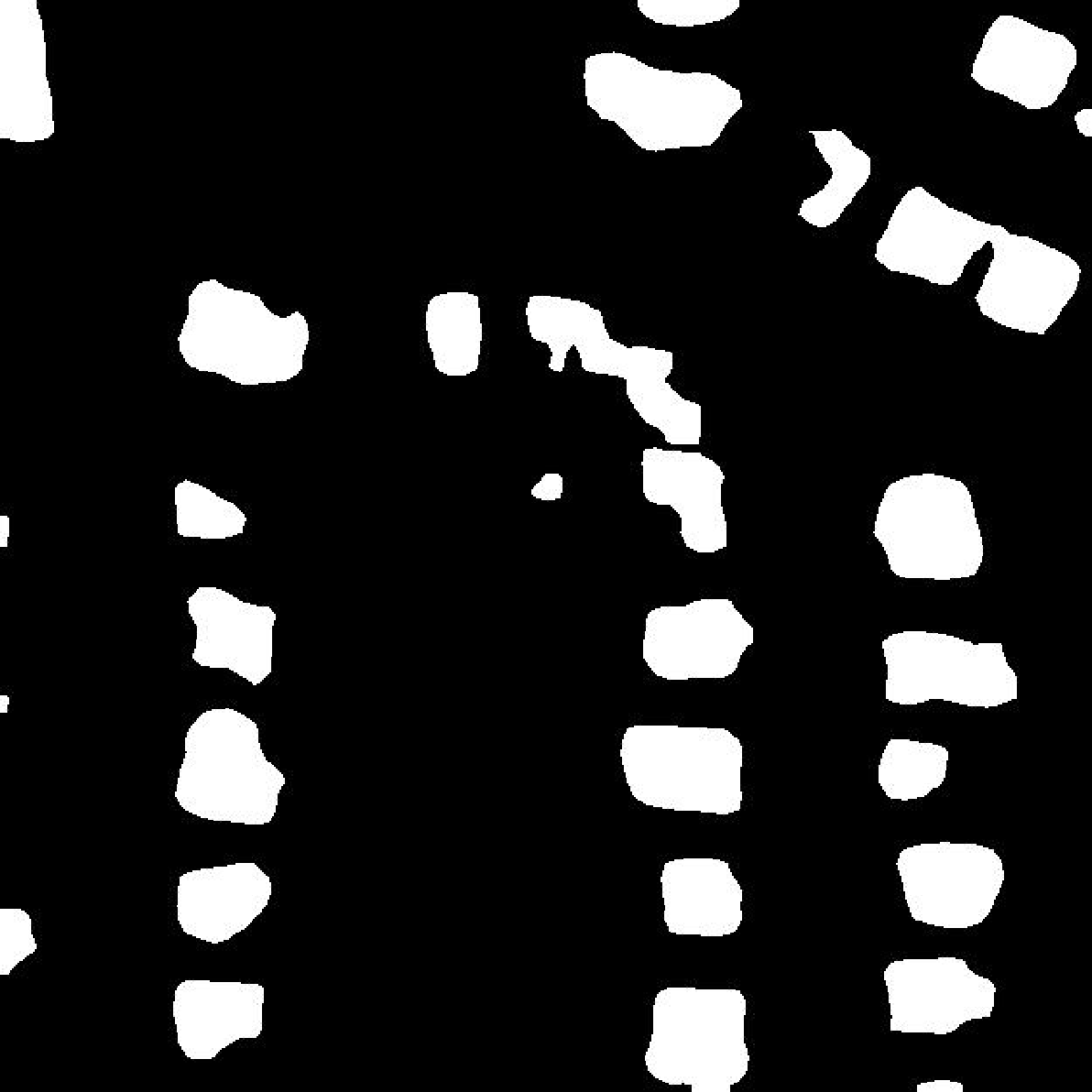}
\end{minipage}}
\caption{Qualitative results of the Inria val dataset (Source domain: Massachusetts Buildings)}
\label{fig:InriaResult}
\end{center}
\end{figure*}

%\begin{figure*}
%\vspace{-0.5cm}
%\centering \footnotesize Fig. 3. Qualitative results of the Inria val dataset (Source domain: Massachusetts Buildings)
%\end{figure*}

 \subsubsection{Vaih-Pots} we use the following two UDA datasets for multi-category semantic segmentation.
\begin{itemize}
 \item {\bfseries ISPRS Vaihingen 2D Semantic Labeling Challenge} contains 33 images of different sizes at 9cm spatial resolution, taken over the city of Vaihingen (Germany). Each image consists of a true orthophoto extracted from a larger orthophoto mosaic. There are 6 labeled categories: impervious surface, building, low vegetation, tree, car, Clutter/background. This dataset is considered to be a source domain.

 \item {\bfseries ISPRS Potsdam 2D Semantic Labeling Challenge dataset} is comprised of 38 ortho-rectified aerial IRRGB images with size of $6000\times6000$ at 5cm spatial resolution, taken over the city of Potsdam in Germany. The ground truth is provided for 24 tiles alike Vaihingen dataset. We randomly choose 12 images as train set, and other 12 images as test set.
\end{itemize}
%While Vaihingen is a relatively small village with many detached buildings and small multi story buildings, Potsdam shows a typical historic city with large building blocks, narrow streets and dense settlement structure. Therefore, there are domain discrepancies between the two datasets besides the different resolution.

Note that the resolution gap of Mass-Inria (around 3.333 times) is greater than Vaih-Pots (around 2.0 times).

\subsection{Evaluation Metric and Implementation Details}

\subsubsection{Evaluation Metric}  

The intersection-over-union (IoU) is adopted as the main evaluation metric. And it is defined as:
\begin{equation}
\operatorname{IoU}\left(P_{m}, P_{g t}\right)=\left|\frac{P_{m} \cap P_{g t}}{P_{m} \cup P_{g t}}\right|
\end{equation}
where $P_{m}$ is the prediction and $P_{g t}$ is the ground truth. Mean IoU (mIoU) is used to evaluate model performance on all classes.

\begin{table*}[!ht]
\small    % 控制字体
\renewcommand{\arraystretch}{1.3} %控制行距
\begin{center}
\caption{The comparison results of domain adaptation from Vaih to Pots val dataset}
\label{tab:VaihToPots}
\setlength{\tabcolsep}{3mm}{%调整宽度
\begin{tabular}{lllllllllll}
\toprule[1.2pt] %加粗的横线
\noalign{\smallskip} %
Methods \% & \rotatebox{90}{BaseNet}& \rotatebox{90}{Source}& \rotatebox{90} {Target}&  \rotatebox{90} {Impervious} & \rotatebox{90} {Building} & \rotatebox{90} {vegetation} & \rotatebox{90} {Tree} & \rotatebox{90} {Car} & \rotatebox{90} {Clutter} & \rotatebox{90}{{\bfseries mIoU}}    \\
\noalign{\smallskip}
\hline
\noalign{\smallskip}
NoAdapt \cite{Tsai2018Learning}& Resnet-101 \cite{He2016Deep}& Vaih &  Pots  &51.8 &45.5&46.2 &11.8 &35.3&18.5&34.9 \\
AdaptSegNet \cite{Tsai2018Learning}&Resnet-101 \cite{He2016Deep}&Vaih& $\downarrow$ Pots&59.4 & 54.2&47.0&26.3&52.2&{\bfseries 32.2}&45.2 \\
AdaptSegNet \cite{Tsai2018Learning}&Resnet-101 \cite{He2016Deep}& $\uparrow$ Vaih&Pots&55.1&55.6&43.0&31.5&60.6 &1.6&41.2\\
\noalign{\smallskip}
\hline
\noalign{\smallskip}
NoAdapt \cite{Zhang2018Fully}&Resnet-101 \cite{He2016Deep}&Vaih&Pots&51.8 &45.5&46.2 &11.8 &35.3&18.5&34.9 \\
CycleGan-FCAN \cite{Zhang2018Fully}&Resnet-101 \cite{He2016Deep}&Vaih& $\downarrow$ Pots&50.1&42.5&33.1&31.6&44.1&22.6&37.3  \\
CycleGan-FCAN \cite{Zhang2018Fully}&Resnet-101 \cite{He2016Deep}&$ \uparrow$ Vaih&Pots&47.9&51.2&43.0&41.7&61.1&23.8&44.8  \\
\noalign{\smallskip}
\hline
\noalign{\smallskip}
NoAdapt & ResidualASPP \cite{Wang2019Learning}  & Vaih & Pots & 29.1 &36.3&37.6 &19.3 &2.8&23.4&24.7        \\
SRS & ResidualASPP \cite{Wang2019Learning}  & Vaih & Pots & 26.5 &32.0&35.2 &17.3 &32.0&17.5&26.7        \\
SRS + PDC    & ResidualASPP \cite{Wang2019Learning}   & Vaih & Pots & 58.3 &51.1 &51.8& 27.9&62.5&20.5&45.4	         \\
SRS + ODC  & ResidualASPP \cite{Wang2019Learning}     & Vaih & Pots & 51.2&21.7&17.9&12.3&54.2&13.0&28.4 \\
Full (SRDA-Net)& ResidualASPP \cite{Wang2019Learning}& Vaih & Pots &{\bfseries 60.2}&{\bfseries 61.0}&{\bfseries 51.8}&{\bfseries 36.8}&{\bfseries 63.4}&18.3&{\bfseries 48.6}\\
\noalign{\smallskip}
\bottomrule[1.2pt] %加粗的横线
\end{tabular}}
\end{center}
\end{table*}

%\begin{figure*}[!ht]
 %\centering
 %\includegraphics[width=18cm, height=1cm]{color/colormap.eps}
%\end{figure*}

\begin{figure*}
\begin{center}
\includegraphics[width=17.8cm, height=0.8cm]{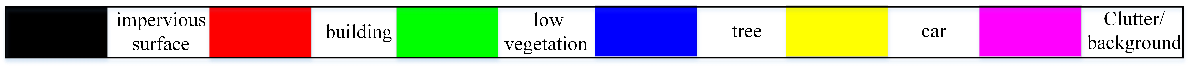}\vspace{-1.5pt}
\subfigure[Input image]{
\begin{minipage}[b]{0.11\linewidth}
\includegraphics[width=1\linewidth]{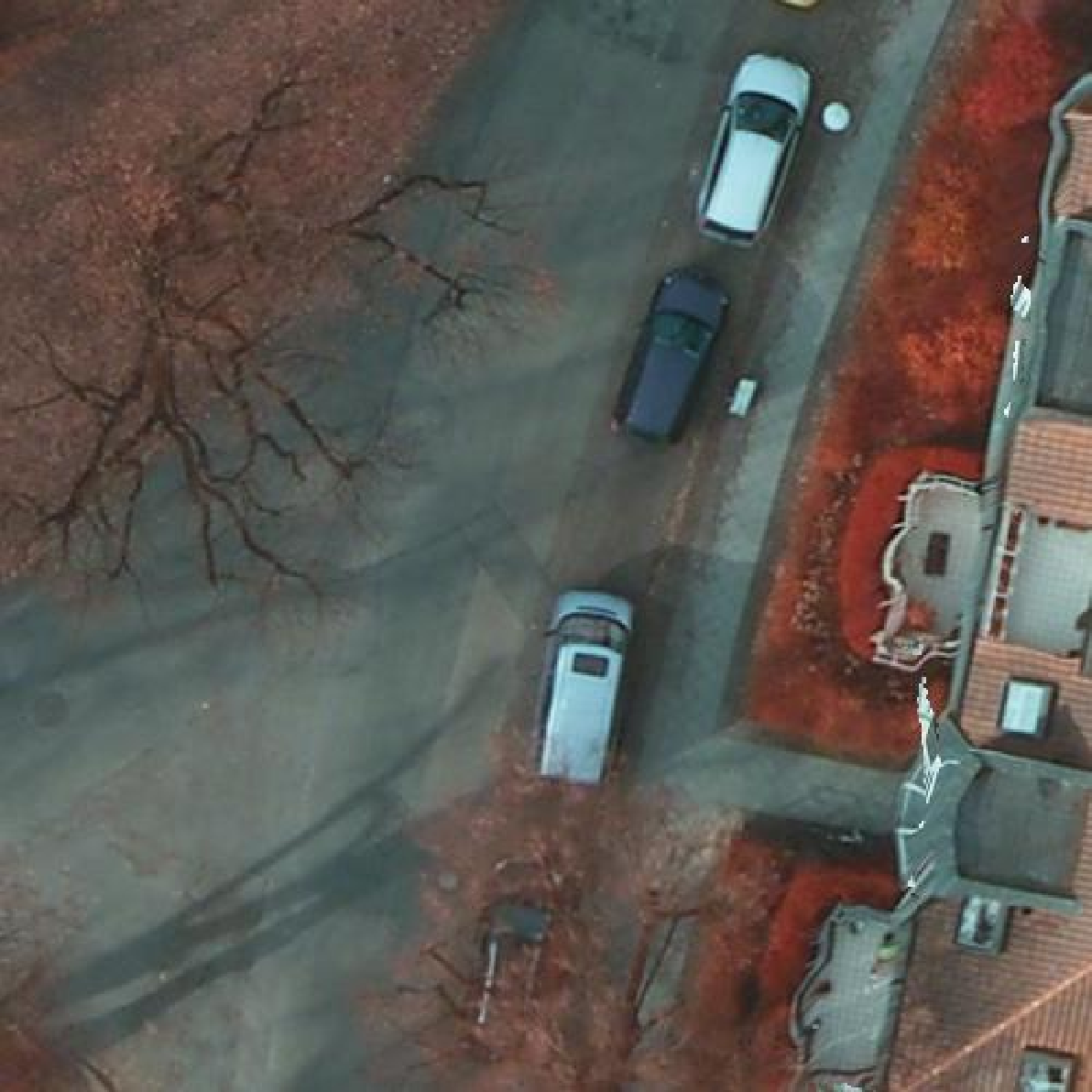}\vspace{4pt}
\includegraphics[width=1\linewidth]{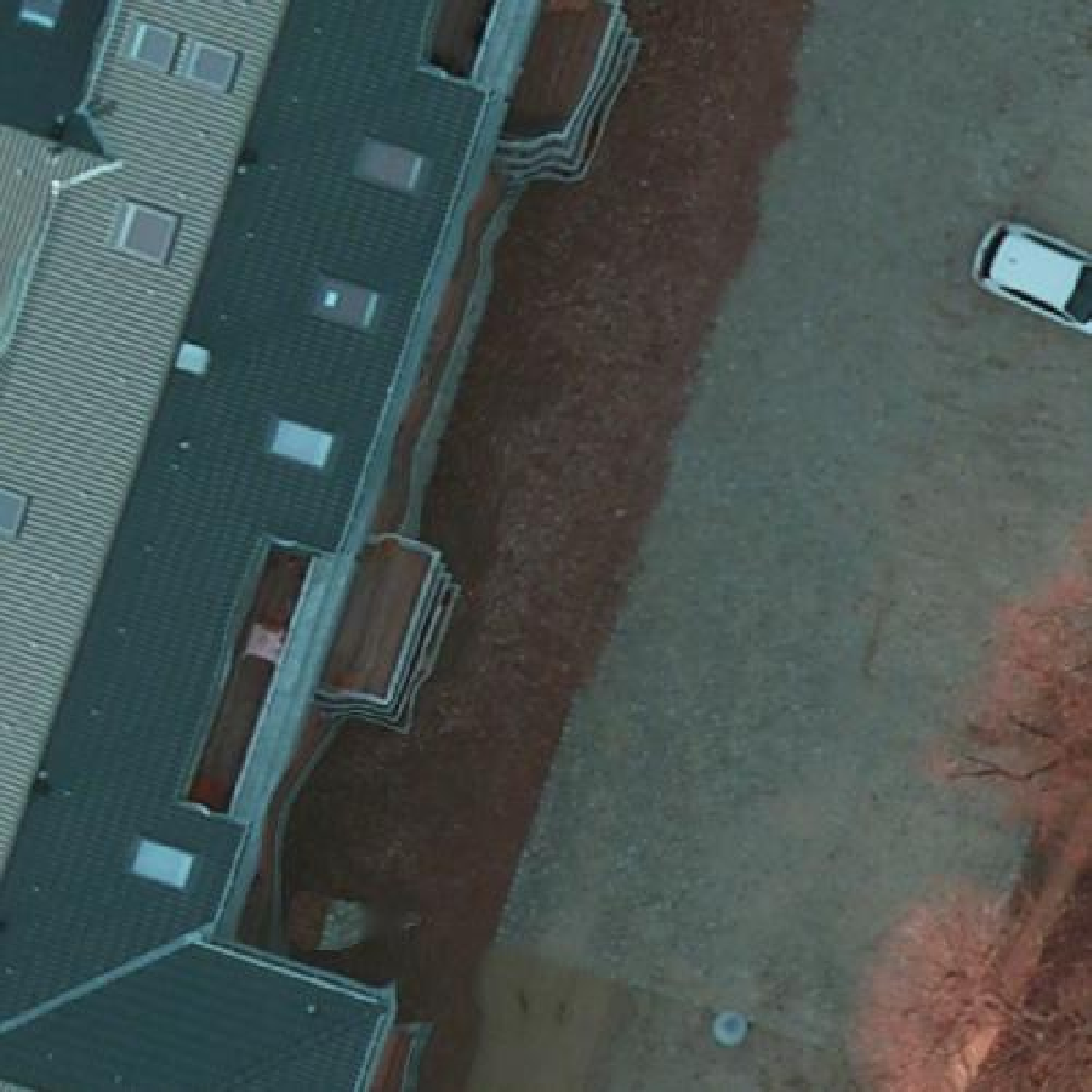}\vspace{4pt}
\includegraphics[width=1\linewidth]{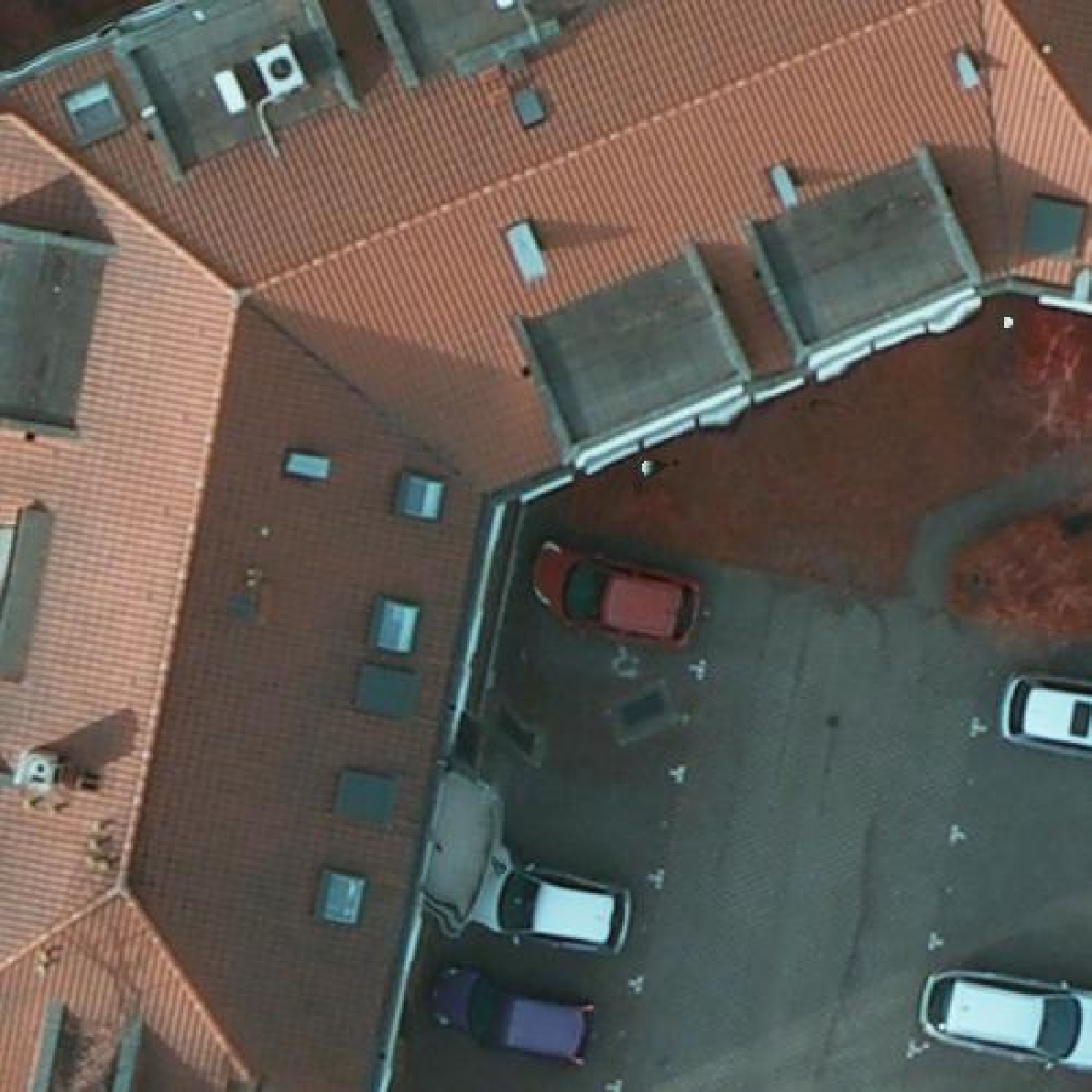}
\end{minipage}}
\subfigure[Ground Truth]{
\begin{minipage}[b]{0.11\linewidth}
\includegraphics[width=1\linewidth]{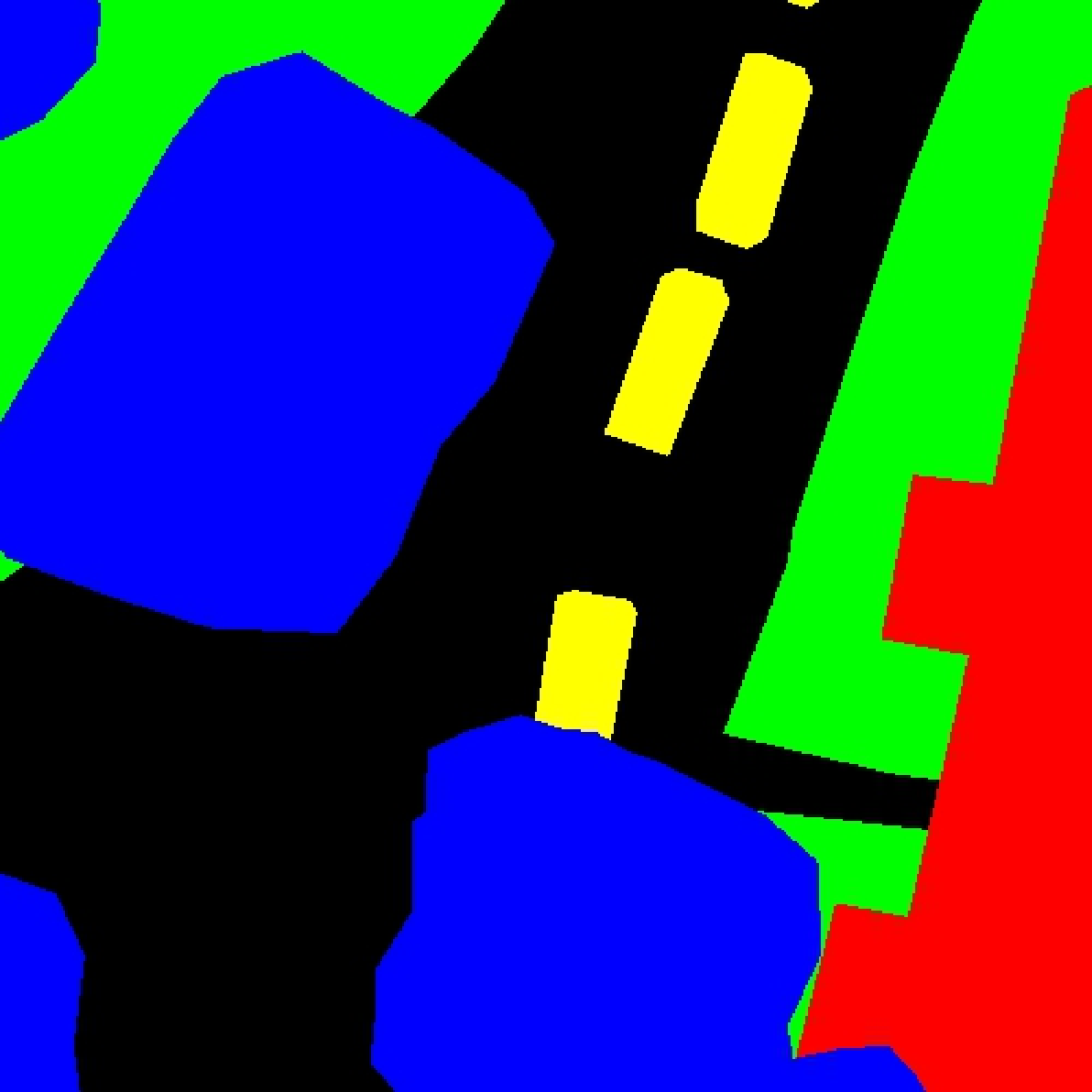}\vspace{4pt}
\includegraphics[width=1\linewidth]{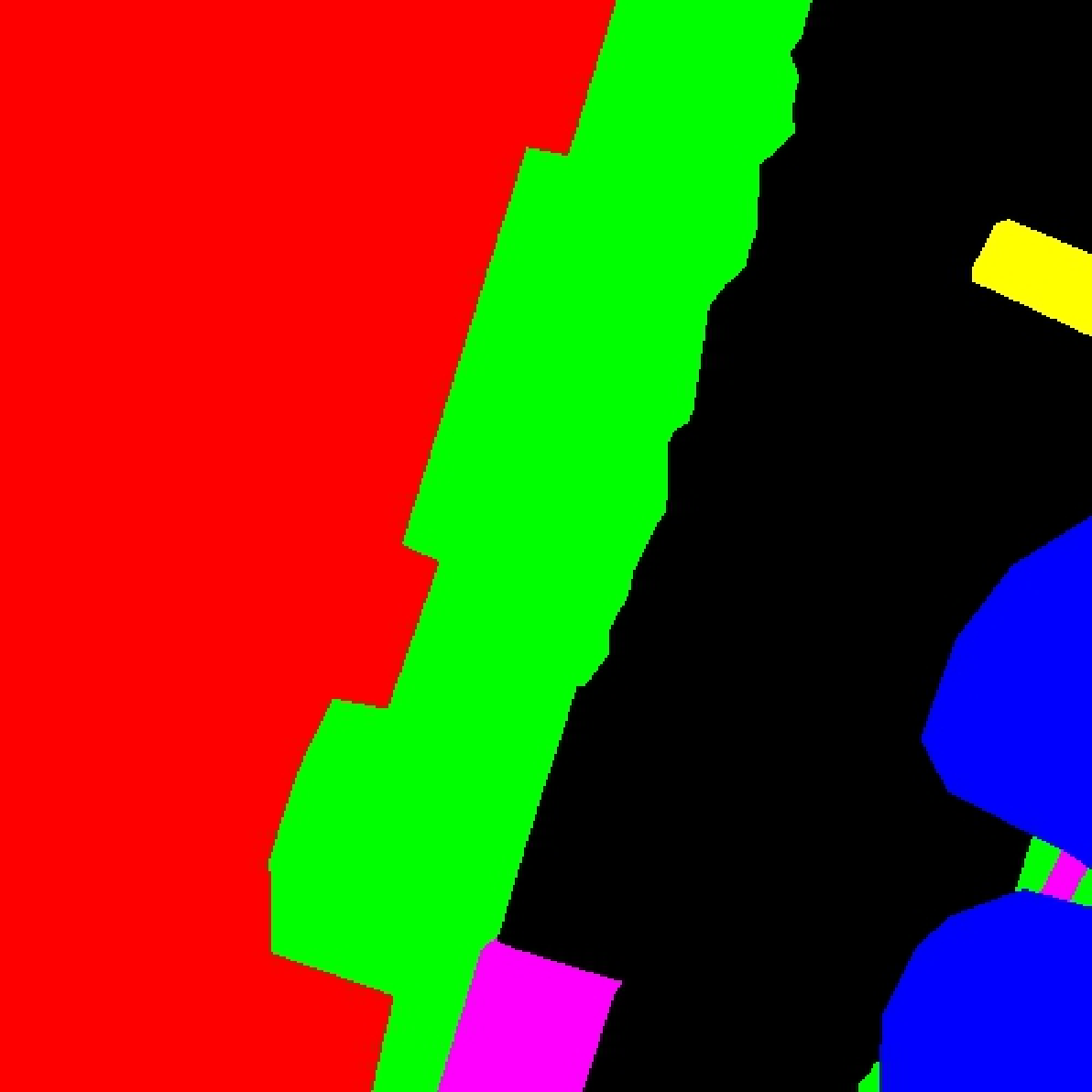}\vspace{4pt}
\includegraphics[width=1\linewidth]{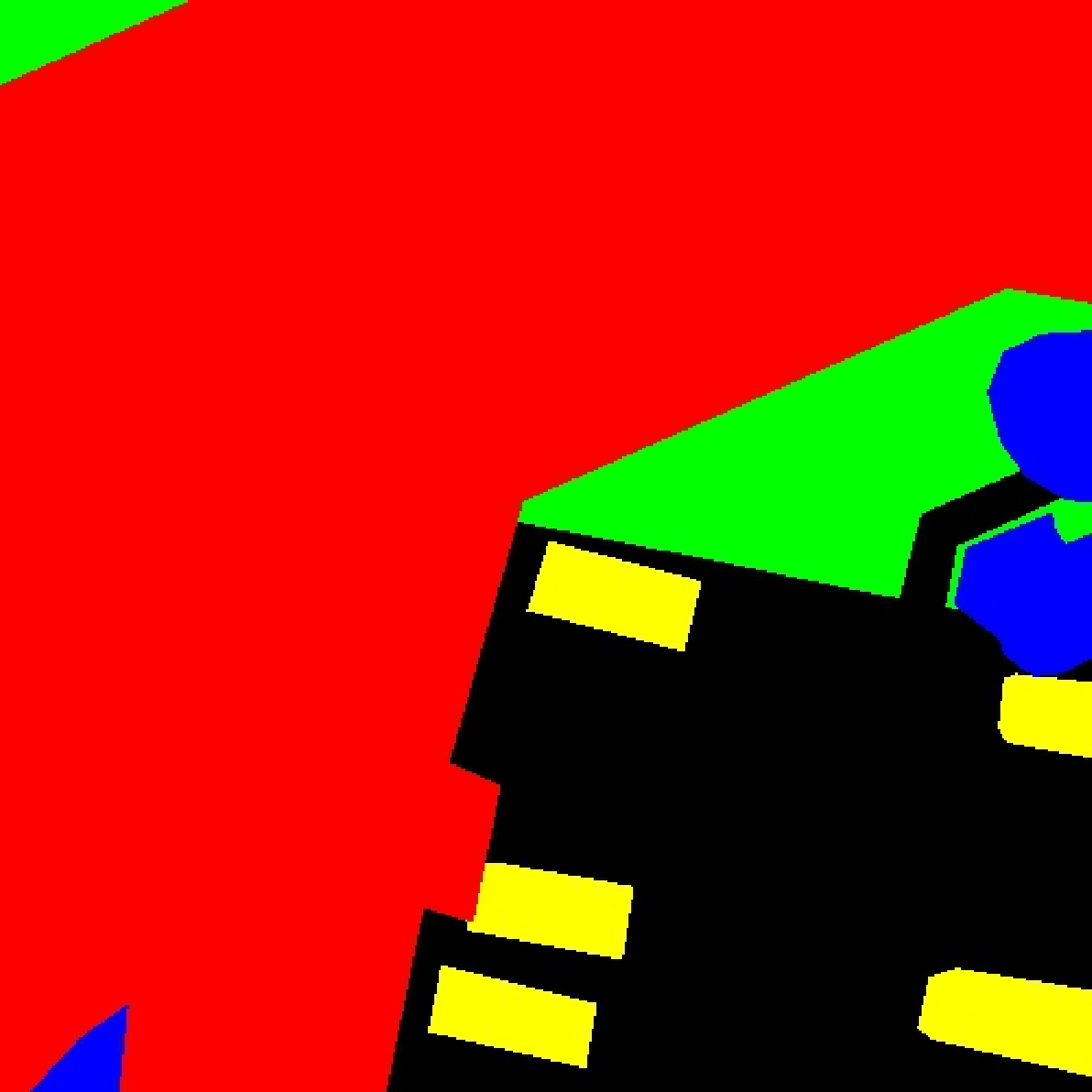}
\end{minipage}}
\subfigure[SRS]{
\begin{minipage}[b]{0.11\linewidth}
\includegraphics[width=1\linewidth]{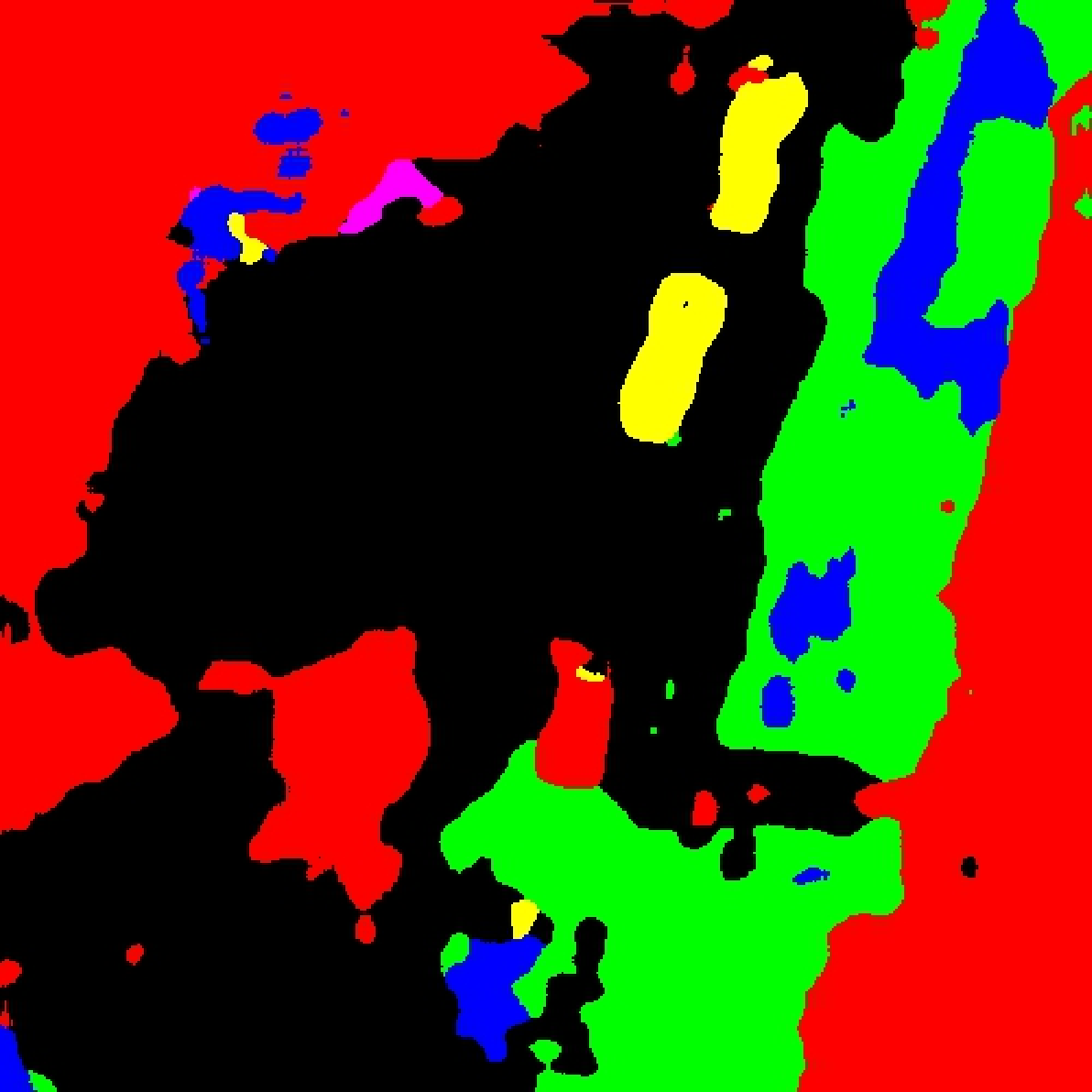}\vspace{4pt}
\includegraphics[width=1\linewidth]{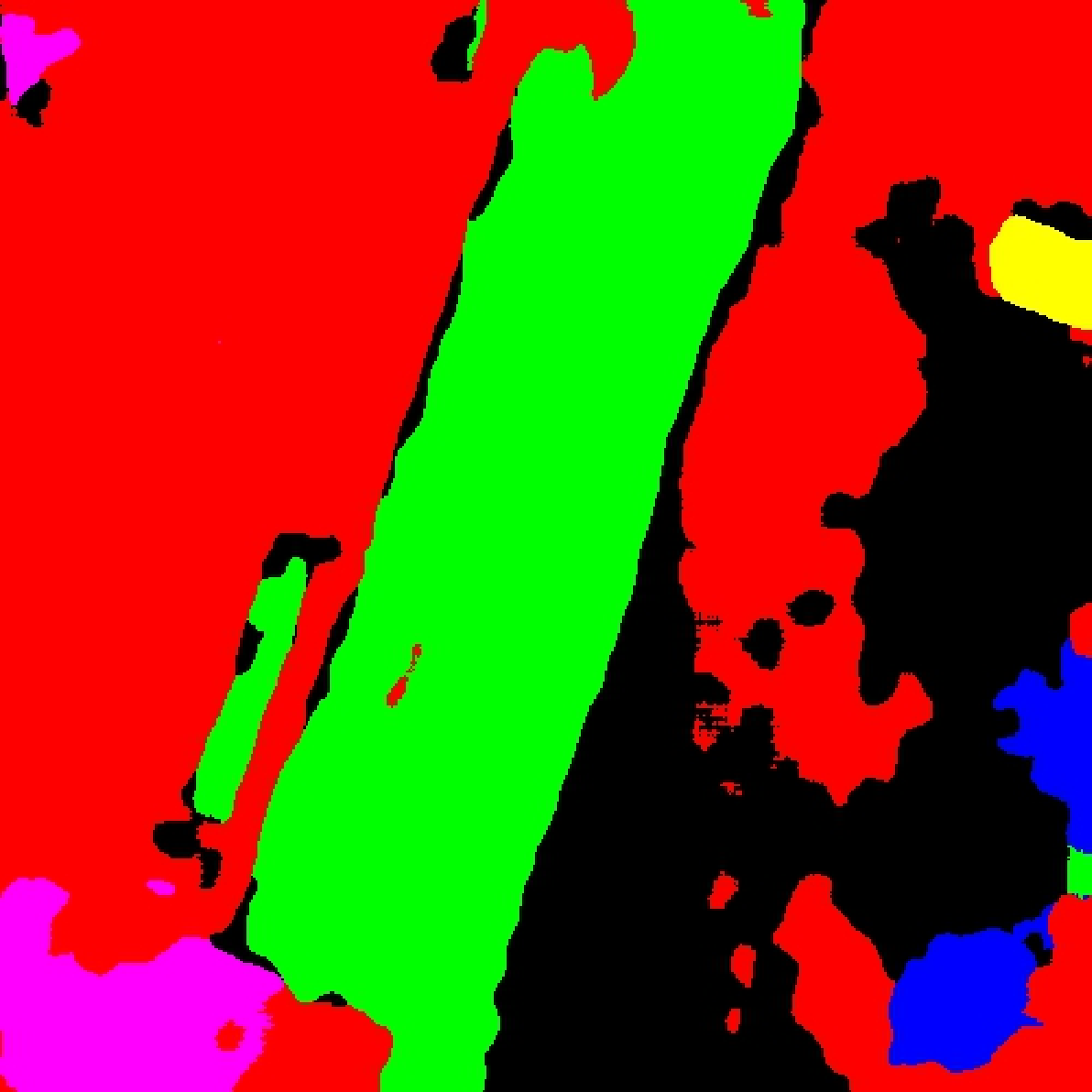}\vspace{4pt}
\includegraphics[width=1\linewidth]{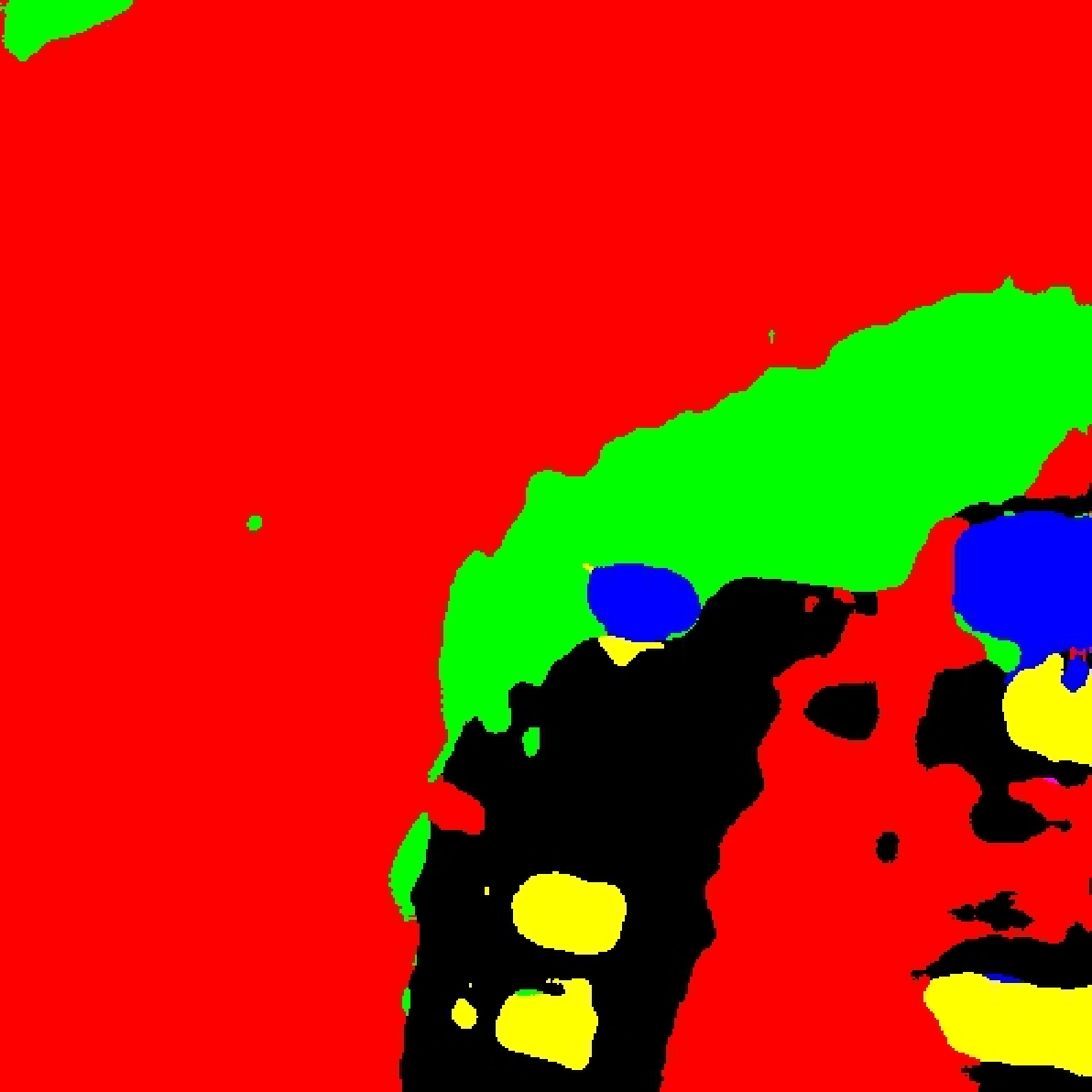}
\end{minipage}}
\subfigure[SRS + PDC]{
\begin{minipage}[b]{0.11\linewidth}
\includegraphics[width=1\linewidth]{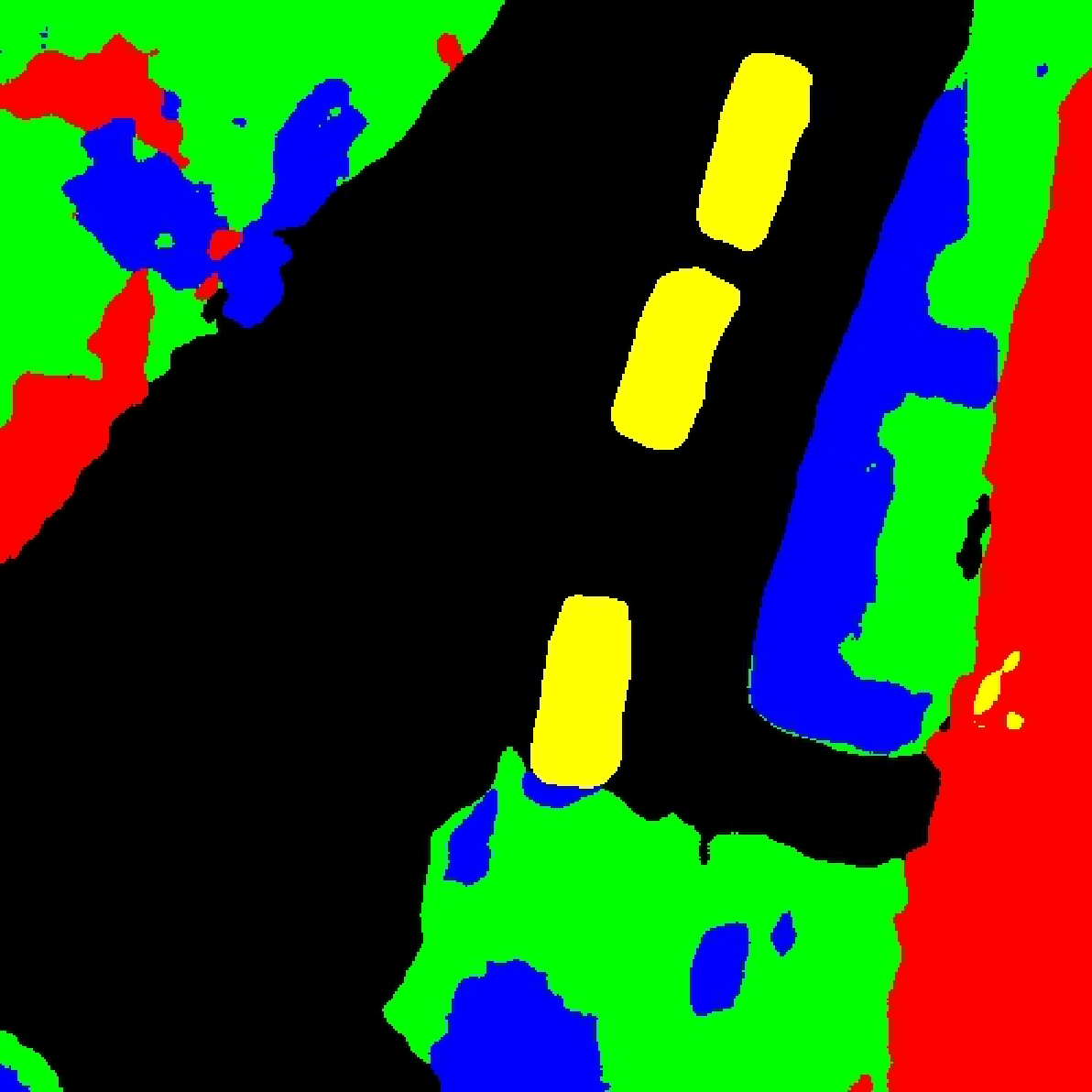}\vspace{4pt}
\includegraphics[width=1\linewidth]{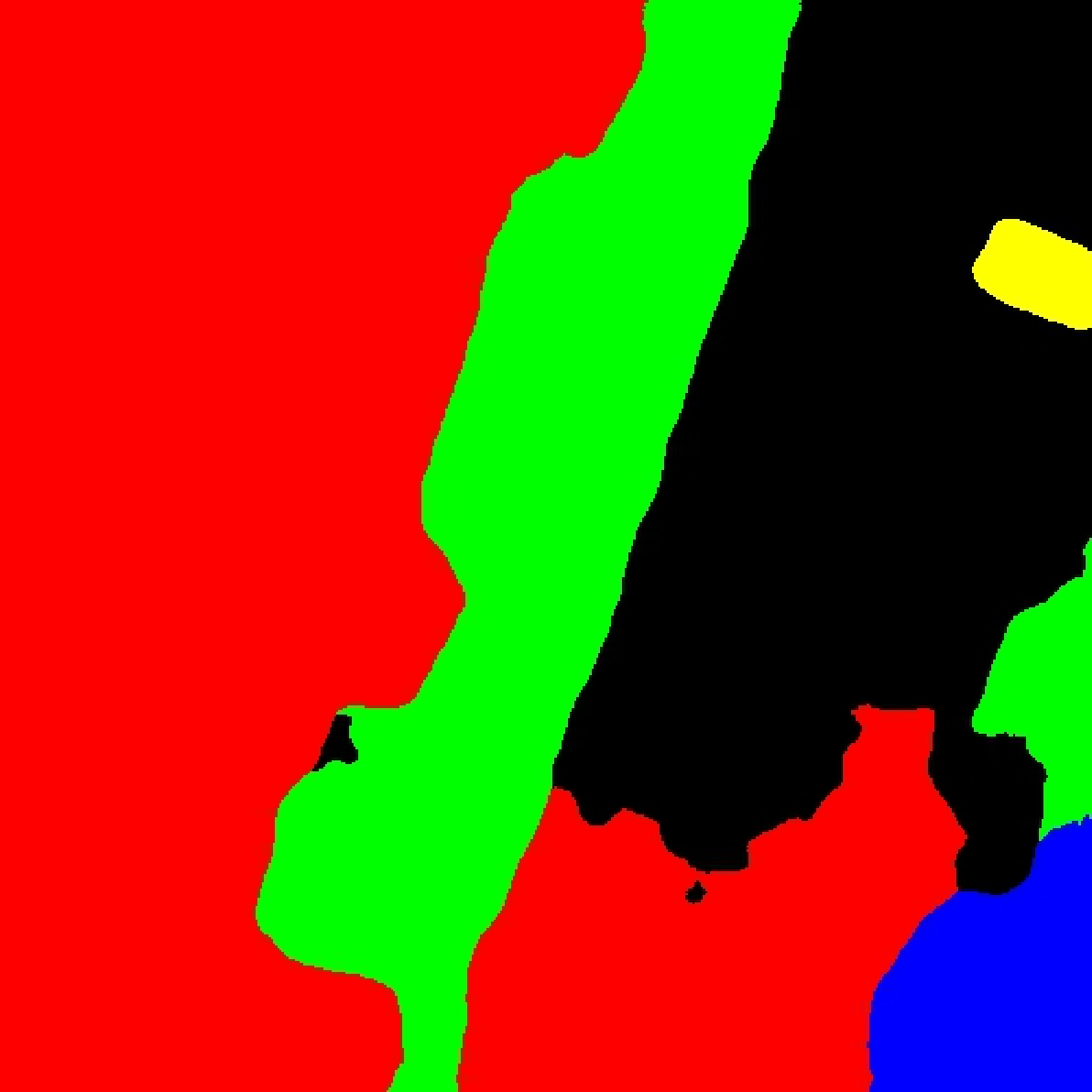}\vspace{4pt}
\includegraphics[width=1\linewidth]{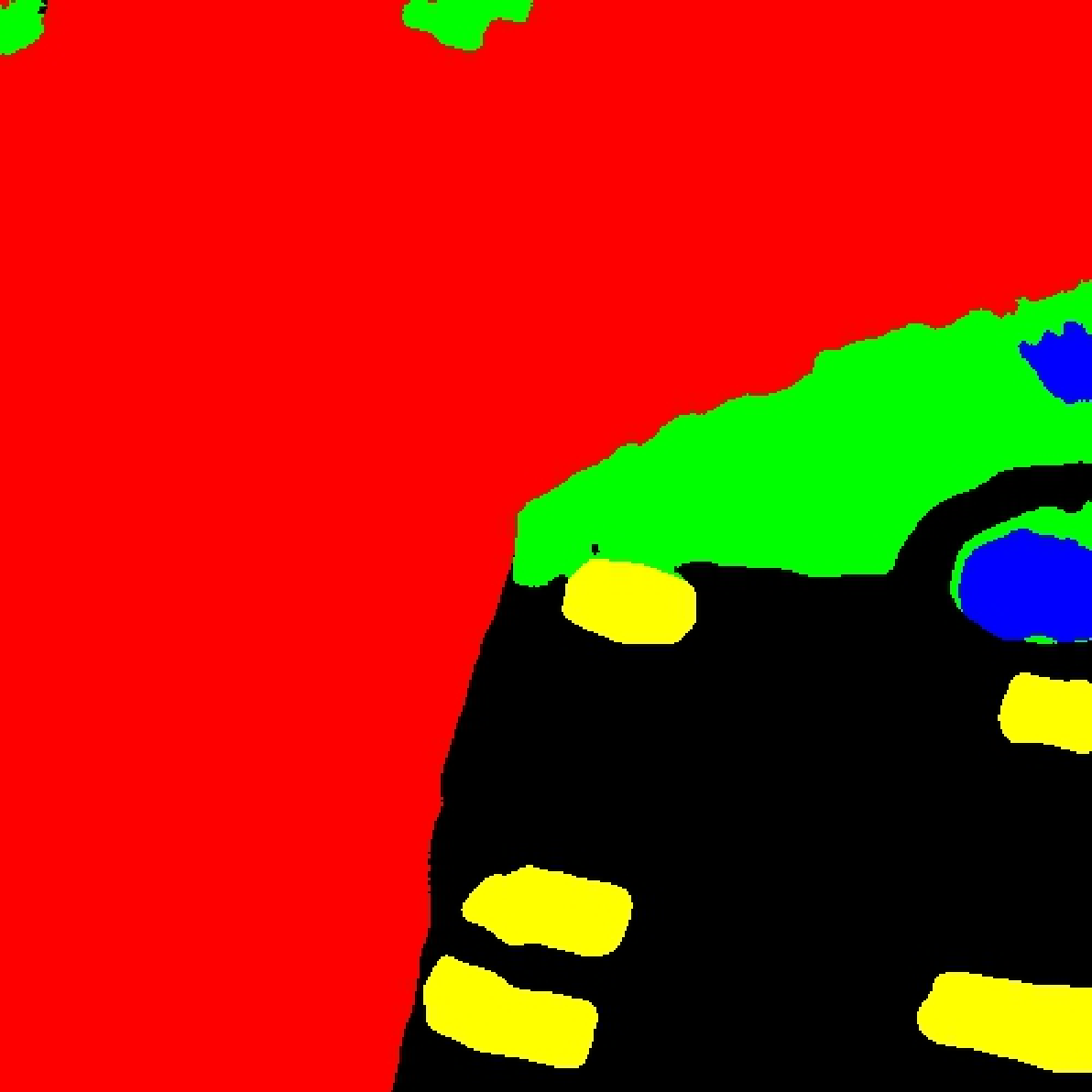}
\end{minipage}}
\subfigure[SRS + ODC]{
\begin{minipage}[b]{0.11\linewidth}
\includegraphics[width=1\linewidth]{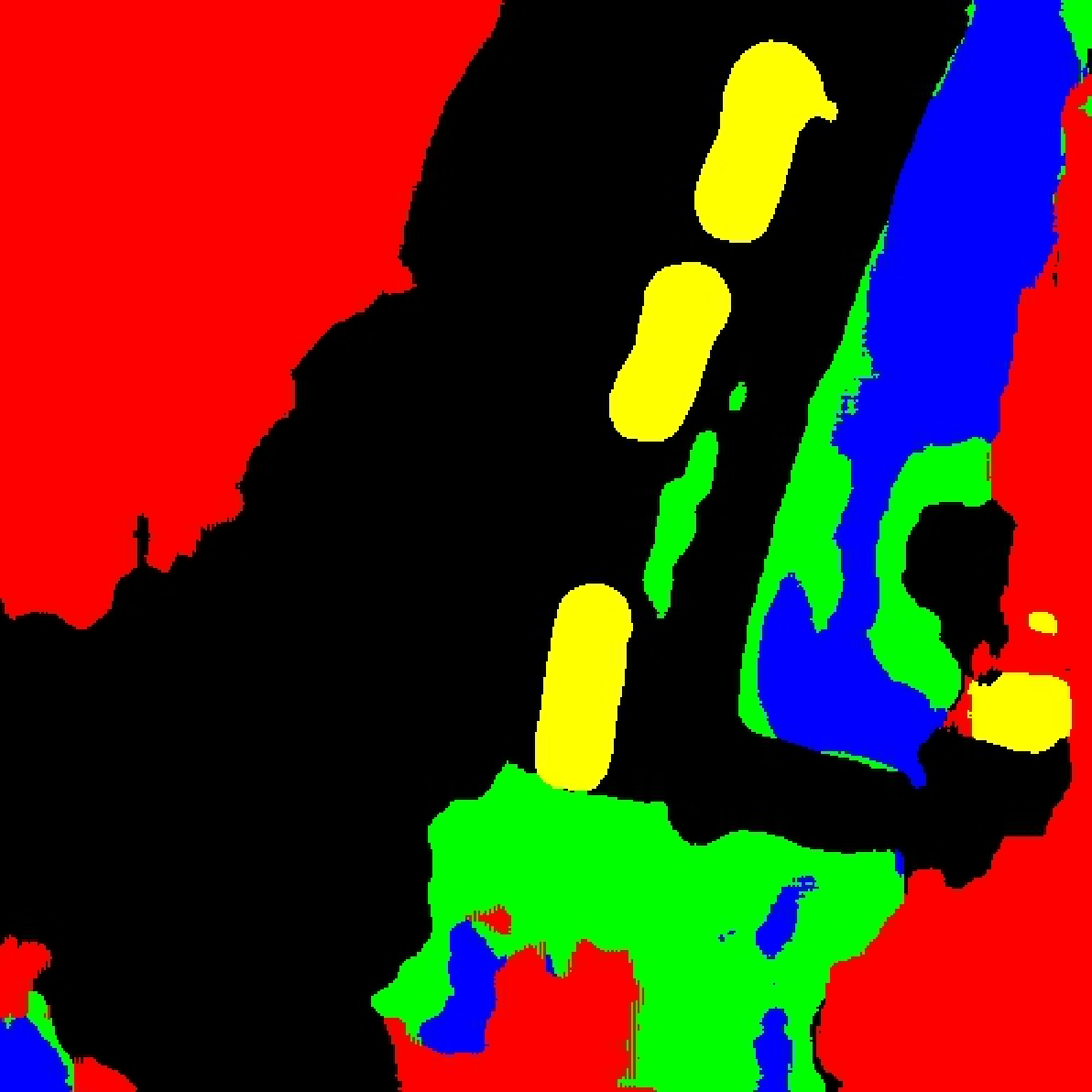}\vspace{4pt}
\includegraphics[width=1\linewidth]{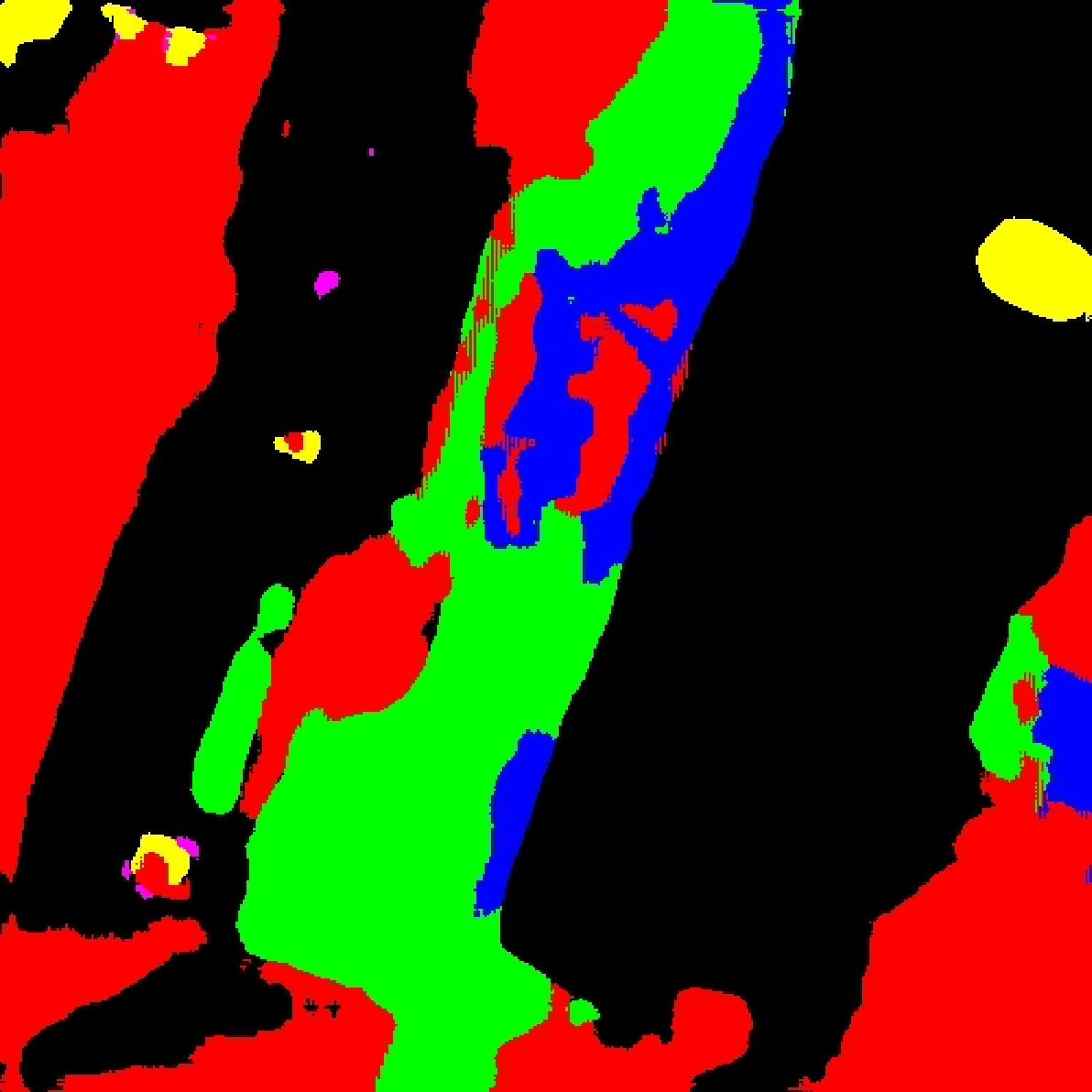}\vspace{4pt}
\includegraphics[width=1\linewidth]{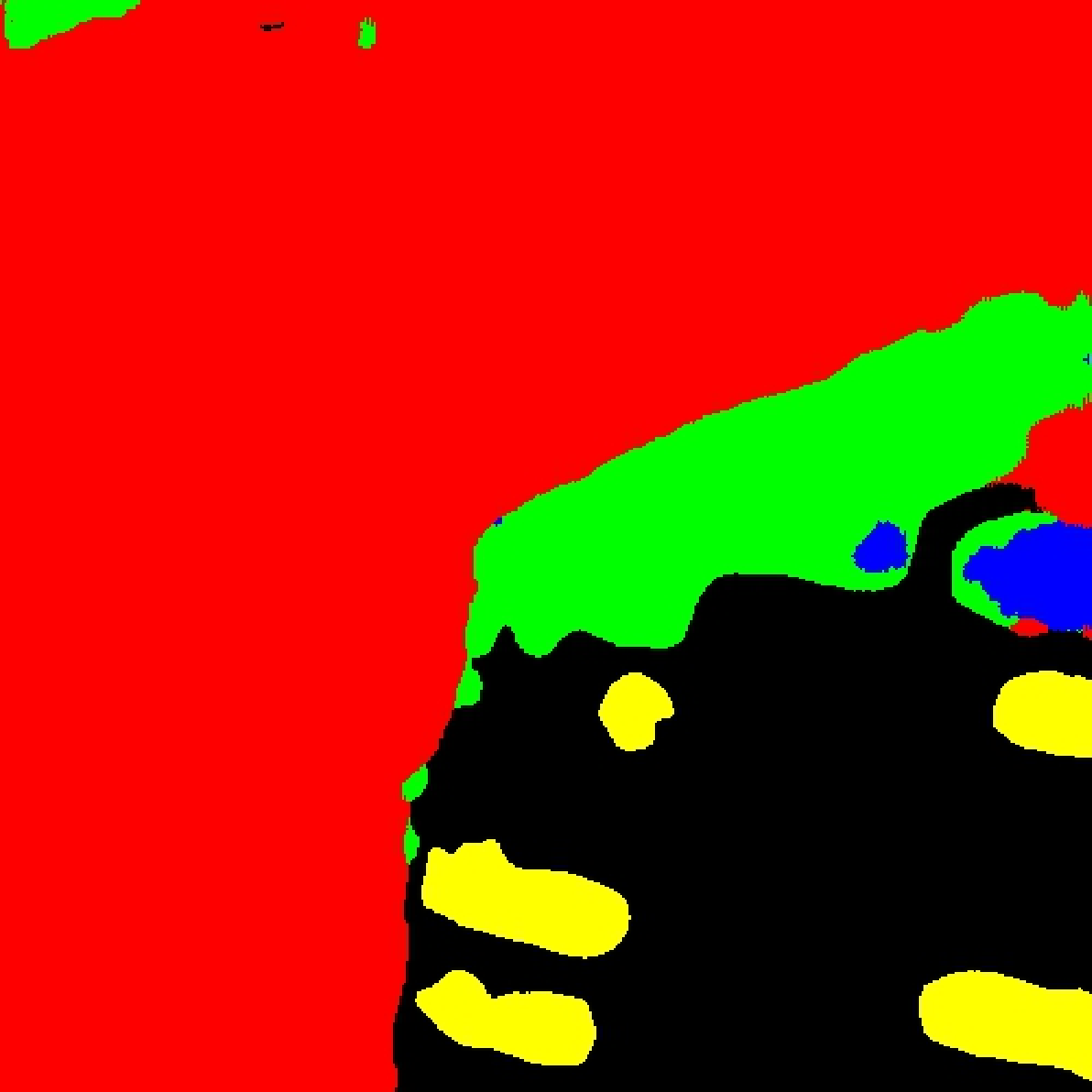}
\end{minipage}}
\subfigure[{\bfseries SRDA-Net}]{
\begin{minipage}[b]{0.11\linewidth}
\includegraphics[width=1\linewidth]{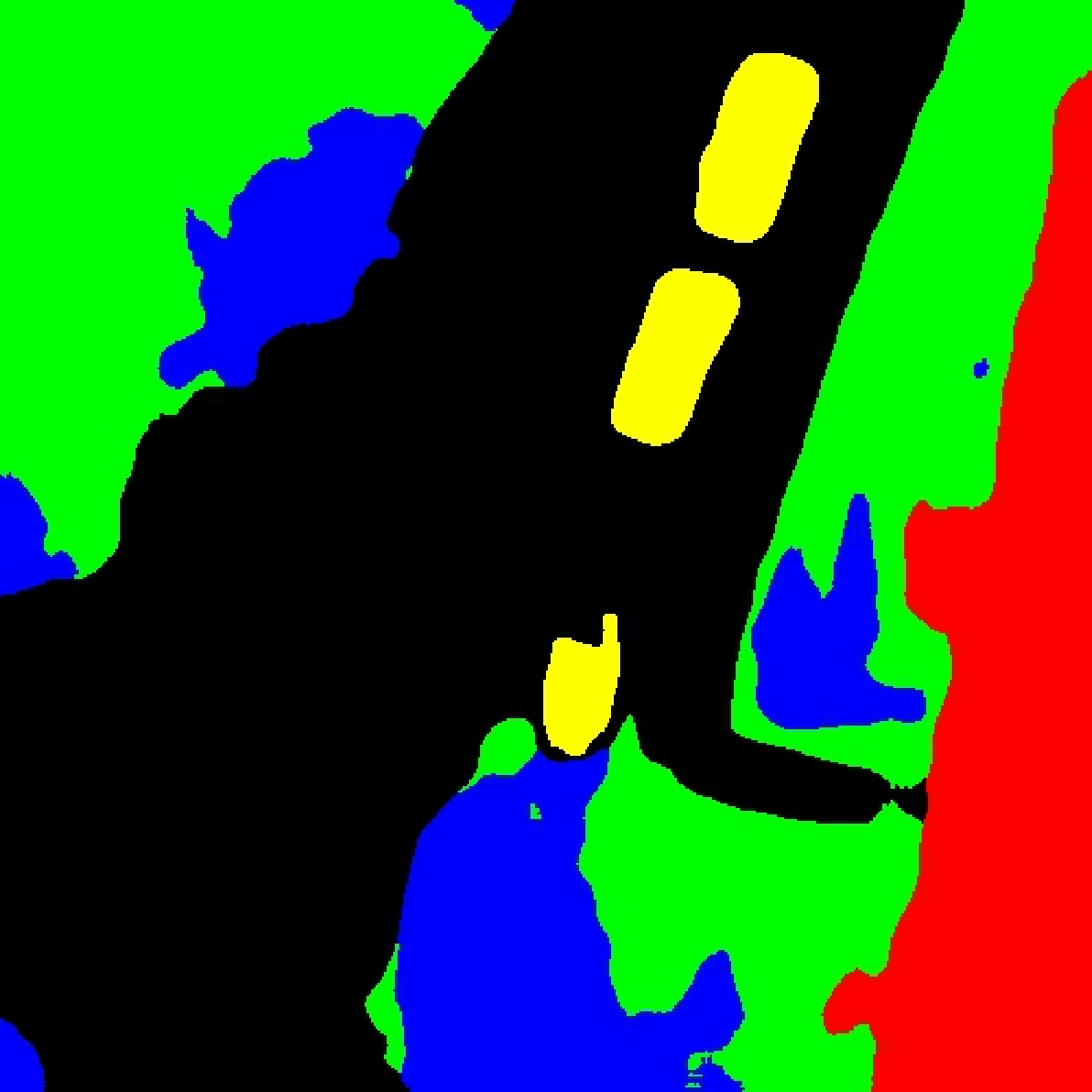}\vspace{4pt}
\includegraphics[width=1\linewidth]{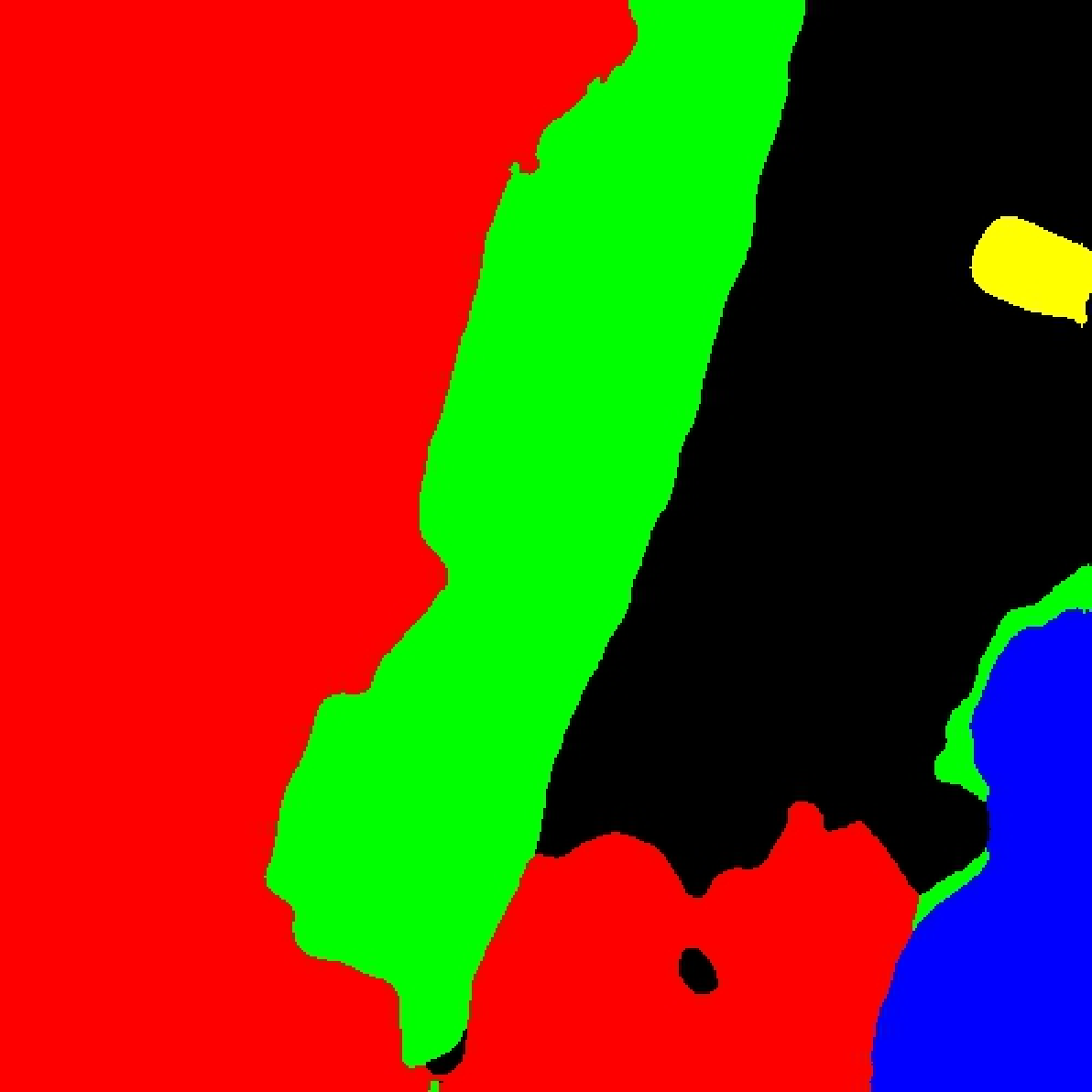}\vspace{4pt}
\includegraphics[width=1\linewidth]{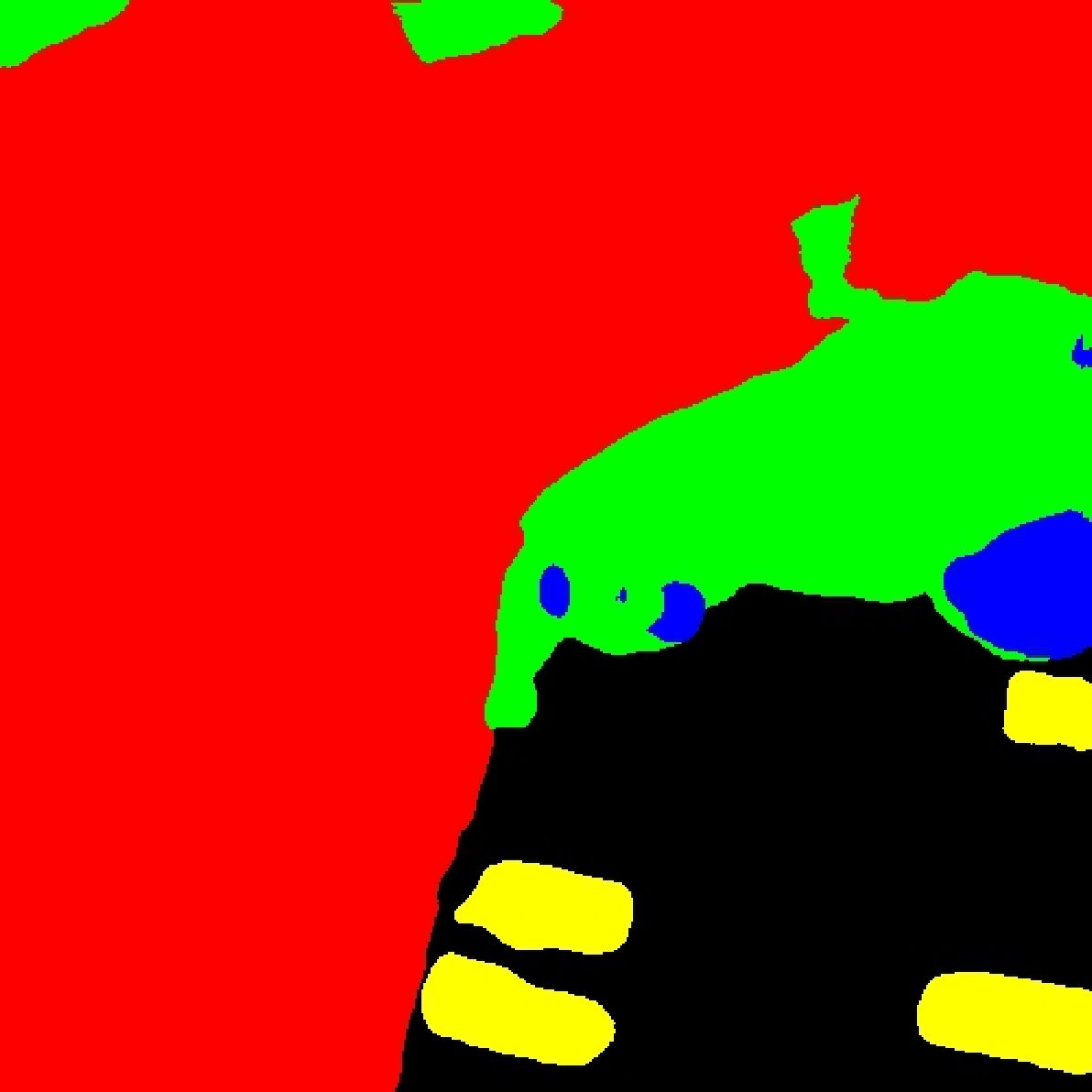}
\end{minipage}}
\subfigure[AdSgNt(45.2)]{
\begin{minipage}[b]{0.11\linewidth}
\includegraphics[width=1\linewidth]{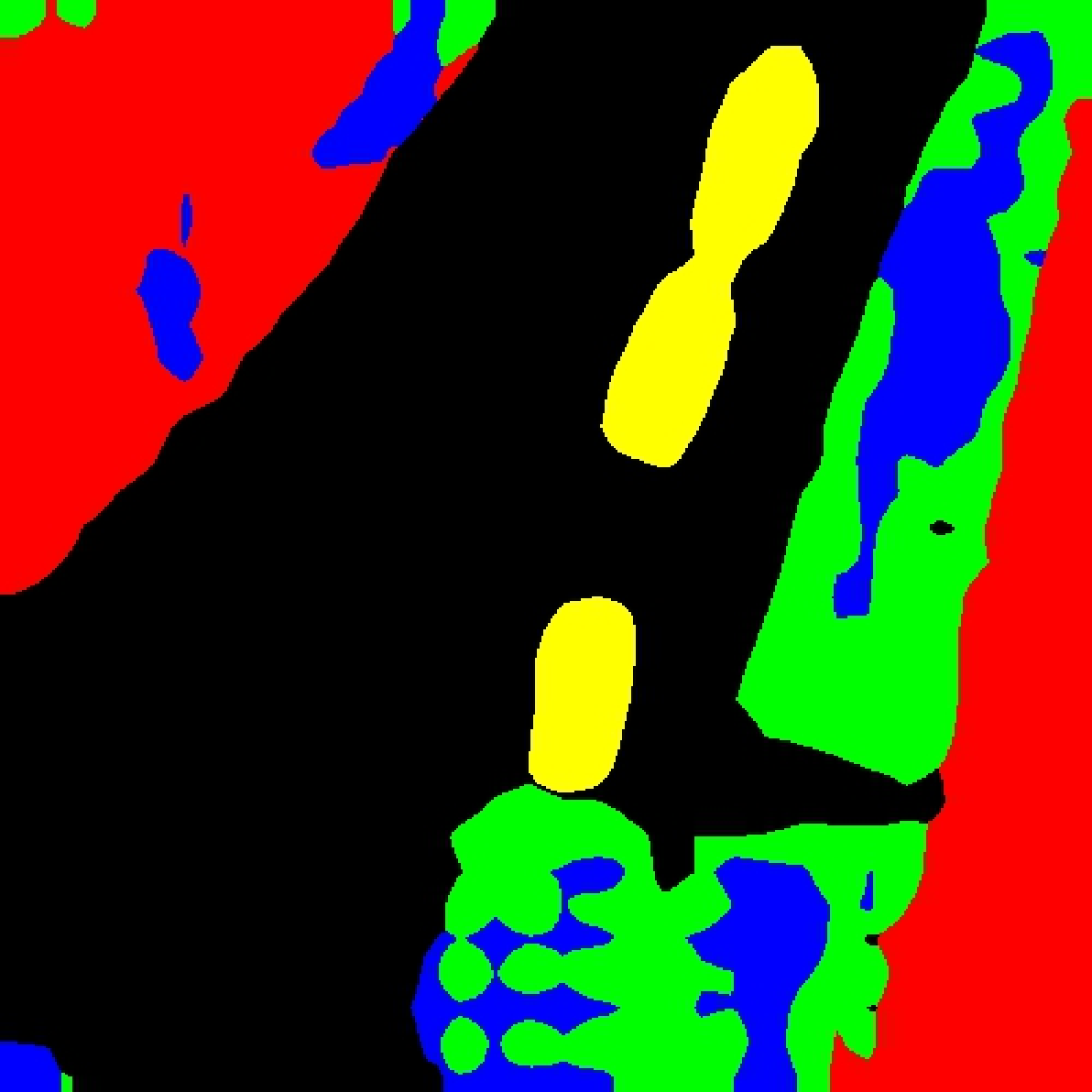}\vspace{4pt}
\includegraphics[width=1\linewidth]{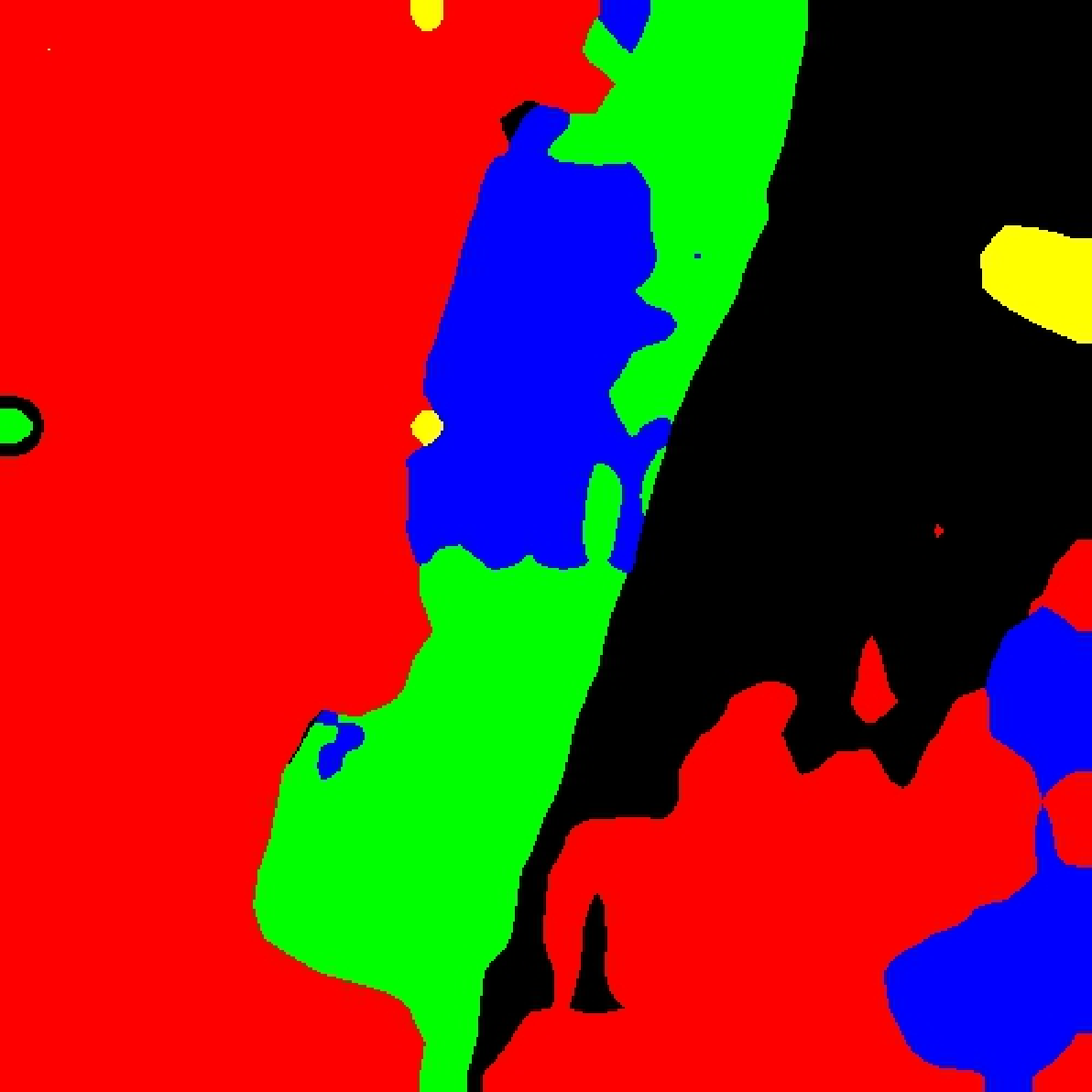}\vspace{4pt}
\includegraphics[width=1\linewidth]{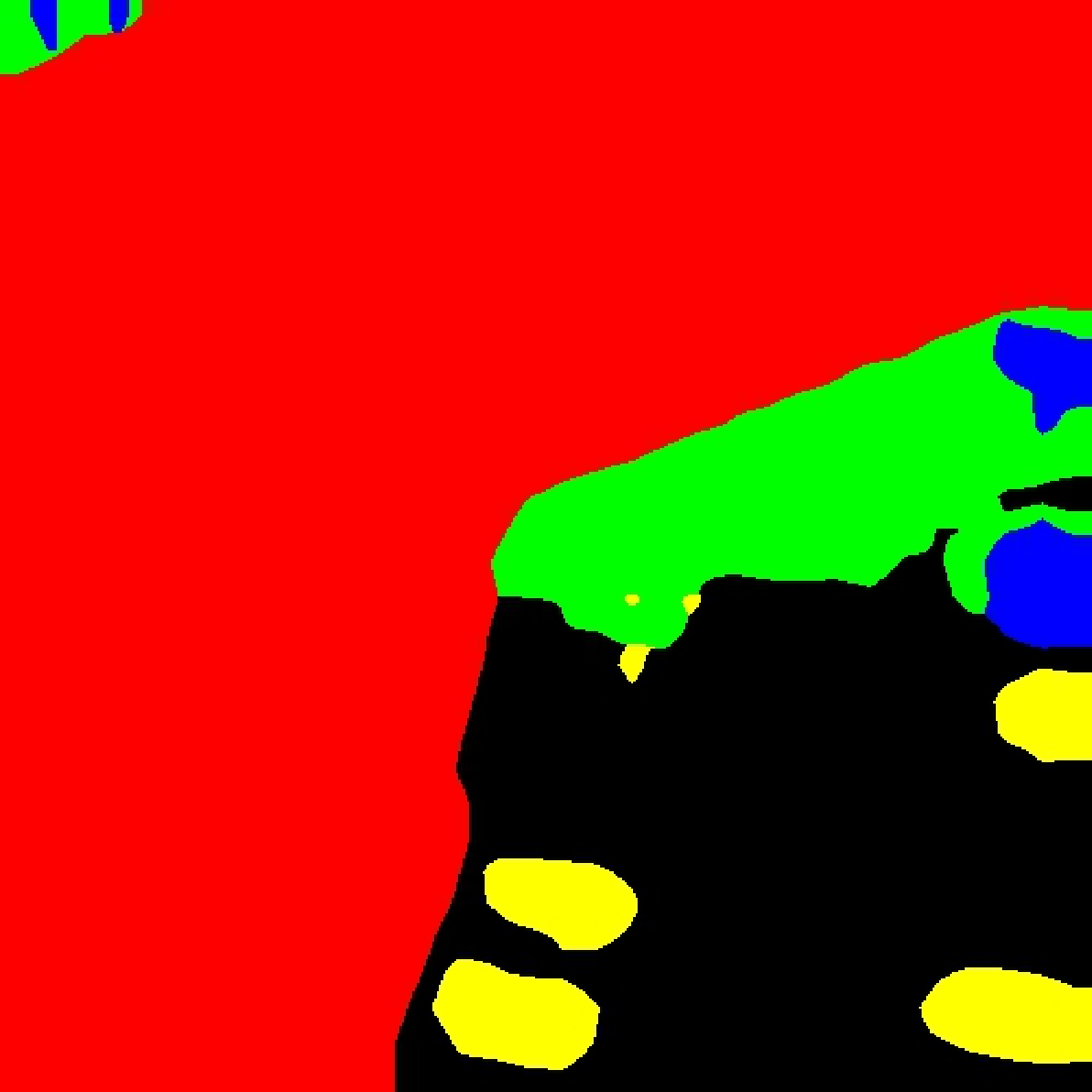}
\end{minipage}}
\subfigure[CFCAN(44.8)]{
\begin{minipage}[b]{0.11\linewidth}
\includegraphics[width=1\linewidth]{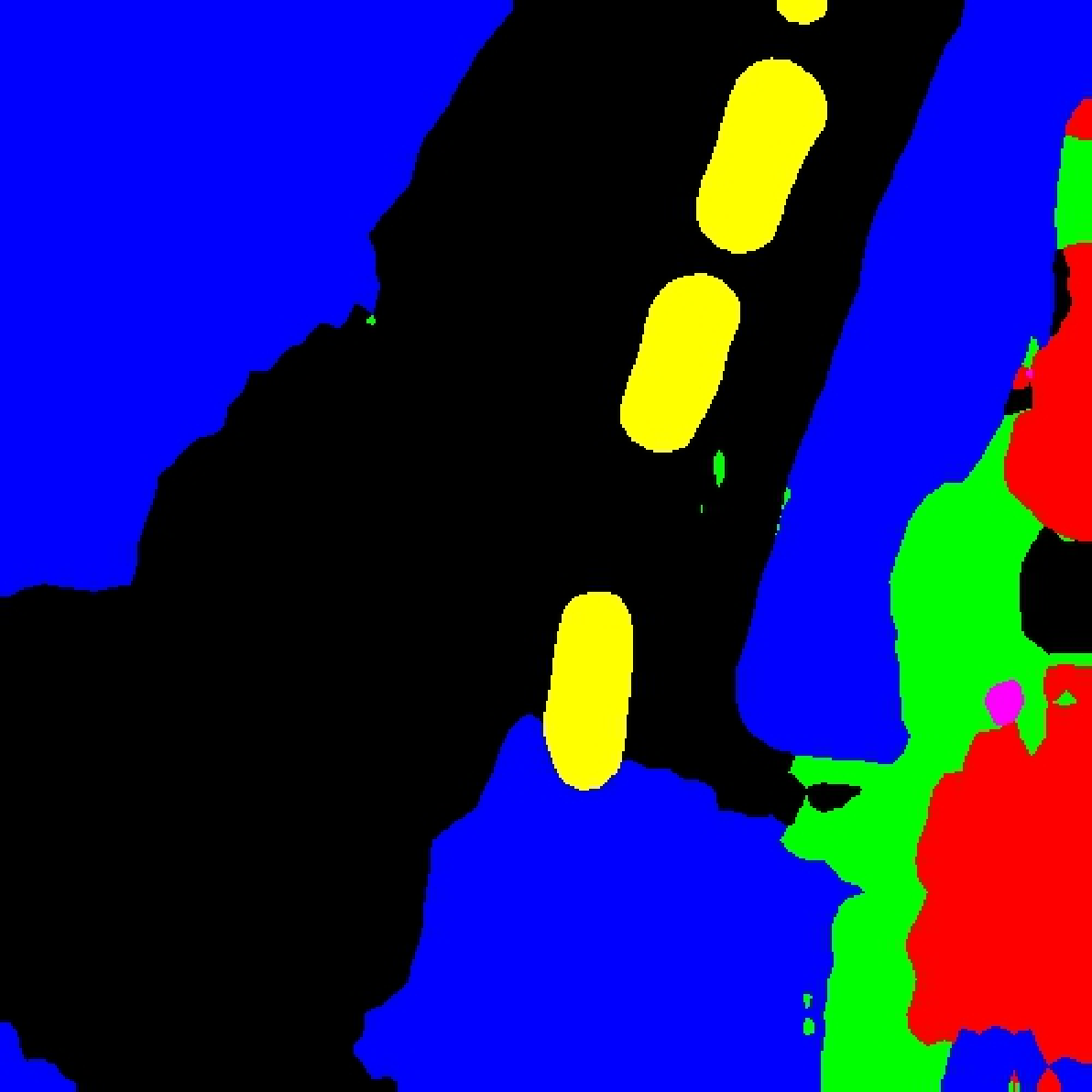}\vspace{4pt}
\includegraphics[width=1\linewidth]{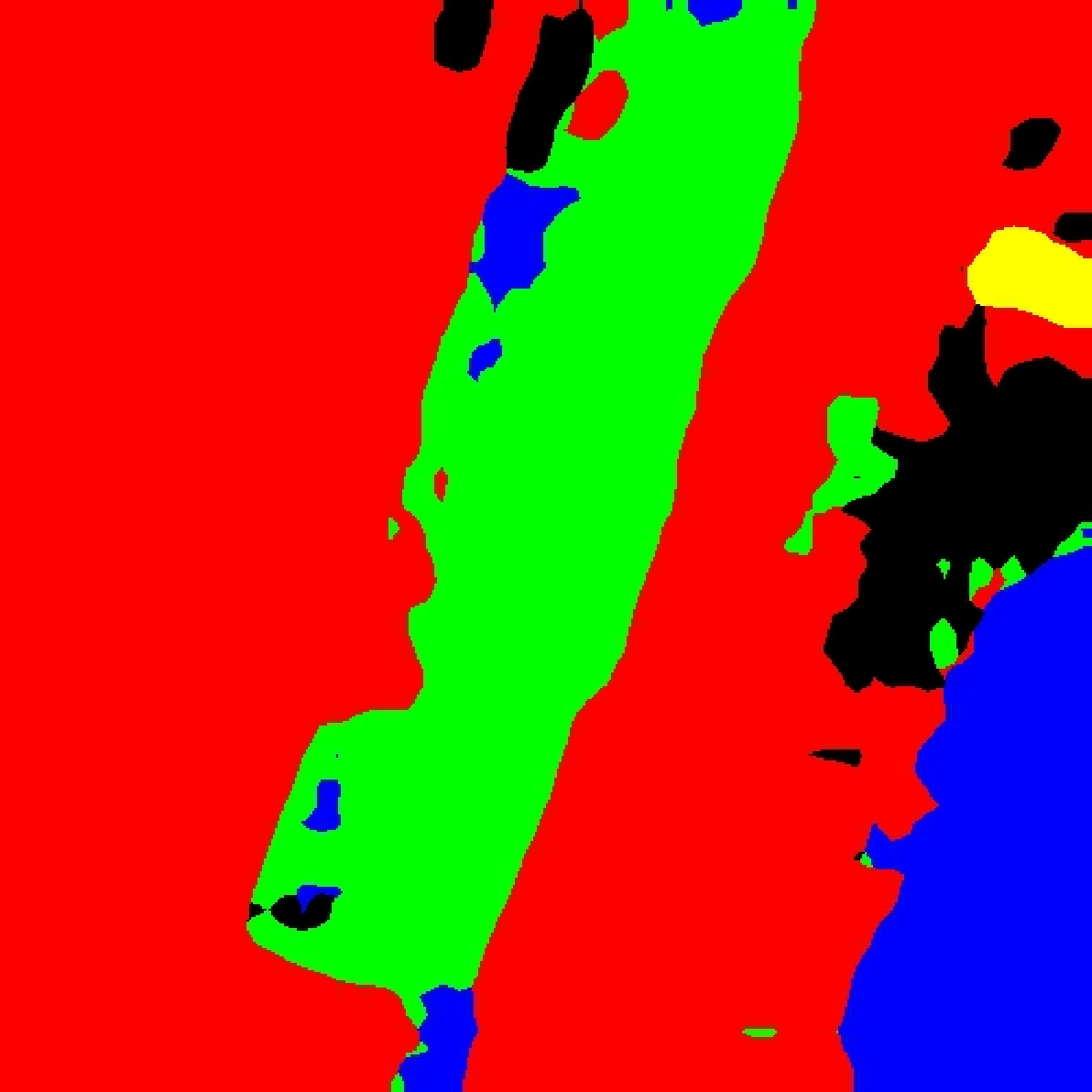}\vspace{4pt}
\includegraphics[width=1\linewidth]{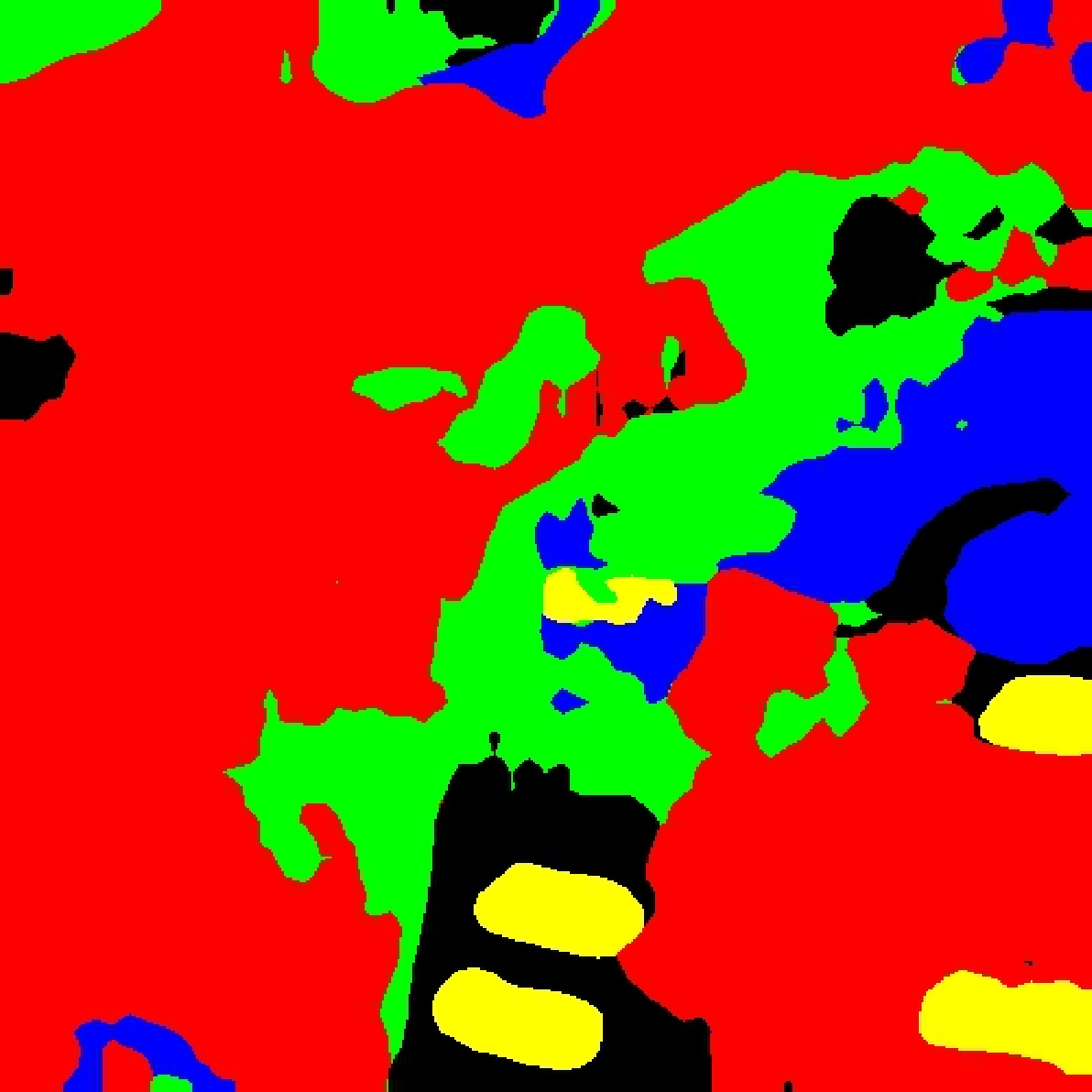}
\end{minipage}}
\caption{Qualitative results of the Pots val dataset (Source domain: ISPRS Vaihingen 2D Semantic dataset)}
\label{fig:PotsResult}
\end{center}
\end{figure*}

%\begin{figure*}
%\vspace{-0.5cm}
%\centering \footnotesize Fig. 4. Qualitative results of the Pots val dataset (Source domain: ISPRS Vaihingen 2D Semantic dataset)
%\end{figure*}

\subsubsection{Implementation Details} 

Network architectures: In SRS, due to GPU memory limitation, we choose the Residual ASPP Module \cite{Wang2019Learning} to capture contextual information, as the shared feature Extractor. For the super-resolution stream, we only use a few deconvolutions to recover the high-resolution images. As for two discriminators, we apply the Patch Generative Adversarial Network (PatchGAN) \cite{Li2016Precomputed} classifier as the PDC Network, and for ODC network we choose is similar to \cite{Tsai2018Learning} which consists of 5 convolution layers with kernel of $4\times4$ and stride of 2, where the channel number is 64, 128, 256, 512, 1, respectively.

Training and testing Details: In the training stage, adam optimization is applied with a momentum of 0.9.  For the Mass to Inria experiments, $\alpha$ is set to 2.5, and $\beta$ set to 10.  Due to different resolution, the Mass images and labels are cropped to 114 $\times$ 114 pixels and then the labels are interpolated to 380 $\times$ 380 pixels. The Inria images are cropped to 380 $\times$ 380 pixels and resized to 114 $\times$ 114 pixels. During the stage of testing, images from Inria are cropped to 625 $\times$ 625 patches without overlap and resized to 188 $\times$ 188. In the Vaih to Pots experiments, $\alpha$ and $\beta$ are set to 5 and 10, respectively. The low-resolution image is cropped to 160 $\times$ 160 pixels and the high-resolution is cropped to 320 $\times$ 320 pixels during training stage. During the testing stage, images of Pots are cropped to 500 $\times$ 500 pixels without overlap and resized to 250 $\times$ 250 pixels. In the actual training process, we first pre-train the model with learning rate 2 $\times {10}^{-4}$. Then the framework is trained with a learning rate of 1.5 $\times {{10}^{-4}}$. For image and label, bicubic interpolation and nearest neighbor interpolation are used respectively.\\

\begin{figure*}[!t]
\centering
\includegraphics[width=17.5cm, height=14.3cm]{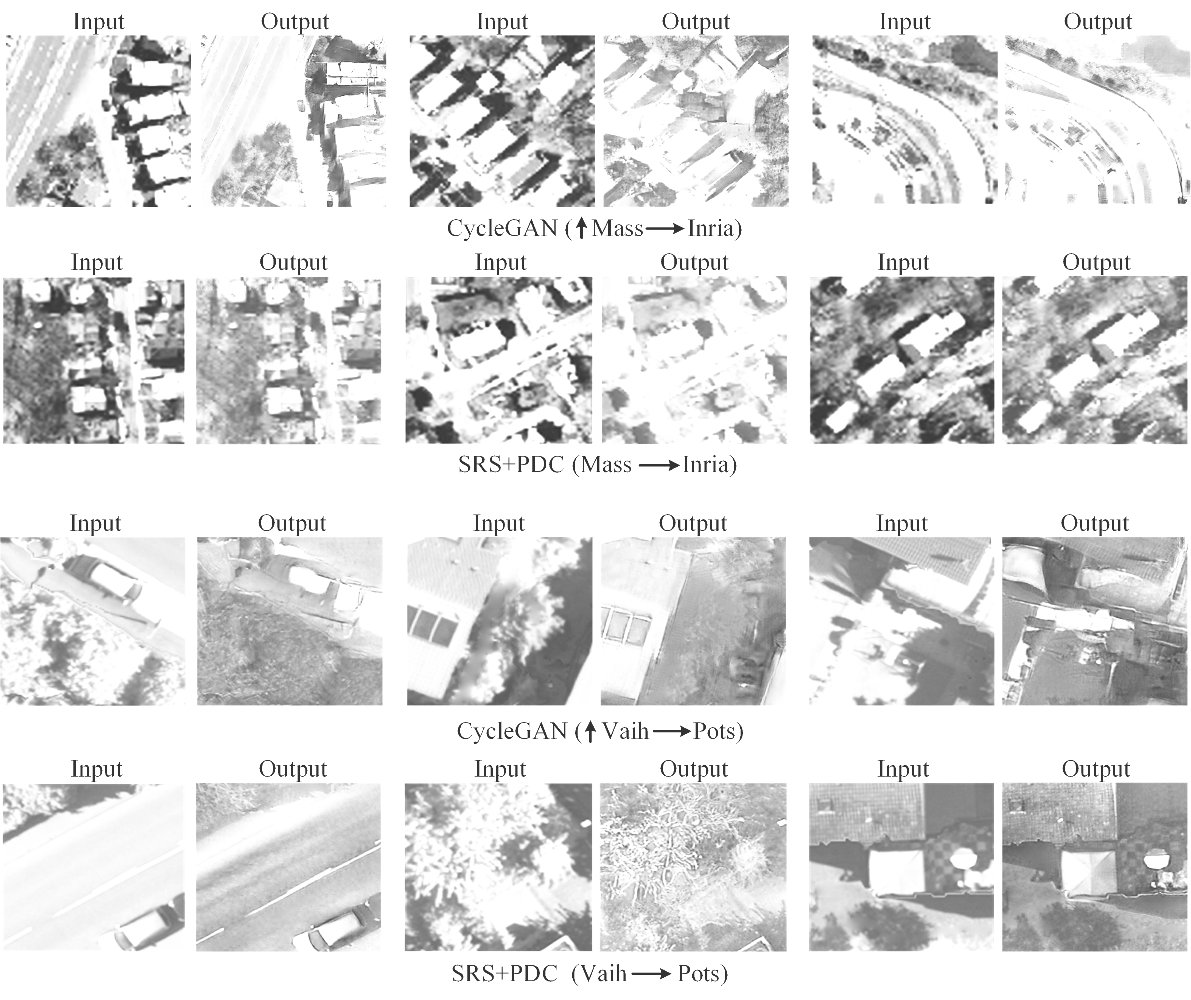}
\caption{Qualitative results of super-resolution with style transfer from CycleGan and SRS+PDC}
\label{fig:superresolution}
\end{figure*}

%\begin{figure*}
%\vspace{-0.5cm}
%\centering \footnotesize Fig. 5. Qualitative results of super-resolution with style transfer from CycleGan and SRS+PDC
%\end{figure*}

{\bfseries Our stepwise experiments.}
\begin{itemize}
    \item {\bfseries NoAdapt:} As a contrast model, NoAdapt is directly trained without domain adaptation from the source domain to the target domain.
	\item {\bfseries SRS:} Based on NoAdapt model, SRS is trained to eliminate resolution gap between source and target domains.
	\item {\bfseries SRS + PDC:}  Based on SRS model, PDC is added furtherly to the training process by the adversarial learning.
	\item {\bfseries SRS + ODC:}  On the basis of SRS model, ODC is introduced into the training process by the adversarial learning.
	\item {\bfseries SRDA-Net (SRS + PDC + ODC):} the proposed SRDA-Net model.\\
\end{itemize}

{\bfseries Other comparison experiments.}
\begin{itemize}
	\item {\bfseries AdaptSegNet :}\cite{Tsai2018Learning}  This work employs the adversarial feature learning in output-space of the base segmentation model. Instead of having only one discriminator over the feature layer, Tsai \emph{et al}. \cite{Tsai2018Learning} propose to install another discriminator on one of the intermediate layers as well.
	\item {\bfseries CycleGan-FCAN: } Fully Convolutional Adaptation Networks (FCAN) \cite{Zhang2018Fully} is a two-stage method, where Appearance Adaptation Networks (AAN) first adapts source-domain images to appear as if drawn from the “style” in the target domain, then Representation Adaptation Networks (RAN) attempts to learn domain-invariant representations. To better adapt the source images to appear as if drawn from the target domain, we replace AAN in FCAN with Cycle Generative Adversarial Network (CycleGan) \cite{Zhu2017Unpaired}.
\end{itemize}
In the experiments, we reported their no adaptation and final results, for comparison with our stepwise experiments.
\subsection{Mass $\rightarrow$ Inria}
The experimental results of the mothods mentioned above for the shift from Mass to Inria are summarized in Table~\ref{tab:MassToInria}, including AdaptSegNet \cite{Tsai2018Learning}, CycleGan-FCAN \cite{Zhang2018Fully} and our stepwise experiments: NoAdapt, SRS, SRS + PDC, SRS + ODC, SRDA-Net. The bold values denote the best scores in the corresponding column.

From the Table~\ref{tab:MassToInria}, it can be seen that our proposed method (SRDA-Net) achieves the best result: IoU of 52.8\%. Under the same training condition, result (52.8\%) of SRDA-Net outperforms that of AdaptSegNet (best result of 48.5\%) and CycleGan-FCAN (best result of 49.7\%), increased by $8.87\%$ and $6.24\%$ respectively. Moreover, in order to explore the effect of resolution problem on the domain adaptation results, we construct experiments on two training data settings (source domain: Mass, target domain: Downsampled Inria; source domain: Upsampled Mass, target domain: Inria) for each comparison method. From the results, we observe that the setting (source domain: Upsampled Mass, target domain: Inria) achieves better result, which confirms that when the resolution difference between the source and target domains is larger, the gain obtained by eliminating the resolution difference is greater than the error introduced by interpolation. In comparison, our method does not need to consider this, and obtained better results.

As for the results compared with baseline method, the adaptation results of the three methods outperform their corresponding results of NoAdapt. Besides, SRS improves the IoU significantly, from 31.9\% to 36.7\%, increasing by 4.8\% , which shows the effectiveness of combining image super-resolution in segmentation network to eliminate the resolution gap between source and target domains.

In order to explore the semantic segmentation performance further, Figure~\ref{fig:InriaResult} shows the visualization results of our step-by-step and the AdaptSegNet/CycleGan-FCAN methods. The images in the first column are selected from the Inria val dataset. The second column shows the ground truth, and the remaining columns illustrate the prediction results of SRS, SRS + PDC,  SRS + ODC, SRDA-Net, AdaptSegNet (48.5\%), CycleGan-FCAN (49.7\%). On the whole, after adding the PDC or ODC, some segmentation mistakes are removed effectively. According to the results of SRS + PDC and SRS + ODC, PDC plays a more important role in learning domain-invariant features than ODC (improvement of 9.3\% versus 2.7\%). When the domain gap is reduced by integrating PDC and ODC to SRS, a better segmentation result can be obtained. Moreover, we can observe that the visualization segmentation results of SRDA-Net outperform results of best AdaptSegNet/CycleGan-FCAN.

\subsection{Vaih $\rightarrow$ Pots}
The results of AdaptSegNet \cite{Tsai2018Learning}, CycleGan-FCAN \cite{Zhang2018Fully} and our stepwise experiments are listed in Table~\ref{tab:VaihToPots},  which are adapted from Vaih to Pots. The bold fonts represent the best scores of the corresponding columns. It can be observed that the proposed method obtains the best performance with mIoU of 48.6\%. Compared with the best comparative result (45.2\%, obtained by AdaptSegNet), the SRDA-Net contributes 3.4\% relative mIoU improvement. Under the condition of no adaptation, among the three baseline methods, mIoU of our method is slightly lower. This is because that parameters of Residual  ASPP Module are less than half of resnet101, which limits the learning power of network.
According to the mIoUs of SRS+PDC and SRS + ODC, PDC (18.7\%) is more effective at learning domain-invariant features than ODC (1.7\%).

From the results of comparison experiments, we find that when the gap of resolution between source and target domains is relatively small, it is difficult to determine whether to upsample the source domain or downsample the target domain to obtain better results. However, there is no need to our proposed method (SRDA-Net).

For reporting the effect of our algorithm, Figure~\ref{fig:PotsResult} gives the three typical example labeling results. From the visual results, we can find that the segmentation results are getting more and more refined in our step-by-step experiments. And our SRDA-Net obtains the finer segmentation results.

%SRDA-Net vs. CycleGan

\begin{table}
%\vspace{1mm}
\renewcommand{\arraystretch}{1.5} % Control the heigh
\begin{center}
\caption{Ablation experiments of SRS}\label{tab:Ablation}
%\vspace{1.5mm}
%\footnotesize
\setlength{\tabcolsep}{6pt}  % Control the width
\begin{tabular}{cccccccc}
 \toprule[1pt]
  $ $  &  NoAdapt &  +strategy1  & +strategy2 &   SRS   \\
\hline
IoU &   31.9   &   33.5  &   36.2  &  36.7   \\
\bottomrule[1pt]
\end{tabular}
\normalsize
\end{center}
\end{table}
\subsection{Study of two strategies in SRS}

To make super-resolution and segmentation promote each other, we propose two strategies in the SRS model: (1) a pyramid feature fusion structure between the two tasks; (2) a cross-entropy segmentation loss is applied to the generated high-resolution source domain images to train the segmentation network. In this section, we construct ablation experiments of SRS and SRS+PDC vs CycleGan experiments to illustrate the effectiveness of two strategies. From the Table \ref{tab:Ablation}, we can observe that both strategies improve the segmentation performance compared to the baseline model (NoAdapt). The SRS (strategy1 + strategy2) achieves the best segmentation accuracy, which shows that both strategies can transfer detailed information from super-resolution to improve segmentation performance.

\subsection{SRS+PDC vs. CycleGan}
The SRS+PDC model is essentially a super-resolution style transfer model. The qualitative super-resolution results of source domain images with style transfer from CycleGan and SRS+PDC are shown in Figure~\ref{fig:superresolution}, in order to illustrate the super-resolution style transfer effect of SRS+PDC. Despite of superimposing the style of target domain, CycleGan generates monotonous and unnatural textures, like buildings in Figure~\ref{fig:superresolution}. Moreover, we find that some objects in the results of CycleGan get distorted, like cars in Figure~\ref{fig:superresolution}. The reason is that upsampled source domain images are blurry, which drops some information. SRS utilizes semantic category priors to help capture the characteristics of each category and produce more natural and realistic textures, which suggests that both strategies can leverage the semantic information of segmentation to improve super-resolution results. And by adversarial training with PDC, the super-resolution images of SRS simultaneously achieve style conversion.

\section{Conclusion}
In this paper, we propose a novel end-to-end framework named SRDA-Net to explicitly address the resolution adaptation problem in the field of semantic segmentation. SRDA-Net can simultaneously deal with the super-resolution task and the domain adaptation task, thus meeting the requirement of semantic segmentation for remote sensing images which usually involve various resolution images. To be specific, a multi-task model is built to simultaneously accomplish the semantic segmentation, as well as eliminate the difference in resolution between source and target domains. By means of the adversarial learning, the pixel-level and output-space domain classifiers are designed to guide the SRS model to learn domain-invariant features, which can eliminate the domain gap effectively. In order to verify the effectiveness of the proposed method, two datasets are constructed which have different resolutions in their source and target domains: Mass-Inria and Vaih-Pots. Extensive experiments demonstrate the effectiveness of SRDA-Net when domain adaptation involving the resolution difference.
\ifCLASSOPTIONcaptionsoff
  \newpage
\fi

\bibliographystyle{IEEEtran}
%\bibliography{ref}

\end{document}